\documentclass[10pt,a4paper]{article}
\usepackage[utf8]{inputenc}
\usepackage[english]{babel}

\usepackage{graphicx}
\usepackage[left=16mm,right=16mm,top=25mm,bottom=25mm]{geometry}

\usepackage{amssymb}         
\usepackage{amsmath}         
\usepackage{mathtools}			
\usepackage{cancel}
\usepackage{float}

\newtheorem{proof}{Proof}

\usepackage{xcolor}

\usepackage{bm}             
\usepackage{nicefrac}       
\usepackage{multirow}       

\usepackage[lined,linesnumbered,ruled]{algorithm2e} 

\usepackage{pifont}         

\usepackage{cancel}

\usepackage[colorlinks = true,linkcolor  = black,urlcolor   = blue,citecolor  = black]{hyperref}
            
\usepackage{authblk}    

\newenvironment{changemargin}[2]{%
\begin{list}{}{%
\setlength{\topsep}{0pt}%
\setlength{\leftmargin}{#1}%
\setlength{\rightmargin}{#2}%
\setlength{\listparindent}{\parindent}%
\setlength{\itemindent}{\parindent}%
\setlength{\parsep}{\parskip}%
}%
\item[]}{\end{list}}

\newcommand{\bdelta}{\bm{\delta}}            
\newcommand{\btheta}{\bm{\theta}}            
\newcommand{\bphi}{\bm{\phi}}                

\newcommand{\Eq}[1]{Eq.~(#1)}           
\newcommand{\Fig}[1]{Figure~#1}         
\newcommand{\Tab}[1]{Table~#1}          
\newcommand{\Sec}[1]{Section~#1}        
\newcommand{\App}[1]{Appendix~#1}       

\newcommand{\cmark}{{\color[rgb]{0,0.5,0}\ding{51}}} 
\newcommand{\xmark}{{\color[rgb]{0.8,0,0}\ding{55}}} 

\newcommand{\coma }{,}                       
\newcommand{\punto}{.}                       

\usepackage{multicol}			
\usepackage{titling}			

\begin{document} 

\title{\huge{\textbf{Information Assisted Dictionary Learning \\for fMRI data analysis}}}

\author{Manuel~Morante$^a$, Yannis~Kopsinis$^b$, Sergios~Theodoridis$^{c,d}$, Athanassios~Protopapas$^e$}
\date{%
    $^a$Computer Technology Institutes \& Press ``Diophantus'', Patras (Greece)\\
    $^b$LIBRA MLI Ltd, Edinburgh (UK)\\
	$^c$Kapodistrian University of Athens (Greece)\\
	$^d$Chinese University of Hong Kong, Shenzhen, (China).\\
	$^e$University of Oslo (Norway)
}

\maketitle
\mbox{}
\vspace{15mm}

\begin{abstract}
\textbf{
  In this paper, the task-related fMRI problem is treated in its matrix factorization formulation, focused on the Dictionary Learning (DL) approach. The new method allows the incorporation of a priori knowledge associated both with the experimental design as well as with available brain Atlases. Moreover, the proposed method can efficiently cope with uncertainties related to the HRF modeling. In addition, the proposed method bypasses one of the major drawbacks that are associated with DL methods; that is, the selection of the sparsity-related regularization parameters.  In our formulation, an alternative sparsity promoting constraint is employed, that bears a direct relation to the number of voxels in the spatial maps. Hence, the related parameters can be tuned using information that is available from brain atlases. The proposed method is evaluated against several other popular techniques, including GLM. The obtained performance gains are reported via a novel realistic synthetic fMRI dataset as well as real data that are related to a challenging experimental design.}
\end{abstract}

\newpage

\begin{multicols}{2}

\section{Introduction}
\label{Sec:Intro}
In order to perform several actions/tasks, the human brain relies on the simultaneous activation of many \emph{Functional Brain Networks} (FBN), which are engaged in proper interaction to execute the tasks effectively. Such networks, potentially distributed over the whole brain, are defined as \emph{segregated} regions exhibiting high functional connectivity. The latter is quantified via the underlying correlations among the associated activation/deactivation time patterns, which are referred to as {\it time courses}  \cite{Exploratory}. Examples of brain networks are the visual, somatosensory, neuronal default mode, dorsal attention, and executive control networks. Functional Magnetic Resonance Imaging (fMRI) is the dominant data acquisition technique for the detection and study of FBNs \cite{Handbook}. fMRI measures the Blood Oxygenation Level-Dependent (BOLD) contrast \cite{FunMRI} that constitutes the evoked hemodynamic response of the brain to the corresponding neuronal activity. This process can be modeled as a convolution between the actual neuronal activation and a \emph{person-dependent} impulse response function, called Hemodynamic Response Function (HRF). The fMRI captures 3D images with a typical resolution being $64\times 64\times 32$ voxels per image, whereas a sequence of 200 to 300 images are acquired per session (usually, one image every one or two seconds). The obtained fMRI measurements, associated with each voxel, comprise a mixture of the time courses corresponding to those FBNs, which are active at the specific voxel. Moreover, beyond the brain-induced sources, additional machine-induced interfering sources are also present that contribute to the measured mixture. The ultimate goal of the fMRI analysis is to detect the set activated of voxels, referred to as {\it spatial map}, in which each brain source of interest manifests itself in revealing the corresponding FBN. 

In the case of block- or event-related experimental designs, i.e., when the subject is presented with a fixed number of pre-selected stimuli, which are referred to as \emph{conditions}, the shape of the time courses that correspond to the  conditions of the experimental design are estimated as the convolution of the conditions with the canonical Hemodynamic Response Function (cHRF) \cite{SPM-web}. Hereafter, such time courses are referred to as \emph{task-related time courses}.

A prominent approach for fMRI data analysis is via Blind Source Separation (BSS), which is usually performed via appropriate matrix factorization schemes \cite{Sergios}. Independent Component Analysis (ICA) \cite{ICA-Decade}, \cite{ICA-Novel}, \cite{ICA-Novel-2}, and Dictionary Learning (DL) are the most popular paths. A drawback of ICA is the independnce assumption, which can be violated in fMRI, especially in high spatial overlap, \cite{Adali-Answer}, \cite{Daubechies-Q}, \cite{Sparse-SPM}. Unlike the independence hypothesis, DL relies on sparsity that is a reasonable assumption concerning the brain activity \cite{Sparse-2018}, \cite{Sparse-Brain}, \cite{Multi-Sparse}.


However, DL approaches are not beyond shortcomings. The tuning of the associated regularization parameters is not easy in practice and it is usually performed via cross validation techniques;  however, in real experiments, this is not possible due to the lack of ground truth data, \cite{Sergios,SPAMS,Yannis-DL,ICASSP-Paper,DL_2017}. Therefore, the only way to proceed with the parameters' fine-tuning is via the visual inspection of the results, a process that requires the active judgment of the user; this can be inconsistent and susceptible to errors  that may hamper the adoption of DL approaches in practice.

Alternatively, another candidate family of BSS for the fMRI data analysis is the Non-negative Matrix Factorization (NMF) approach~\cite{Comparisons, NMF, S-NMF, T-NMF}. Unlike the previous techniques, the NMF methods impose a non-negativity constraint over the matrix factorization. Nevertheless, the non-negativity constraint may not be valid in practice~\cite{NMF-Issue_2015}. In the fMRI framework, the aim of the non-negativity constraint conceptually consists of eliminating the negative contribution of the BOLD signal response, leaving only the positive activations~\cite{Comparisons}, which potentially weakens the performance of the NMF methods. Furthermore, NMF algorithms, often, require the tunning of several regularization parameters, sharing the same limitations related to standard DL techniques.

Conventional analysis of fMRI data relies on approaches focused on the General Linear Model (GLM), which assumes the prior availability of the task-related time courses \cite{TotalActivation}. This approach suffers from a critical limitation: It assumes that the HRF is known and fixed, whereas the truth is that the HRF significantly varies across subjects \cite{HRF-Var}, \cite{HRF-Var-2}. In contrast, BSS methods do not need to make any assumption regarding the HRF. Moreover, they inherently model interfering artifacts, such as machine-induced artifacts, uncorrected head-motion residuals or other unmodeled physiological signals that may obscure the brain activity of interest. In addition, they can discover other brain-induced sources beyond the task-related ones.

Despite their advantages, BSS methods share a major drawback compared to GLM: when two or more task-related sources manifest themselves in highly overlapped brain regions, ICA, to a larger, and DL to a smaller extent, can fail to discriminate them. This situation is typical in most experimental designs of interest, in which the different experimental conditions can activate the same major functional brain network, e.g., the auditory, the visual, the somatomotor, etc. In contrast, GLM-based methods offer the option of studying the \emph{contrast} among different conditions, aiming at detecting brain areas, which are activated due to a specific task. 

In an attempt to overcome the aforementioned fundamental drawbacks of the BSS methods against GLM, efforts have been made to create new approaches \cite{2005_SPM-ICA,Sparse-GLM}. In the ICA framework, the most relevant is to impose tast-related information. Collectively, such methods are referred to as constrained ICA~\cite{Semi-ICA-1, Semi-ICA-Calhoun, CSICA_2010, CSICA_2010, CSTICA_2014, CSTICA_2018}. Although constrained ICA methods often lead to enhanced performance, compared with their fully blind counterparts~\cite{2006_ICA-Ref}, they suffer from a critical limitation: the embedded constraint, e.g., the imposed task-related time courses, must not violate the independence assumption. This requirement poses stringent constraints either on the total number of allowable time sequences, e.g. \cite{Morpho-ICA}, or on the nature of the imposed time courses, which need to be independent of each other \cite{Lu_2005,CSTICA_2009,CSTICA_2014,CSTICA_2018}. Both restrictions heavily limit the application of constrained ICA to fMRI, since the most common case is  to have experimental designs that comprises more than two BOLD sequences. Moreover, they often correspond to significantly overlapping regions and, therefore, to highly statistically dependent sources. Furthermore, in contrast to many unconstrained ICA algorithms, which require a reduced number of relatively easy to tune parameters, all the constrained ICA algorithms require extensive regularization paramenter fine-tuning \cite{CSTICA_2018}, following cross-validation arguments. Even the most recent constrained ICA techique, referred as CSTICA \cite{CSTICA_2018}, involves three regularization parameters and, as it is pointed out by the authors, that the algorithm still needs further improvement to ``enable these parameters not to be determined by the experiments''. Beyond the constrained ICA, there are also NMF algorithms that allow the incorporation of external information \cite{S-NMF, T-NMF}; yet, they suffer from similar drawbacks that limit their applicability in practice.

Recently, a DL method called \emph{Supervised Dictionary Learning} (SDL) \cite{SDL} was introduced, which allows the incorporation of external information from the task-related time courses in a rationale similar to to GLM. As a result, SDL is greatly benefited in the case of highly overlapping spatial maps,  and leads to performance similar to that of GLM. However, SDL inherits from GLM two major drawbacks: a) it builds upon the cHRF, which is fixed and inevitably different from the true one, and b) it adopts a regularized formulation of the DL, which inherits the difficulties associated with  the tuning of the related parameter. In \Sec{\ref{Sub:Exp_AllComp}}, \Tab{II}, we provide a thorough comparison among all competitive approaches and their characteristics, which are of interest in the fMRI case.

In this paper, a novel DL formulation, referred to as Information Assisted Dictionary Learning (IADL) is proposed, which, among other merits, alleviates the two aforementioned critical disadvantages of SDL as well as those related to constrained ICA approaches. More specifically: 

\begin{itemize}
	\item A new semi-blind DL approach is proposed that incorporates HRF-related information. However, this information is provided in a {\it relaxed} way that allows flexibility around the cHRF and, implicitly, it can accommodate related inaccuracies and variations. 

	\item A new sparsity constraint is adopted, which incorporates sparsity in a systematic way that accepts a physical interpretation that relates to the relative size of the areas, which are activated by the underlying FBNs. This latter information is readily available in brain atlases and it naturally complies with the FBNs' segregated nature.
	
	\item The proposed sparsity constraint also offers the flexibility of dealing with dense sources that is, sources that are not sparse. Indeed, in real fMRI, some of the sources, which are usually related to machine artifacts, can be dense.
	
	\item A highly realistic synthetic dataset is constructed allowing to conduct a thorough performance evaluation of the new method against state-of-the-art ICA and DL-based approaches.

\end{itemize}

\subsubsection*{Notation}
A lower case letter, $x$, denotes scalars, a bold capital letter, $\mathbf{X}$, denotes a matrix, and a bold lower case letter, $\mathbf{x}$, denotes a vector with its $i^{\mathrm{th}}$ component denoted as $x_{i}$. The $i^{\mathrm{th}}$ row and the $i^{\mathrm{th}}$ column of a matrix, $\mathbf{X}\in\mathbb{R}^{M\times N}$, are represented as $\mathbf{x}^{i}\in\mathbb{R}^{1\times N}$ and $\mathbf{x}_{i}\in\mathbb{R}^{M\times 1}$, respectively. Moreover, $x_{ij}$ denotes the element ``located'' at row $i$ and column $j$ of the matrix $\mathbf{X}$.

\section{Novel DL constraints tailored to task-related fMRI}
\label{Sec:Statment}

\subsection{Preliminaries on DL-based fMRI analysis}
The data collected during an fMRI experiment form a two dimensional data matrix as follows: Each one of the acquired 3D images is unfolded and stored into a vector, $\mathbf{x}=\left[x_{1},x_{2},\ldots,x_{N}\right]\in\mathbb{R}^{1\times N}$, where $N$ is the total number of voxels per image. Such vectors correspond to the sequence of, say $T$, successively obtained images and they are concatenated as rows to form the data matrix $\mathbf{X}\in\mathbb{R}^{T\times N}$. Note that in practice, prior to the formation of the data matrix,  $\mathbf{X}$, several standardized pre-processing steps are conducted to account for many detrimental effects related to the fMRI image acquisition process, such as slice timing correction, head motion, realignment, normalization, etc.

From a mathematical point of view, the source separation problem can be described as a matrix factorization task of the data matrix, i.e.,
\begin{equation}
	\mathbf{X}\approx\mathbf{D}\mathbf{S}\coma
\end{equation}
where, following the dictionary learning jargon, $\mathbf{D}\in\mathbb{R}^{T\times K}$ is the dictionary matrix, whose columns represents different time courses, $\mathbf{S}\in\mathbb{R}^{K\times N}$ is the coefficient matrix, whose rows are the spatial maps associated with the corresponding time courses, and $K$ is the number of sources. In general, such a matrix factorization can be obtained via the solution of a constrained optimization task:
\begin{equation}
	(\hat{\mathbf{D}},\hat{\mathbf{S}})=\underset{\mathbf{D},\mathbf{S}}{\text{argmin }}\left\Vert \mathbf{X}-\mathbf{D}\mathbf{S}\right\Vert _{F}^{2}\;\text{ s.t. }\;\begin{array}{c}
\mathbf{D}\in\mathfrak{D}\\
\mathbf{S}\in\mathfrak{L}
\end{array}
\coma
\label{Eq:Main}
\end{equation}
where $\mathfrak{D}$, $\mathfrak{L}$, are two sets of admissible matrices, defined by an adopted set of appropriately imposed constraints  and $\Vert\cdot\Vert_F$ denotes the Frobenious norm of a matrix.

The concept of sparsity refers to discrete signals that involve a sufficiently large number of zero values. The typical way to quantify sparsity is via the $\ell_{0}$-norm: given an arbitrary vector, $\mathbf{x}\in\mathbb{R}^{N}$, the $\ell_{0}$-norm (which, strictly speaking, is not a norm its strict mathematical definition~\cite{Sergios}) is defined as the number of non-zero components of a vector. In DL, the coefficient matrix is constrained to be sparse. In standard DL methods, the sparsity constraints are usually implemented in two ways: either each column of $\mathbf{S}$ is separately constrained to be sparse, e.g., $\left\Vert \mathbf{s}_{i}\right\Vert _{0}\leqslant\gamma_{i}$, where $\gamma_{i}$ is the maximum number of the non-zero values of the $i^{\mathrm{th}}$ column of $\mathbf{S}$, or the full coefficient matrix, $\mathbf{S}$, is constrained to involve, at most, $\hat{\gamma}$ non-zeros \cite{Yannis-DL,SDL,Davies}.

The new proposed method introduces novel constraints on the spatial maps (i.e., on the coefficient matrix, $\mathbf{S}$) and time courses (i.e., on the dictionary, $\mathbf{D}$), whose design serves the specific needs of the task-related fMRI data analysis. These constraints are discussed next.

\subsection{Information bearing sparsity constraints on the spatial maps}
\label{Sub:Stat_wL1}
In the fMRI framework, sparsity seems a natural assumption to model the segregated nature of the spatial maps of the FBNs. In other words, each row, say $\mathbf{s}^{i}$, of the coefficient matrix, $\mathbf{S}$, should have non-zeros values only at entries that correspond to voxels activated by the corresponding time course $\mathbf{d}_{i}$. The smaller the area, which a specific FBN occupies in the brain, the sparser the corresponding row of the coefficient matrix should be. At this point, it is worth to recall that, so far, all the DL methods applied in fMRI do not impose sparsity row-wise. However, imposing sparsity column-wise assumes that only a few sources are active in each voxel. Although this is generally true, this piece of information is voxel-dependent and it is hard to accommodate a suitable regularization parameter that simultaneously work optimally for all the thousands of columns. On the other hand, imposing sparsity in the full coefficient matrix may be easier to handle, yet, it is a general constraint that fails to exploit relevant information regarding the spread of each FBN.

In this paper, to the best of our knowledge, it is the first time that the DL framework is extended in order to allow sparsity promotion along the rows of the coefficient matrix. Looking  at \Eq{\ref{Eq:Main}}, sparsity in the rows of the coefficient matrix can be imposed using the following admissible set of constraint: 
\begin{equation}
	\mathfrak{L}_{0}=\left\{ \mathbf{S}\in\mathbb{R}^{K\times N}\;|\;\left\Vert \mathbf{s}^{i}\right\Vert _{0}\leqslant\phi_{i}\;,\;\forall ~i=1,2,\ldots,K\right\} 
	\coma\label{Eq:L0}
\end{equation}
where $\phi_{i}$ is a user-defined parameter, which denotes the maximum number of non-zero elements of the $i^{\mathrm{th}}$ row of $\mathbf{S}$. In the fMRI case, the parameter $\phi_{i}$ bears a clear physical interpretation and it corresponds to the total number of voxels that are active due to the $i^{\mathrm{th}}$ source. This corresponds to the number of voxels that the corresponding FBN occupies, an estimate of which can be obtained from brain atlases.

It is well known that the $\ell_{0}$-norm constraint results in an NP-hard optimization, and it is usually relaxed to its closer convex relative, i.e., the $\ell_1$-norm \cite{Equiv-Cond}. Then, the corresponding constrained set becomes:
\begin{equation}
	\mathfrak{L}_{1}=\left\{ \mathbf{S}\in\mathbb{R}^{K\times N}\,|\,\left\Vert \mathbf{s}^{i}\right\Vert _{1}\leqslant\lambda_{i}\;,\;\forall ~ i=1,2,\ldots,K\right\} 
	\coma\label{Eq:L1}
\end{equation}
where, $\lambda_{i}$, are \emph{new} user-defined parameters to implicitly control the sparsity of the rows of the coefficient matrix. In contrast to the parameters, $\phi_{i}$, the new parameters, $\lambda_{i}$, are not directly related to the sparsity level, rendering them hard to tune in practice, unless cross-validation is an option, where in fMRI this is not the case. 

An additional novelty of IADL that allow us to bypass the previous drawback, is the application across rows of a weighted version of the $\ell_{1}$-norm. In particular, given an arbitrary vector $\mathbf{x}\in\mathbb{R}^{N}$, the weighted $\ell_{1}$-norm is defined as:
\begin{equation}
	\left\Vert \mathbf{x}\right\Vert _{1,\mathbf{w}}=\sum_{i=1}^{N}w_{i}\left|x_{i}\right|
	\coma\label{Eq:wln}
\end{equation}
where $\mathbf{w}$ is a real positive vector given by
\begin{equation}
	w_{i}=\frac{1}{\left|x_{i}\right|+\varepsilon}\qquad\forall i=1,2,\ldots,N
	\coma\label{Eq:wdef}
\end{equation}
and $\varepsilon\in\mathbb{R}^{+}$ is a real positive number, which is introduced in order to avoid division by zero, providing enhanced numerical stability, \cite{wl1-norm,Yan-wl1-norm,Sergios}, and can be set directly to a small value, e.g. $10^{-6}$. Accordingly, the row-sparsity constraint now becomes:
\begin{equation}
	\mathfrak{L}_{w}=\left\{ \mathbf{S}\in\mathbb{R}^{K\times N}\;|\;\left\Vert \mathbf{s}^{i}\right\Vert _{1,\mathbf{w}^{i}}\leqslant\phi_{i}\quad\forall i=1,2,\ldots,K\right\} 
	\coma
	\label{Eq:Lw}
\end{equation}
where $\mathbf{w}^{i}$ is the vector of weights that correspond to the vector $\mathbf{s}^i$ computed as in \Eq{\ref{Eq:wdef}}. The key point  is that, similarly to the $\ell_{0}$ case, the constraint is imposed via the sparsity level $\phi_i$ and, consequently, $\phi_{i}$ can now be used as an upper bound. This is theoretically substantiated, that the weighted $\ell_{1}$-norm is bounded by the corresponding $\ell_{0}$-norm~\cite{Yan-wl1-norm}. As it will be further discussed in \Sec{\ref{Sec:IADL}}, this facilitates the transferring of knowledge that relates to the brain structure and functional connectivity into the optimizations task. Moreover, as it shown in \App{\ref{Sub:Math_Convex}}, given $\mathbf{w}$, $\mathfrak{L}_{w}$ remains \emph{convex}.

Hereafter, an equivalent but conceptually easier to handle sparsity measure that is independent of the length of the vector, known as sparsity percentage, will be used interchangeably with the sparsity level. Sparsity percentage expresses the proportion of zeros within a vector, $\mathbf{x}$ and it is given by
\begin{equation}
	\theta=\left(1-\frac{\phi}{N}\right)\times100
	\coma
\end{equation}
where $\phi=\left\Vert \mathbf{x}\right\Vert _{0}$, and $N$ is the total number of elements. 

\subsection{Task-related dictionary soft constraints}

The enhanced discriminative power of the GLM over the BSS methods comes from the fact that in GLM, the task related time courses are explicitly provided into the GLM modeling via the estimated BOLD sequences \cite{Handbook}. Such information is left unexploited in the BSS framework. Hence, it seems reasonable to incorporate this information in the BSS methods, leading naturally to a semi-blind formulation.

As it has already been stated before in the Introduction,  in contrast to ICA techniques, DL-based methods can easily incorporate, in principle, any constraint in the time courses, since sparsity is not affected by such assumptions. This fact has been exploited in SDL by splitting the dictionary into two parts:
\begin{equation}
	\mathbf{D}=\left[\mathbf{\Delta},\mathbf{D}_{F}\right]\in\mathbb{R}^{T\times K}
	\coma
\end{equation}
where the first part, $\mathbf{\Delta}\in\mathbb{R}^{T\times M}$, comprises fixed columns, which are set equal to the imposed task-related time courses, $\bdelta_{1},\bdelta_{2},\ldots,\bdelta_{M}$. The second part, $\mathbf{D}_{F}\in\mathbb{R}^{T\times (K-M)}$, is left to vary and it is learned during the DL optimization phase. Nevertheless, SDL inherits the same drawback associated with the GLM: the constrained atoms of the fixed dictionary (columns of the matrix $\mathbf{\Delta}$) lead to improvement only if the imposed task-related time courses are sufficiently accurate. Otherwise, the task-related time courses will be mis-modelled and their contribution can introduce detrimental effects, leading to innaccurate results.

In this paper, we relax the strong \emph{equality} requirement of SDL to a looser \emph{similarity}-based distance-measuring norm constraint. Then, if part of the a priori information is inaccurate, e.g., the assumed HRF differs from the true one, the method can efficiently adjust the constrained atoms since they are not forced to remain fixed and equal to the preselected time-courses. Moreover, the proposed modeling accounts, also, for multiple factors that potentially alter the functional shape of the task-related time courses across subjects and brain regions, such as, vascular differences, partial volume imaging, brain activations, \cite{HRF-Var-2}, hematocrit concentrations \cite{HRF-Hematocrit}, lipid ingestion \cite{HRF-Fat} and even nonlinear effects due to short interstimulus intervals \cite{NonLinear-HRF}.

Mathematically, the starting point lies on splitting the dictionary into two parts:
\begin{equation}
\label{DicSplit}
	\mathbf{D}=\left[\mathbf{D}_{C},\mathbf{D}_{F}\right]\in\mathbb{R}^{T\times K}
	\coma
\end{equation}
where, in contrast to SDL, the part, $\mathbf{D}_{C}\in\mathbb{R}^{T\times M}$, has columns which are constrained to be \emph{similar} rather than equal to the imposed task-related time courses. Then, we can define a new \emph{convex} set of admissible dictionaries:
\begin{equation}
	\mathfrak{D}_{\delta}=\left\{ \mathbf{D}\in\mathbb{R}^{T\times K}\;\left|\begin{array}{ll}
\left\Vert \mathbf{d}_{i}-\bdelta_{i}\right\Vert _{2}^{2}\leqslant c_{\delta} & i\in [1,M]\\
\left\Vert \mathbf{d}_{i}\right\Vert _{2}^{2}\leqslant c_{d} & i\in [M+1,K]
\end{array}\right.\right\} 
\coma
\label{Eq:Dd}
\end{equation}
where $\left\Vert \,\cdot\,\right\Vert _{2}$ denotes the Euclidean norm, $\mathbf{d}_{i}$ is the $i^{\mathrm{th}}$ column of the dictionary $\mathbf{D}$ and $\bdelta_{i}$ is the $i^{\mathrm{th}}$ a priori selected task-related time course. The constant $c_{\delta}$ is a user-defined parameter, which controls the \emph{degree of similarity} between the constrained atoms and the imposed time courses; essentially, this reflects our confidence on how accurate is the cHRF for the subject under consideration. In particular, as it is further explained in \App{\ref{Sub:Par_cdelta}}, $c_{\delta}$ accounts for the natural variability that the HRFs among subjects are expected to have giving rise to consistent strategies for its tuning. Moreover, the free atoms have a bounded norm controlled by $c_{d}$, another user-defined parameter to avoid ill conditioned phenomena, which can be fixed and set equal to~1~\cite{Allen}.

\section{The IADL algorithm}
\label{Sec:IADL}

In this section, we present an implementation of IADL that solves \Eq{\ref{Eq:Main}}, incorporating the two proposed sets of constraints, i.e. $\mathfrak{D}=\mathfrak{D}_{\delta}$ and $\mathfrak{L}=\mathfrak{L}_{w}$. 

The simultaneous minimization for $\mathbf{D}$ and $\mathbf{S}$ is challenging due to both the non-convexity of task in \Eq{\ref{Eq:Main}}, as well as the potentially large size of the data matrix. The size of the latter is restrictive for optimisation frameworks, which are computationally demanding. For the above reasons, we adopt the \emph{Block Majorized Minimization} (BMM) rationale, which provides a powerful framework for the solution of such type of optimization tasks, e.g. \cite{Gen-MM}. Essentially, the BMM scheme simplifies the optimization task by adopting a two-step alternating minimization, under certain assumptions \cite{Conv-MM}. The proposed DL approach is algorithmically described next and the full mathematical derivation and convergence proof are analytically provided in \App{\ref{Sec:Mathematical}}.

Put succinctly, the proposed DL algorithm follows the standard scheme of classical DL methods, which iteratively alternate between a sparse coding step and a dictionary update step. Concerning the sparse cording step, the corresponding recovery mechanism is a soft thresholding operator similar to the one that corresponds to the standard $\ell_{1}$-norm constraint as it is explained in \cite{Yan-wl1-norm}.

In the Algorithm 1, we present the pseudo-code for solving \Eq{\ref{Eq:Main}}, given the number of sources, $K$, an arbitrary set of estimates, $\mathbf{S}^{[0]},\mathbf{D}^{[0]}$, $M$ estimates of the task-related time courses, $\bdelta_{1},\bdelta_{2},\ldots,\bdelta_{M}$, grouped by columns in the matrix $\mathbf{\Delta}$, the sparsity for each row, $\bphi=[\phi_{1},\phi_{2},\ldots,\phi_{K}]^{T}$, and the number of iterations, $Iter$. Observe that the free parameters of the algorithm that need to be tuned are: a) the number of sources, $K$, b) the maximum sparsity per row $\phi_i$, c) the radius parameter $c_{\delta}$ and d) the parameters $c_d$ and $\varepsilon$ involved in equations \Eq{\ref{Eq:Dd}} and \Eq{\ref{Eq:wdef}}, respectively. The last two can be directly fixed to 1 and a small value, e.g. $10^{-6}$, respectively.  All the rest of the parameters can be straightforwardly tuned based on physical arguments, which can be easily drawn from the fMRI study at hand. The algorithm is insensitive in overestimating $K$, which renders this parameter easily tunable (see discussion in \App{\ref{Sub:Par_Sources}} and \App{\ref{Sub:Extra_EvalK}}. Also, the maximum sparsity per row is readily obtained from published brain atlases (see \App{\ref{Sub:Par_Sparsity}} and \App{\ref{Sec:Atlas}}) and $c_{\delta}$ is easily tuned considering the task related time courses (see \App{\ref{Sub:Par_cdelta}}); the latter results from established HRF models and their expected variability. Furthermore, we provide a Matlab implementation for IADL based analysis, which is available in the IADL\footnote{IADL repository: \url{https://github.com/MorCTI/IADL}\label{fnt:AIDL}} github repository. The implementation provided comprises automatic initialization and parameter selt-tuning for $c_{\delta}$. Accordingly, its default parameter setup can be adopted out-of-the-box with any task-related fMRI dataset. At the same time, a more thoughtful specification of the parameters, e.g. incorporating the expected sparsity level of the task-related FBNs at hand, can improve results further.

\begin{algorithm}[H]
\SetKwInOut{Input}{input}\SetKwInOut{Output}{output}
\Input{$K$, $\mathbf{S}^{[0]}$, $\mathbf{D}^{[0]}$, $\mathbf{\Delta}$, $\bphi$, $Iter$}
$\mathbf{B}=\mathbf{D}^{[0]}$\;
\For{$t=0$ \KwTo $Iter$}{
	$c_{S}$ equal (or bigger) than $\left\Vert \mathbf{B}^{T}\mathbf{B}\right\Vert $\;	
	$\mathbf{A}=\frac{1}{c_{S}}\left[\mathbf{B}^{T}\mathbf{X}+\left(c_{S}\mathbf{I}_{K}-\mathbf{B}^{T}\mathbf{B}\right)\mathbf{S}^{[t]}\right]$\;
	$\mathbf{W}\leftarrow w_{ij}=\frac{1}{\left|a_{ij}\right|+\varepsilon}$ with ${\scriptstyle(\varepsilon >0)}$\;
	\For{$i=1$ \KwTo $K$}{
		\If{$\left\Vert \mathbf{a}^{i}\right\Vert _{1,\mathbf{w}^{i}}>\phi_{i}$}{
			$\mathbf{a}^{i}=\mathcal{P}_{B_{\ell_{i}}[\mathbf{w}^{i},\phi_{i}]}(\mathbf{a}^{i})$\;	
		}
	}	
	$\mathbf{S}^{[t+1]}=\mathbf{A}$\;
	\vspace{3mm}
	$c_{D}$ equal (or bigger) than $\left\Vert \mathbf{A}\mathbf{A}^{T}\right\Vert $\;
	$\mathbf{B}=\frac{1}{c_{D}}\left[\mathbf{X}\mathbf{A}^{T}+\mathbf{D}^{[t]}\left(c_{D}\mathbf{I}_{K}-\mathbf{A}\mathbf{A}^{T}\right)\right]$\;
	\For{$i=1$ \KwTo $M$}{
		\If{$\left\Vert \mathbf{b}_{i}-\bdelta_{i}\right\Vert ^{2}> c_{\delta}$}{$\mathbf{b}_{i}=\frac{c_{\delta}^{\nicefrac{1}{2}}(\mathbf{b}_{i}-\bdelta_{i})}{\left\Vert \mathbf{b}_{i}-\bdelta_{i}\right\Vert }+\bdelta_{i}$\;
		}		
	}
	\For{$i=M+1$ \KwTo $K$}{
		\If{$\left\Vert \mathbf{b}_{i}\right\Vert ^{2}>c_{d}$}{
				$\mathbf{b}_{i}=\frac{c_{d}^{\nicefrac{1}{2}}}{\left\Vert \mathbf{b}_{i}\right\Vert }\mathbf{b}_{i}$\;		
		}
	}	
	$\mathbf{D}^{[t+1]}=\mathbf{B}$\;
}
\Output{$\mathbf{D}=\mathbf{D}^{[t+1]}$, $\mathbf{S}=\mathbf{S}^{[t+1]}$}
\caption{Information Assisted DL}
\label{Alg:Main}

{\footnotesize{where $\mathcal{P}_{B_{\ell_{1}}[\mathbf{w}^{i},\phi_{i}]}$ is the projection operator over the weighted $\ell_{1}$-norm ball, $B_{\ell_{1}}[\mathbf{w}^{i},\phi_{i}]=\{ \mathbf{x}\in\mathbb{R}^{N}\;|\;\left\Vert \mathbf{x}\right\Vert _{1,\mathbf{w}^{i}}\leqslant\phi_{i}\} $, of weights $\mathbf{w}^{i}$ and radius $\phi_{i}$. This projection operator onto the weighted $\ell_{1}$-norm ball is derived in closed form in \cite{Yan-wl1-norm}.}}
\end{algorithm}

\section{Experimental Results}
\label{Sec:Experiments}

\subsection{Performance results based on synthetic data}
\label{Sub:Exp_Synthetic}
In \App{\ref{Sec:SyntheticDataset}}, a novel synthetic data set is presented. This set is highly realistic that emulates demanding experimental tasks, where substantial spatial map overlap exists. 
This allows us to effectively evaluate the performance of the proposed DL method in comparison with the state-of-the-art of blind and semi-blind approaches under a more realistic  setting. The brain-like sources comprising the dataset are, for convenience, depicted in \Fig{\ref{Fig:Srcs}} in \App{\ref{Sec:SyntheticDataset}}.

\begin{table*}[t]
\begin{center}
\caption{Sparsity parameters for IADL}
\label{Tab:Lvls}
\begin{tabular}{cc}
\multicolumn{2}{c}{\textbf{(a)}}\tabularnewline
\multicolumn{2}{c}{\textbf{True Sparsity}}\tabularnewline
\hline 
N$^{\circ}$ & Sparsity (\%)\tabularnewline
\hline 
\hline 
1 & 95.28 \tabularnewline
11 & 91.60 \tabularnewline
14 & 94.57 \tabularnewline
\hline 
6 & 97.04 \tabularnewline
3 & 95.53 \tabularnewline
2 & 95.33 \tabularnewline
13 & 94.51 \tabularnewline
5 & 93.30 \tabularnewline
10 & 92.67 \tabularnewline
8 & 91.82 \tabularnewline
12 & 91.53 \tabularnewline
4 & 88.25 \tabularnewline
7 & 88.07 \tabularnewline
19 & 86.14 \tabularnewline
9 & 85.51 \tabularnewline
15 & 71.95 \tabularnewline
20 & 71.84 \tabularnewline
18 & 1.99 \tabularnewline
16 & 1.00 \tabularnewline
17 & 1.00 \tabularnewline
\hline
 & \tabularnewline
 & \tabularnewline
 & \tabularnewline
 & \tabularnewline
 & \tabularnewline
 
\end{tabular}\hspace{5mm}
\begin{tabular}{ccc}
\multicolumn{3}{c}{\textbf{(b)}}\tabularnewline
\multicolumn{3}{c}{\textbf{Sparsity (\%)}}\tabularnewline
\hline 
$\btheta_{1}$ & $\btheta_{2}$ & $\btheta_{3}$\tabularnewline
\hline 
\hline 
95 & 95 & 85\tabularnewline
90 & 90 & 85\tabularnewline
94 & 90 & 85\tabularnewline
\hline 
95 & 95 & 95\tabularnewline
94 & 90 & 90\tabularnewline
93 & 90 & 90\tabularnewline
92 & 90 & 90\tabularnewline
91 & 90 & 90\tabularnewline
90 & 90 & 90\tabularnewline
89 & 85 & 85\tabularnewline
88 & 85 & 85\tabularnewline
87 & 85 & 85\tabularnewline
86 & 85 & 85\tabularnewline
85 & 85 & 85\tabularnewline
80 & 80 & 80\tabularnewline
80 & 80 & 80\tabularnewline
75 & 75 & 75\tabularnewline
75 & 75 & 75\tabularnewline
70 & 70 & 70\tabularnewline
60 & 70 & 70\tabularnewline
10 & 10 & 10\tabularnewline
5 & 5 & 5\tabularnewline
0 & 0 & 0\tabularnewline
0 & 0 & 0\tabularnewline
0 & 0 & 0\tabularnewline
\hline
\end{tabular}\hspace{5mm}
\begin{tabular}{c}
\textbf{(c)} \tabularnewline
\textbf{Sp. (\%)}\tabularnewline
\hline 
$\btheta_{r}$\tabularnewline
\hline 
\hline 
90\tabularnewline
95\tabularnewline
95\tabularnewline
95\tabularnewline
95\tabularnewline
95\tabularnewline
\hline
95\tabularnewline
95\tabularnewline
90\tabularnewline
90\tabularnewline
90\tabularnewline
85\tabularnewline
80\tabularnewline
70\tabularnewline
50\tabularnewline
30\tabularnewline
20\tabularnewline
10\tabularnewline
0\tabularnewline
0\tabularnewline
\hline 
\tabularnewline
\tabularnewline
\tabularnewline
\tabularnewline
\tabularnewline
\end{tabular}
\end{center}
{\footnotesize Table (a) and (b) depict a comparison between three different choices of sparsity $\btheta_{1}$, $\btheta_{2}$ and $\btheta_{3}$ used in the synthetic experiments (b), and the sorted sparsity percentage values of the implemented realistic synthetic data set (a). The first three values in both cases correspond to the assisted sources of interest. Table (c) shows the values used for the real fMRI analysis.}
\end{table*}

The adopted performance measure, $r$, is associated with the Pearson's correlation coefficient among the estimated and the true sources and it is described in detail in \App{\ref{Sec:Measurements}}.

The aim of our first performance study is twofold: first, to study the effectiveness of the proposed approach in dealing with HRF mis-modeling and second to evaluate the algorithm's decomposition performance with respect to sets of sources of interest. As benchmarks, the following competitive algorithms are considered: a) McICA\footnote{Morphologically constrained ICA is a constrained ICA algorithm that subtract one particular signal of interest using given a priori information In this paper we implemented the open implementation from \url{http://dsp.ucsd.edu/~zhilin/Software.html}}, which is a constrained ICA algorithm~\cite{Morpho-ICA} that allows to assist a source using the task-related time courses, b) SDL, an Online DL algorithm (ODL) \cite{SPAMS}, which is included in SPAMS toolbox\footnote{The SPAMS (SPArse Modeling Software) is an open source optimization toolbox for solving various sparse estimation problems. \url{http://spams-devel.gforge.inria.fr}} and c) two ICA algorithms, namely, Infomax\footnote{Infomax is an algorithm for ICA using maximum likelihood. In this paper, we used an open source implementation of this algorithm from the DTU Toolbox: \url{http://cogsys.imm.dtu.dk/toolbox/menu.html}} \cite{Infomax}, a widely used ICA algorithm within the fMRI community and JADE\footnote{The  the Joint Approximate Diagonalization of Eigenmatrices (JADE) is an algorithm for ICA that separates observed mixed signals into latent source signals by exploiting fourth order moments. In this paper, we used the open source implementation from J.F. Cardoso: \url{http://mikexcohen.com/lecturelets/eigen/jader.m}} \cite{jadeR}, which we used as initialization point for all the involved DL algorithms.

\begin{figure}[H]
	\centering
	\includegraphics[width=0.95\columnwidth]{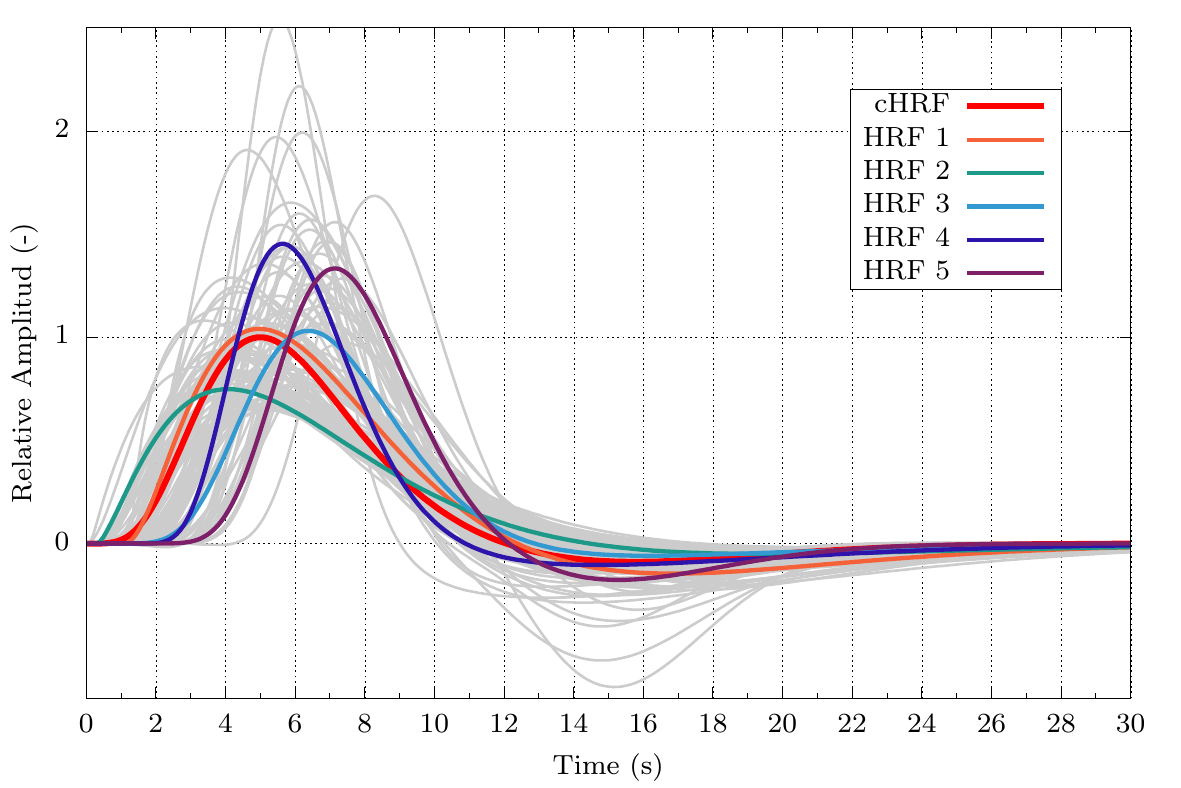}
	\caption{Graphic representation of 100 HRFs (gray) randomly generated from the two gamma distributions model. The red curve represents the canonical HRF (cHRF) and the rest of the colored HFRs stands for the five selected alternatives.}
	\label{Fig:HRFs}
\end{figure}

In order to emulate the HRF variability, we generated six different  ``subjects'', through six different, yet realistic, synthetic HRFs. These selected HRFs correspond to the HRFs that are depicted in the \Fig{\ref{Fig:HRFs}}. In particular, six different datasets have been built, one per each subject, with the only difference among them being the fact that the brain-induced time courses have been generated using the HRF of the corresponding subject. Moreover, the task-related time courses are chosen to be the sources 1, 11 and 14 (see \Fig{\ref{Fig:Srcs}}), since they correspond to realistic scenarios that are often encountered in practice:  Source 1 is easy to identify, since it barely spatially overlaps with other sources and it corresponds to a block-event experimental design. Sources 11 and 14 are more challenging and exhibit notable overlap and they emulate an event-related task (see \Fig{\ref{Fig:Srcs}}). Consequently, we generate the imposed task-related time courses $\bdelta_{1}$, $\bdelta_{2}$ and $\bdelta_{3}$, convolving the experimental conditions related to these sources with the cHRF, according to the standard procedure that is followed in GLM/SPM-based analysis.

\begin{figure*}[t]
	\centering	
	{\footnotesize\textbf{I. Performance with respect to the full sources}}
	\includegraphics[width=1.5\columnwidth]{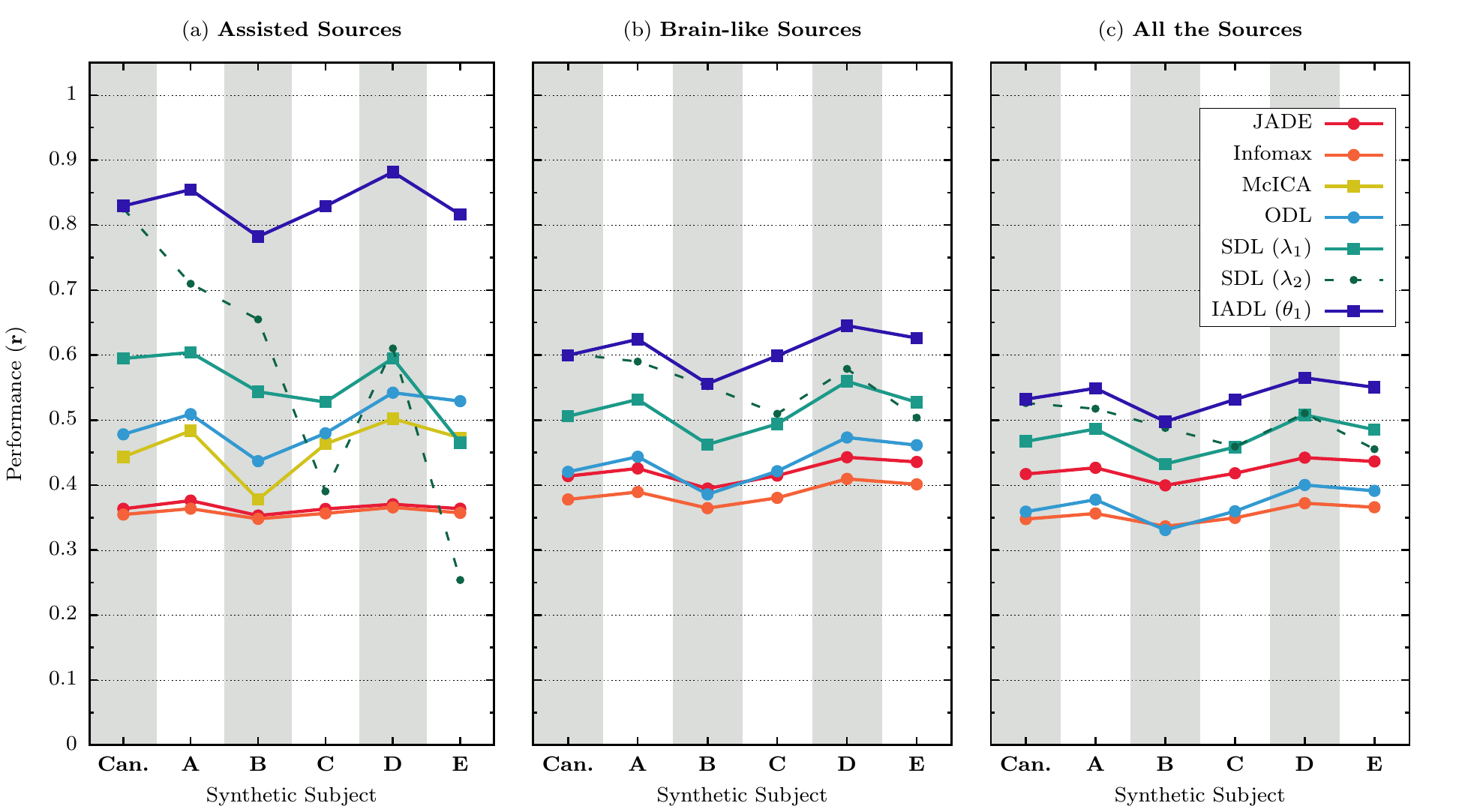}
	
	{\footnotesize\textbf{II. Performance with respect to the time courses}}
	\includegraphics[width=1.5\columnwidth]{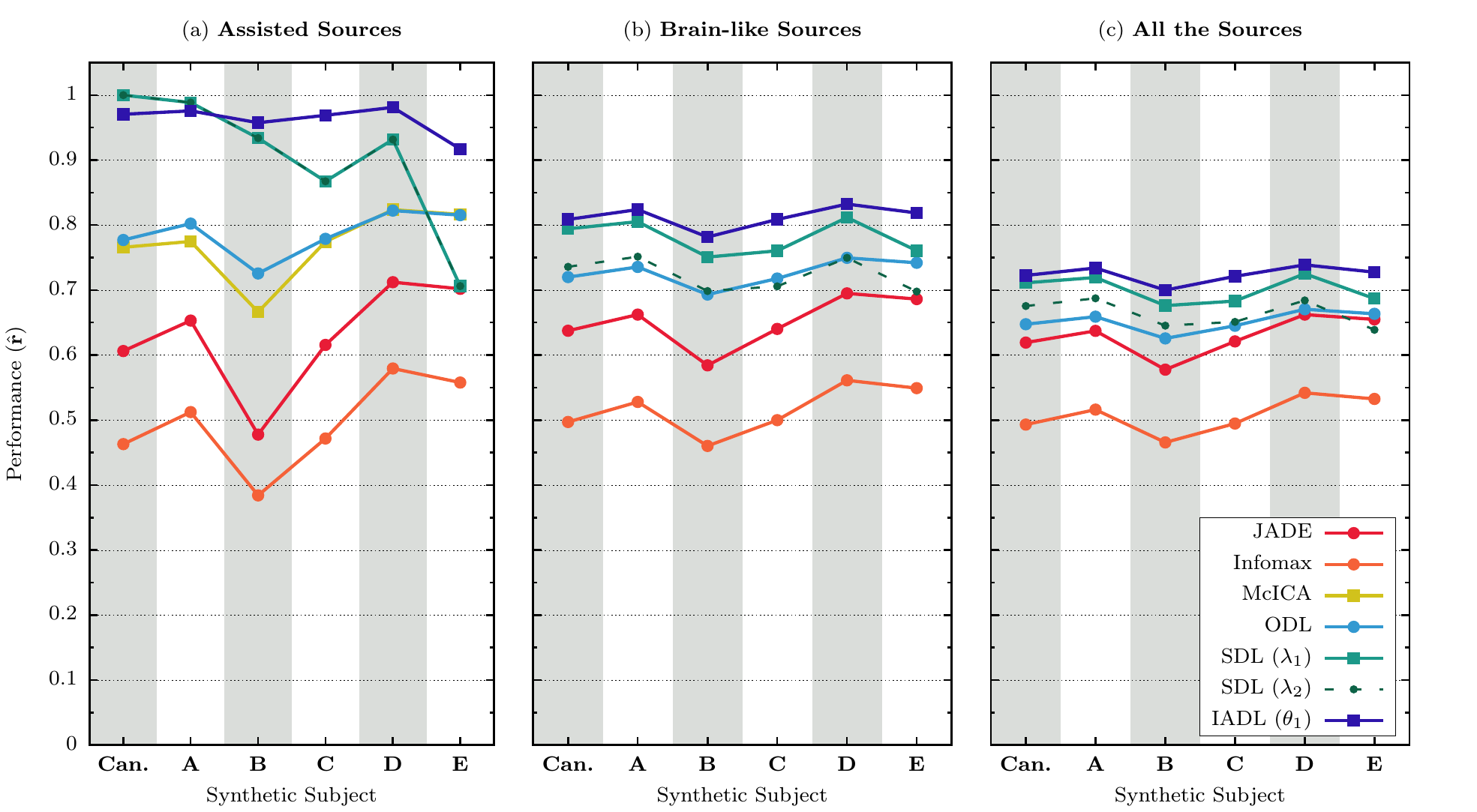}
	\caption{Performance comparison with respect to the full source (I) and the time courses only performance (II). The sub-figures correspond to a) the sources of interest [1, 11, 14], b) the brain-like sources only, and c) all sources (including artifacts)}
	\label{Fig:Exp_1}
\end{figure*}

Concerning the parametrisation of the algorithms, $K$ was overestimated by 20\%, i.e., $K=25$ rather than 20. This is typical in realistic scenarios since it is not possible to know the exact number of sources. All the benchmark methods require an estimate of the number of sources. Thus, the same value was provided to all the algorithms. Moreover, as it has been discussed in \Sec{\ref{Sec:IADL}}, IADL requires to set up three extra quantities: a) the maximum sparsity for each task-related sources, b) the set of maximum sparsity values that the rest of the sources can take and c) the parameter $c_{\delta}$. For (a), the imposed sparsity percentage per task-related sources 1, 11 and 14 are 95\%, 90\% and 94\% respectively, which correspond to a slight overestimation compared to the true sparsity values. Sparsity set-up information related to this experiment is also listed in \Tab{\ref{Tab:Lvls}}. In particular, the true sparsities of the task-related time courses are shown in the first three rows of \Tab{\ref{Tab:Lvls}.a} and the corresponding imposed sparsities are shown in the corresponding rows of the first column of \Tab{\ref{Tab:Lvls}.b} (sparsity set up scheme $\btheta_{1}$). For the point (b) above, the sparsity values, which are provided to the algorithm, are shown in the rest of the rows of $\btheta_{1}$. Moreover, to better grasp the relation among the imposed values and the true sparsity of non task-related sources, both the true and the imposed ones are sorted with decreasing sparsity percentage. In \Sec{\ref{Sub:Exp_Robustness}}, it will be shown that the proposed scheme is largely insensitive to the provided sparsity values. Finally, for the point (c) above, we set $c_{\delta}=0.2$, which is big enough to accommodate the variations among HRFs between the different synthetic datasets. This value has been calculated following the algorithmic procedure described in the \App{\ref{Sub:Par_cdelta}}.

SDL and ODL tune the sparsity constraint via a single regularisation parameter, $\lambda$. In contrast to IADL, $\lambda$ is not directly related to a certain sparsity level and its optimum value is case-specific. Moreover, it does not bear a concise physical interpretation that could serve as a guide for its tuning. Also, the range of values of the optimum $\lambda$ can vary significantly from case to case. It can take a very small value, e.g., 0.01 or a value that can be relatively large, e.g., 10. In the synthetic dataset case, since the ground truth, i.e., the correct decomposition, is fully known, $\lambda$ can get optimized through cross validation. However, such a luxury is never available in practice, where real fMRI datasets need to be analysed. In this study, we evaluate SDL using $\lambda$ values that lead to the best performance according to certain criteria: the first criterion was to optimize SDL with respect to the mean performance of the time courses of all the sources, leading to a $\lambda_{1}=0.02$. The second criterion was to optimize with respect the mean value of the full source performance for the assisted sources only, leading to a $\lambda_{2}=2.8$. In both cases, we conducted the $\lambda$ value optimization for the subject that corresponds to the cHRF. Observe the large variation between the obtained $\lambda$ values, which is indicative of the sensitivity of the SDL approach to $\lambda$ tuning. On the other hand, the optimum value for the fully blind ODL algorithm was found to be $\lambda=1.6$. Moreover, SDL and ODL are initialized from ICA, similarly to IADL.  Such an initialization leads to a better performance compared to the original version presented in \cite{SDL}. These observations agree with the results reported in \cite{Ini-ICA}, where the authors use ICA as initialization point for their proposed DL algorithm to enhance its performance.

McICA requires to fine-tune a set of 4 regularization parameters. We also observed that the optimal selection of these parameters heavily depends on the particular synthetic subject, similarly to the SDL algorithm. Accordingly, we manually optimized these parameters via cross-validation aiming to achieve the best average performance over all the subjects.


\Fig{\ref{Fig:Exp_1}.I} and \Fig{\ref{Fig:Exp_1}.II} show the performance results with respect to the full sources and the time courses, respectively. The horizontal axis indicates the six synthetic subjects that correspond to different HRFs. Both figures comprise three sub-graphs, where each one of them considers the performance with respect to three different sets of sources: (a) the task-related sources 1, 11 and 14, (b) the brain-like sources ($1,2,\ldots,15$), and (c) the whole dataset ($1,2,\ldots,20$) which includes both brain-like sources and artifacts. For completeness, 

\Fig{\ref{Fig:Ind_S}} in \App{\ref{Sec:Extras}} shows the individual performance of the studied methods for each assisted source and depicts some of the obtained time courses and its corresponding spatial maps.

Let us first focus on the two information-assisted DL algorithms, SDL and IADL, whose performance is indicated with green and dark blue curves, respectively. They are both assisted with the task-related time courses that correspond to the cHRF. In the SDL case, the solid and the dashed curves correspond to regularization parameter tuning equal to $\lambda_{1}$ and $\lambda_{2}$ respectively. Recall that these two cases lead to the best performance for the canonical subject in \Fig{\ref{Fig:Exp_1}.II.a} and \Fig{\ref{Fig:Exp_1}.I.c} respectively. In sub-figures \Fig{\ref{Fig:Exp_1}.I.a} and \Fig{\ref{Fig:Exp_1}.II.a}, it is already observed that IADL manages to cope well with the subject variability even in cases where the discrepancy between the canonical and the subject HRF is large, e.g., subject E (see \Fig{\ref{Fig:HRFs}}). On the contrary, in \Fig{\ref{Fig:Exp_1}.I.a}, SDL copes well only in the canonical case, for which the algorithm has been explicitly optimized. In \Fig{\ref{Fig:Exp_1}.II.a}, SDL succeeds a superior performance only in the canonical case --where the exact task-related time courses have been used-- as well as in subject A who, as it can be seen in the \Fig{\ref{Fig:HRFs}}, has an HRF similar to the canonical one. Focusing on the SDL performance for subject B to E, it is apparent that the strategy to fix the assisted time courses can lead to a high deterioration according to both performance measures.

In the cases where only brain-line and all the sources are considered (mid and right-most sub-figures), the proposed approach still outperforms SDL. Note that, the time courses estimated by IADL are overall better than that of SDL even in the case of the canonical subject, in which SDL is fully optimized exploiting the knowledge of the ground truth. In comparison to the fully blind methods, i.e., ODL, JADE and Infomax, task-related course assisting approaches perform clearly better. In this case, the ODL works better than ICA-based approaches in the cost; however, in practice, it is very hard, if possible at all, to optimize its $\lambda$ parameter as it is done here with the synthetic dataset.

The yellow curves in \Fig{\ref{Fig:Exp_1}} depict the performance of McICA. This particular constrained ICA algorithm provides estimates only of the assisted sources \cite{Morpho-ICA}; hence, the relates results correspond only to the assisted brain-like sources. First, we can observe that the McICA algorithm performs better than JADE and Infomax. On the other hand, our proposed IADL algorithm outperforms McICA for all the studied synthetic subjects.

\subsection{IADL robustness against sparsity parameter mistuning.}
\label{Sub:Exp_Robustness}

In this section, the tolerance of the proposed approach to the choice of the maximum sparsity parameters, $\phi_i$, is investigated. As it was apparent in the previous experiment with synthetic data, it is convenient to separately consider the sparsity constraints of the task-related time courses, which need to be explicitly set on an one-to-one basis. The sparsity constraints of the rest of the sources, only need a rough estimation of the sparsity percentage, as it is described in \App{\ref{Sub:Par_Sparsity}}. This convention is followed in \Tab{\ref{Tab:Lvls}.b}, where three such setups, denoted as $\btheta_{1}$, $\btheta_{2}$ and $\btheta_{2}$ are listed. The first one is the closest, overall, to the true sparsities. In $\btheta_{2}$, the sparsities of the non task-related sources, i.e., the 4th up to the 25th, have been grossly assigned in a simplified way; 5 sources with sparsity percentage 90\%, 5 sources with 80\%, etc. In $\btheta_{3}$, the sparsity percentage of the task-related time sources have been largely relaxed by fixing the sparsity percentage value to 85\% for all three of them (i.e., the 1st up to the 3rd). This figure roughly corresponds to the smaller sparsity percentage that can be found in all the FBNs and atlases that we have checked (a list of FBNs' sparsity percentages corresponding to several published brain atlas can be found in \App{\ref{Sec:Atlas}}.

\begin{figure*}[t]
	\centering
	
	{\footnotesize\textbf{I. Performance with respect to the full sources}}
	\includegraphics[width=1.5\columnwidth]{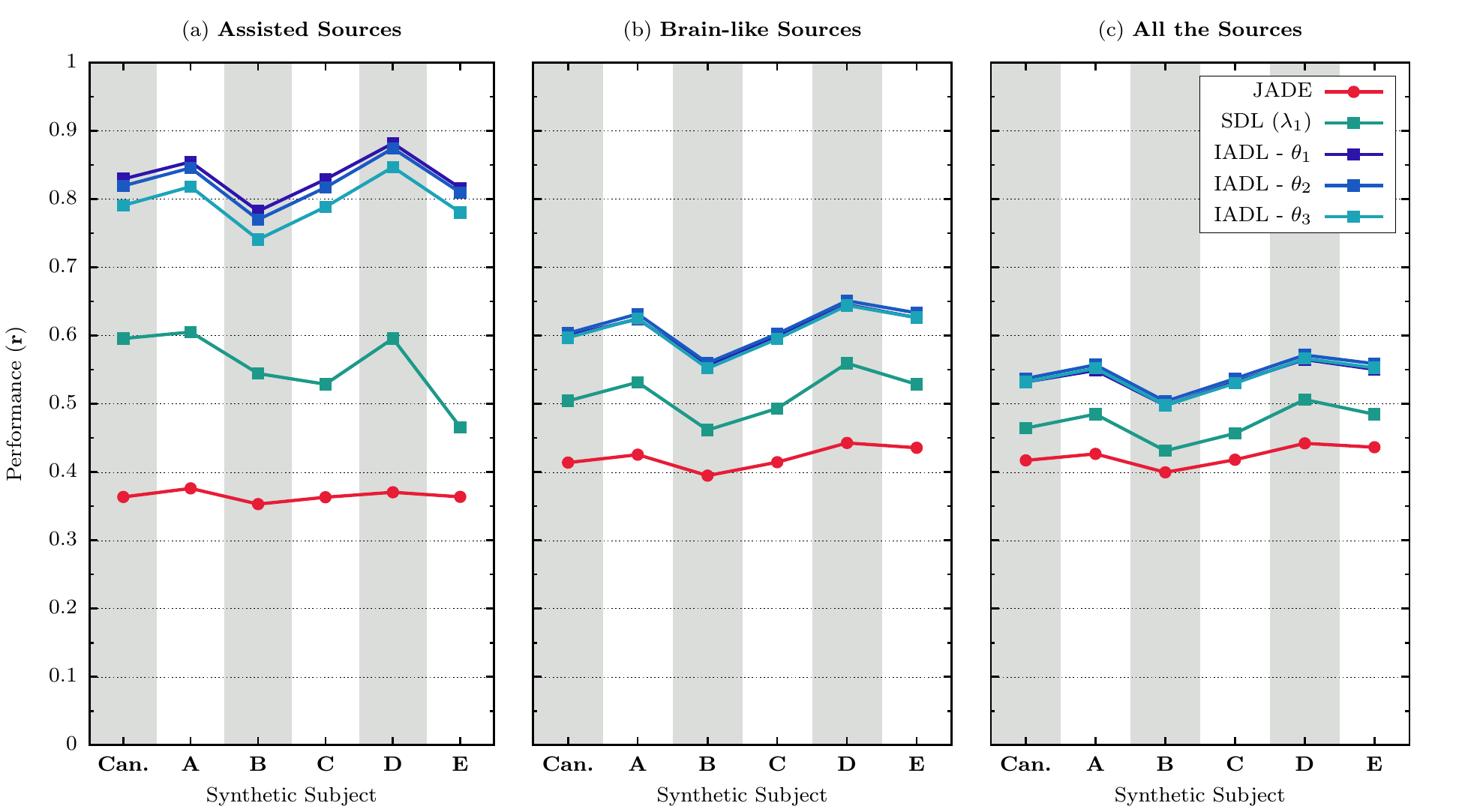}
	
	{\footnotesize\textbf{II. Performance with respect to the time courses}}
	\includegraphics[width=1.5\columnwidth]{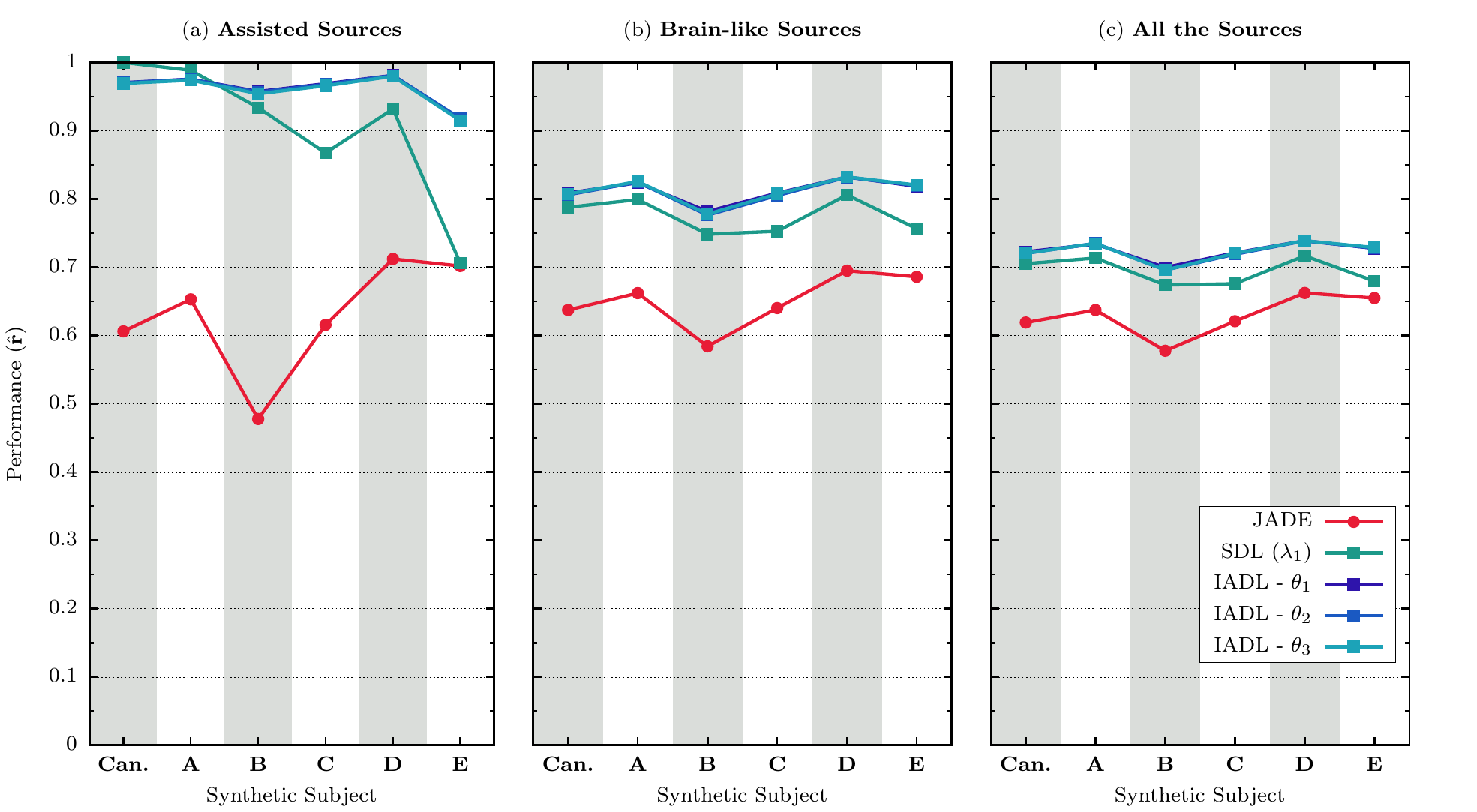}
	
	\caption{IADL performance using three different choices of sparsity. The results of SDL and JADE are also included as a reference.}
	
	\label{Fig:Exp_2}
\end{figure*}

\Fig{\ref{Fig:Exp_2}.I} and \Fig{\ref{Fig:Exp_2}.II} show the performance for these three sparsity setups. The first figure illustrates the performance results with respect to the full sources, whereas the second one shows the performance results with respect to the time courses only. Besides the sparsity levels, the set-up of the experiment is the same as the previous one; also and for reference, the performance curves of JADE and SDL are copied here. It can be readily observed that the proposed approach is remarkably robust to the sparsity specifications. Indeed, in the time-course only measure case, there are not any detrimental effects, whereas the full-source case, $\btheta_3$, led the estimates of the task-related sources to a minor performance degradation. This result is highly welcome and ensures that the requirement to explicitly set the sparsity for the task-related time courses is not an obstacle, when using the proposed algorithm. In fact, if there is extra information concerning the maximum expected sparsity level of the task-related time courses, then this can be used since it can only help. Otherwise, the sparsity percentage for the task-related time courses are all set to a safely small sparsity percentage value, e.g., 85\%, and this will still be beneficial to the algorithm leading to reliable results.

\subsection{Comparison between IADL and GLM}
\label{Sub:Exp_IADLvsGLM}

\begin{table*}[t]

\caption{Comparison between several alternatives for fMRI data analysis.}
\begin{center}
	\begin{tabular}{lccccccccc}
 & \multicolumn{9}{c}{\textbf{Criteria}}\tabularnewline
\textbf{Algorithm} & \textbf{A} & \textbf{B} & \textbf{C} & \textbf{D} & \textbf{E} & \textbf{F} & \textbf{G} & \textbf{H} & \textbf{I} \tabularnewline
\hline 
\hline 
IADL   & \cmark & \cmark & \cmark & \cmark & \cmark & \xmark & \cmark & \cmark & \cmark\tabularnewline
ICA    {\scriptsize \cite{ICA-Decade}, \cite{ICA-Novel}, \cite{ICA-Novel-2}}& \xmark & \xmark & \cmark & \cmark & \cmark  & \xmark & \xmark & \cmark & \cmark\tabularnewline
SPM    {\scriptsize\cite{SPM-web}} & \cmark & \xmark & \xmark & \cmark & \cmark & \xmark & \xmark & \cmark & \cmark\tabularnewline
McICA  {\scriptsize \cite{Morpho-ICA}} & \cmark & \cmark & \cmark & \xmark & \xmark & \xmark & \xmark & \cmark & \cmark \tabularnewline
CSTICA {\scriptsize \cite{CSTICA_2014}, \cite{CSTICA_2018}} & \cmark & \cmark & \cmark & \xmark & \xmark & \cmark & \xmark & \cmark & \xmark \tabularnewline
CTICA  {\scriptsize \cite{Semi-ICA-1}, \cite{Semi-ICA-Calhoun}} & \cmark & \cmark & \cmark & \xmark & \xmark & \xmark & \xmark & \cmark & \xmark \tabularnewline
ADL {\scriptsize \cite{ICASSP-Paper}} & \cmark & \cmark & \cmark & \xmark & \xmark & \xmark & \xmark & \xmark & \cmark \tabularnewline
CSICA  {\scriptsize \cite{CSICA_2010}}& \xmark & \xmark & \cmark & \xmark & \xmark & \cmark & \xmark & \cmark & \xmark \tabularnewline
T-NMF  {\scriptsize\cite{T-NMF}}  & \cmark & \cmark & \cmark & \xmark & \xmark & \xmark & \xmark & \xmark & \xmark\tabularnewline
ODL     {\scriptsize \cite{SPAMS}}  & \xmark & \xmark & \cmark & \xmark & \xmark & \xmark & \xmark & \xmark & \cmark\tabularnewline
SDL    {\scriptsize \cite{SDL}}& \cmark & \xmark & \cmark & \xmark & \xmark & \xmark & \xmark & \xmark & \xmark\tabularnewline
NMF    {\scriptsize \cite{NMF}}    & \xmark & \xmark & \cmark & \xmark & \xmark & \xmark & \xmark & \xmark & \cmark\tabularnewline
S-NMF  {\scriptsize \cite{S-NMF}} & \xmark & \xmark & \cmark & \xmark & \xmark & \cmark & \xmark & \xmark & \xmark\tabularnewline
Sparse-SPM  {\scriptsize \cite{Sparse-GLM}, \cite{Sparse-SPM} }& \xmark & \xmark & \cmark & \xmark & \xmark & \xmark & \xmark & \xmark & \xmark\tabularnewline
\hline 
	\end{tabular}
\end{center}

\textbf{Criteria:}\\
\textbf{A} -- Can impose task-related time information in a strict way.\\
\textbf{B} -- Can impose task-related time information in a relaxed/assisted way.\\
\textbf{C} -- Provides blind estimation of brain-induced sources and other unmodeled physiological noise or artifacts.\\
\textbf{D} -- Cross validation nor experimental visual inspection are not needed for parameter fine tuning. \\
\textbf{E} -- Free parameters are easy to tune and accept a physical interpretation within the fMRI context.\\
\textbf{F} -- Allows to include spatial information from the sources of interest through user-defined masks.\\
\textbf{G} -- Allows to include spatial information for all the sources in a relaxed/assisted way. \\
\textbf{H} -- Can explicitly deal with dense sources.\\
\textbf{I} -- An implementation of the code is freely available.

\label{Tab:CompTab}
\end{table*}

For completeness, we have also performed a comparison between the proposed IADL and the standard GLM approach using the SPM12\footnote{Statistical Parametric Mapping (SPM). Welcome Trust Centre for Neuroimaging, London, UK. \url{https://www.fil.ion.ucl.ac.uk/spm/software/spm12/}} toolbox, where the design matrix comprises the three task-related time courses. For this study, we observed that SPM and IADL recover the assisted sources 1 and 3 correctly. However, for the assisted source 2, the result of SPM is significantly inferior to that of IADL. Full details of this experiment can be found in \App{\ref{Sub:Extra_GLM}}.

\subsection{Comparison between IADL and several alternatives for the analysis of fMRI data}
\label{Sub:Exp_AllComp}

In order to clearly position IADL among all the matrix factorization-based methods that have been used in fMRI, with respect to all their features and analysis capabilities, we present a comprehensive comparison in Table II. Among those different alternatives, observe that IADL complies with all the studied criteria apart from criterion~\textbf{F}, namely to be able to explicitly specify the place/vexels within the brain that a FBN will appear through user-defined masks. However, note that IADL with relatively mild modifications in the spatial map constraint, $\mathfrak{L}_{w}$, can also comply with this criterion. Such development is beyond this work and it will be presented elsewhere.

\subsection{Real fMRI data analysis}
\label{Sub:Exp_Real-fMRI}
To test the capabilities of the proposed method over a realistic scenario, we study the motor task-fMRI experiment from the WU-Minn Human Connectome Project \cite{HCP}. This experiment follows a standard block paradigm, where a visual cue asks the participants to either tap their left/right fingers, squeeze their left/right toes or move their tongue. Each movement block lasted 12 seconds and is preceded by a 3 second of visual cue. In addition, there are 3 extra fixation blocks of 15 second each, as it is detailed in the Human Connectome Project protocols\footnote{Task-fMRI 3T Imaging Protocol Details: \url{http://protocols.humanconnectome.org/HCP/3T/task-fMRI-protocol-details.html}\label{fnt:protocols}}. 

The reason for selecting this specific dataset is twofold: first, the FBNs related to this experimental design are well studied~\cite{DL_2017}, \cite{Motor-ROIs}, \cite{Cortex-Yeo}, \cite{Cerebellum-Buckner}, which facilitates the evaluation of the results by inspection. Second, this dataset is particularly challenging: the FBNs of interest exhibit significant asymmetries in their intensity~\cite{Cortex-Yeo}, and the spatial maps exhibit high overlap, particularly within the cerebellar cortex~\cite{Cerebellum-Buckner}.

For this study, we selected the first 15 participants (10 females, range 26-35 years) that performed the studied motor task from the Q1 release Human Connectome Project repository\footnote{Human Connectome Project: \url{https://www.humanconnectome.org/}}, where the acquisition parameters are summarized in the imaging protocols\footnote{HCP 3T Imaging Protocol Overview: \url{http://protocols.humanconnectome.org/HCP/3T/imaging-protocols.html}}. On top of the standard preprocessing pipeline already conducted on the original dataset, \cite{Minimal}, \cite{Motor-ROIs}, we also smoothed each volume with a 4-mm FWHM Gaussian kernel. 

\begin{figure*}
	\centering
	\includegraphics[width=0.99\textwidth]{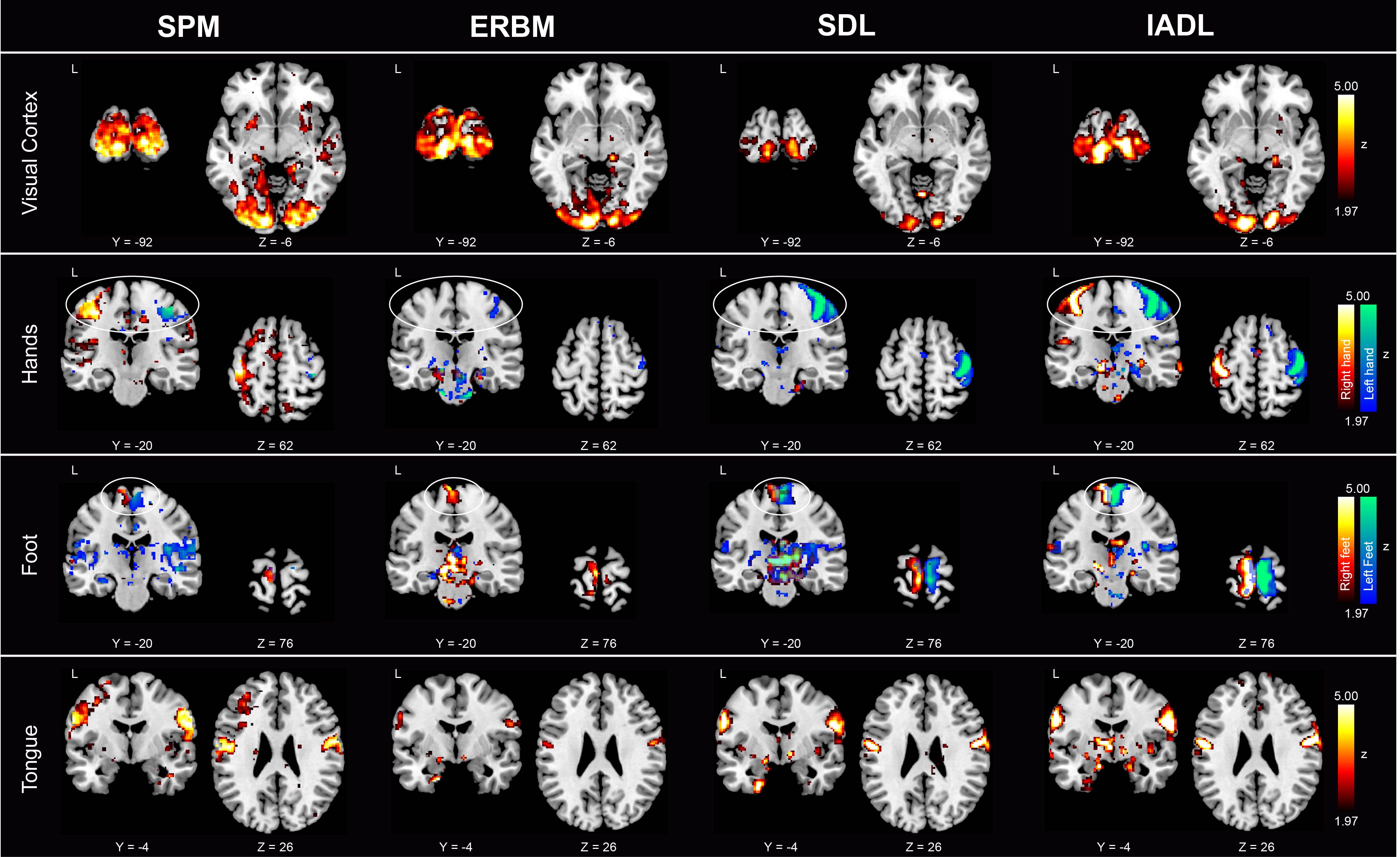}
	\caption{Significant active voxles ($z>1.97$) for the group analysis. Each row shows the most representative positions for each specific task: visual, left/right hand, left/right feet and tongue.}
	\label{Fig:Real}
\end{figure*}

For comparison, in addition to IADL, the analysis of the real data is also conducted with SDL, ICA and GLM. Thus, for this experiment we defined 6 task-related time courses, i.e., one per condition (visual cue, right-hand, left-hand, right-feet, left-feet and tongue), as a prior knowledge. Each condition was modeled as a succession of blocks with duration equal to its presentation time (see protocols\textsuperscript{\ref{fnt:protocols}}) and then convolved with cHRF. This is the standard procedure followed in GLM-based analysis as well \cite{Handbook}.

For the GLM analysis, we used SPM12${}^{6}$ and we followed the same standard procedure as it is described in \cite{Motor-ROIs}. For each subject, we computed a linear contrast to estimate significant activity for each movement type versus the baseline and versus all the other movements, and an extra linear contrast for the visual cue. Then, a second level analysis was performed to asses significant activity among the 15 participants, for each condition.

For the matrix factorization methods, in all the cases, the total number of sources was set equal to 20, which is a value often used in ICA analysis (for further details see discussion in \App{\ref{Sub:Par_Sources}}). Both semi-blind methods, IADL and SDL impose the same task-related time courses used in the SPM analysis, that is, each information-assisted algorithm will contain 6 different assisted sources, one per condition.

For ICA, we used the software toolbox GIFT\footnote{The Group ICA of fMRI Toolbox (GIFT) is a open Matlab software suitable for independent component analysis and blind source separation of group (and single subject) fMRI data: \url{http://mialab.mrn.org/software/gift/}.\label{ftn:GIFT}},  which implements multiple ICA algorithms in the context of fMRI data analysis. The algorithms Fast-ICA, Infomax, Erica and ERBM were tested and all of them produced qualitatively similar results. To save space, the results of the ERBM algorithm \cite{ERBM} are shown only, since by visual inspection it appeared to lead to somewhat better performance compared to the rest.

\begin{figure}[H]
	\centering
	\includegraphics[width=0.5\textwidth]{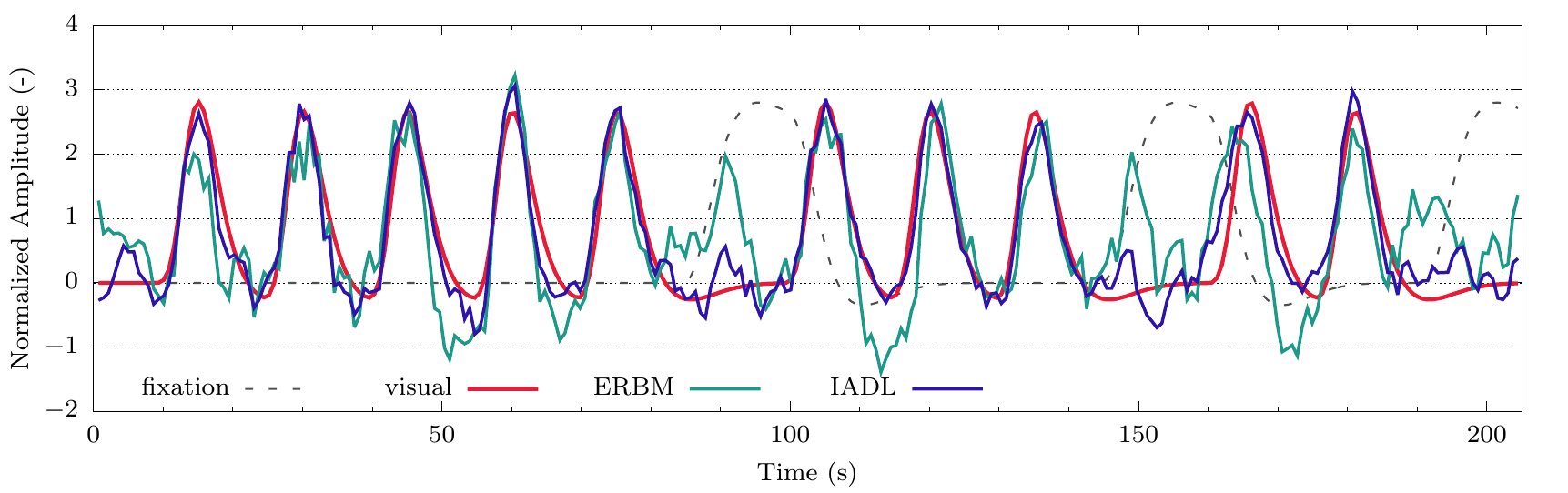}
	\includegraphics[width=0.5\textwidth]{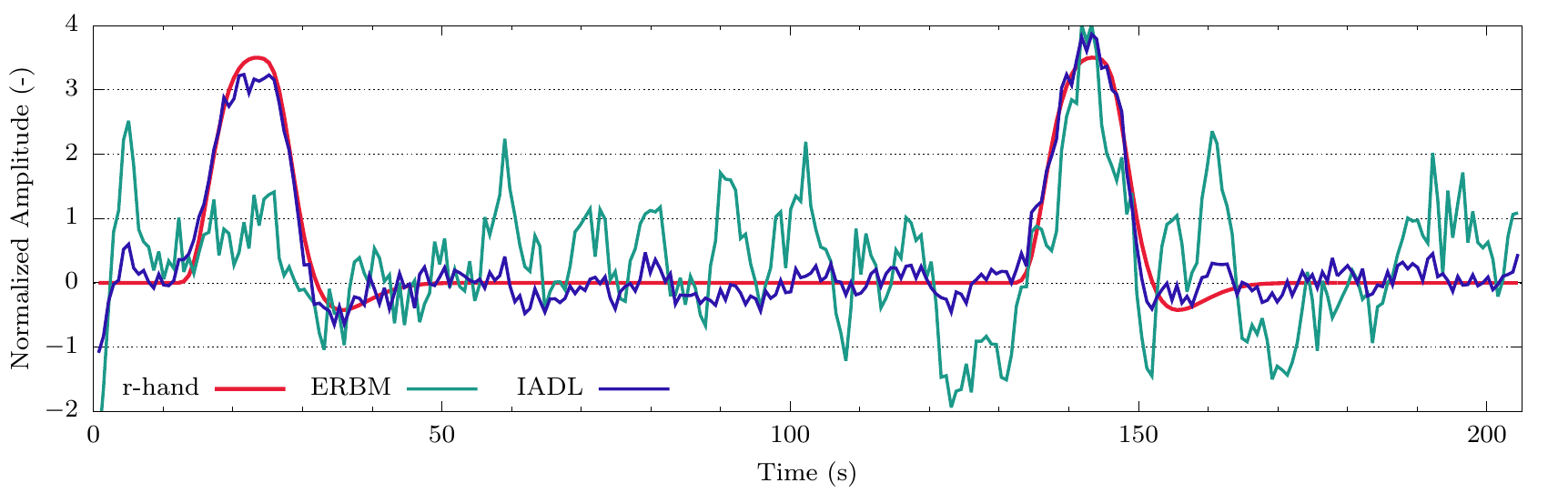}
	\caption{Time courses obtained from the different algorithms in the single subject analysis. The red line shows the task-related time course and the blue and green correspond to IADL and ERBM, respectively.}
	\label{Fig:RealTC}	
\end{figure}

\begin{figure*}[t]
	\centering
	\includegraphics[width=0.7\textwidth]{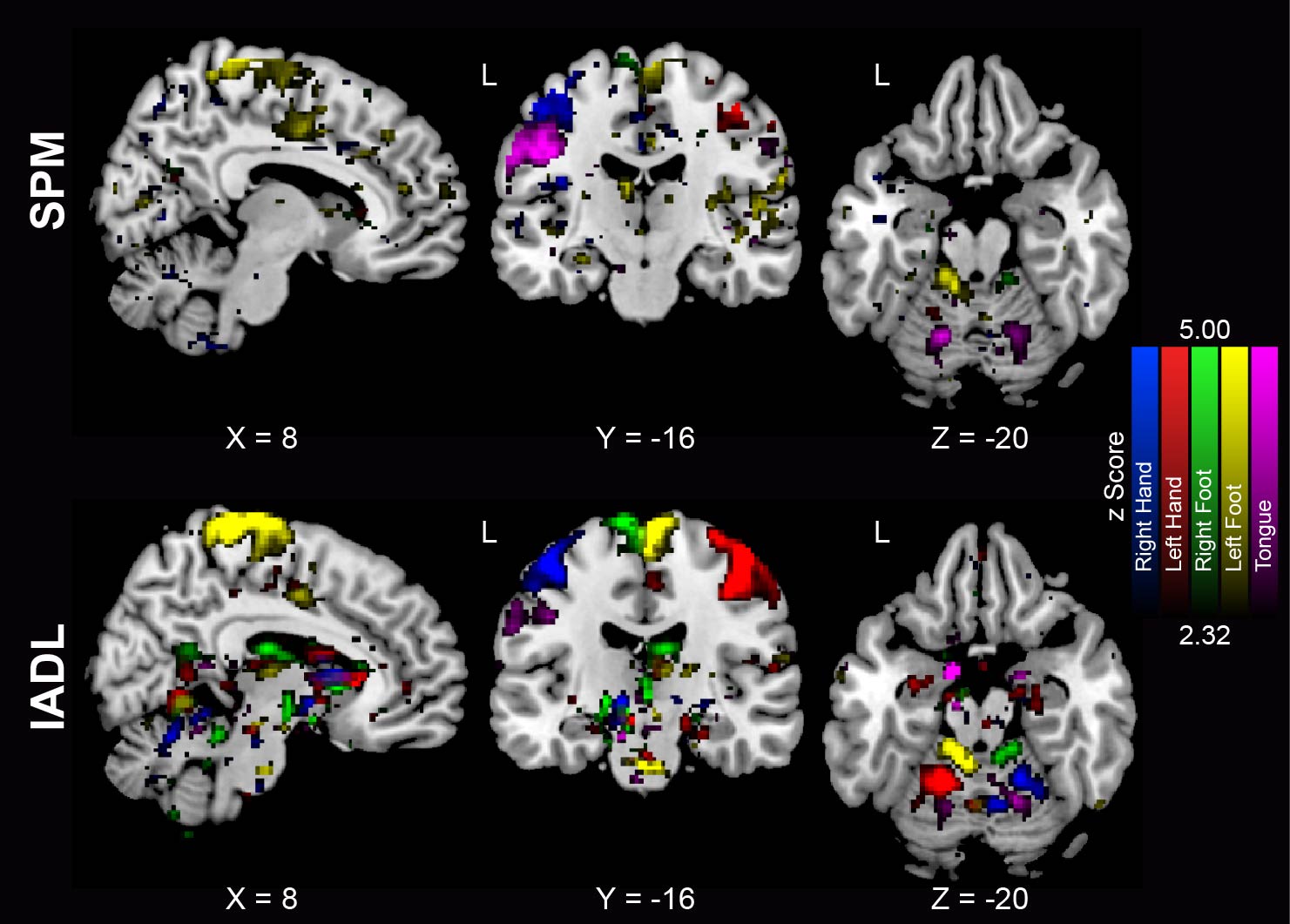}
	\caption{Significant active voxels ($2.32 < z < 5$) of the group analysis for SPM and IADL depicting all the movement conditions together.}
	\label{Fig:All_ROIs}
	
\end{figure*}

With respect to IADL, the sparsity percentage per source is shown in \Tab{\ref{Tab:Lvls}.c}, where the first 6 values correspond to the task-related time courses. Although all FBNs are known to be very sparse,  the visual FBN is one of the denser \cite{MSDL}, \cite{Pow_2011} (see also \Tab{S-I}). Therefore, the sparsity percentage that corresponds to the visual cue was set to 90\% and all the rest were set to 95\%. As it has been discussed in the previous section, IADL is robust to sparsity over determination, therefore not much difference would be expected if all values were set equal to 90\%. For the rest of the sources, (7th - 20th value in \Tab{\ref{Tab:Lvls}.c}) values of gradually diminishing sparsity percentage and $c_{\delta}$ computed to be equal to 4.6, following the same tuning approach as in \Sec{\ref{Sub:Exp_Synthetic}}, which is also described in \App{\ref{Sub:Par_cdelta}}.

We have performed both single subject analysis, where the data set of each subject is decomposed separately, and group analysis, where the datasets of all 15 subjects are concatenated along time \cite{First_GICA}. In essence, group analysis implies that all subjects share a similar common spatial map. In the IADL and SDL cases, the task-related time courses of all subjects are concatenated as well in order to construct the overall task-related time courses. Moreover, the similarity parameter $c_{\delta}$ is recalculated to the value 68 using the same procedure described in \App{\ref{Sub:Par_cdelta}}, in order to account for longer, concatenated task-related time courses.

At the single subject level, all the matrix factorization methods produce similar results for all the participants. \Fig{\ref{Fig:RandomSubject}} in \App{\ref{Sub:Extra_Single}} depicts the results of one randomly selected subject for ICA, SDL and IADL. However, for SPM, some of the subjects did not exhibit significant activity within the expected area of interest, which agree with previous analysis \cite{Motor-ROIs}.

The spatial maps that correspond to the experimental conditions for the group analysis are depicted in \Fig{\ref{Fig:Real}}. All the spatial maps from the matrix factorization methods were computed via the pseudo-inverse approach, namely $\tilde{\mathbf{S}}=\mathbf{D}^{-1}\mathbf{X}$, following the discussion in \App{\ref{Sub:Par_FBNs}}. This approach, besides being the standard in ICA-based analysis, led to enhanced performance for SDL. For the latter algorithm, the regularization parameter, $\lambda$, was optimized by inspection. In fact, due to the absence of physically consistent guidance for its tuning, we run SDL for a large set of $\lambda$ values, namely $\lambda=\{0.01, 0.02, 0.05, 0.1, 0.5, 1, 2.8, 5, 10, 50\}$ (similarly to the range proposed in the SDL paper \cite{SDL}). It turned out that with visual inspection $\lambda=10$, produced spatial maps, which were closer to the ones expected to be detected in the specific experimental design. Note that $\lambda$ was optimized in the single subject case and the same value was used in the group analysis, too. Finally, the spatial maps from SPM are the results obtained from the second level analysis.

From observing the spatial maps that resulted, it is apparent that all the methods detected significant activity within the visual cortex associated to the visual cue. However, ICA has difficulties to detect the rest of the components since only the right foot and the left hand appear recovered partially. Although the group results for ICA are somewhat better than those for the single subject analysis (see  \Fig{\ref{Fig:RandomSubject}} in \App{\ref{Sec:Extras}}), ICA still fails to recover the FBNs related to most of the motion activities. These results are expected since, as we discussed before, the sources related to the motor task are more difficult to separate compared with the visual area, which appear correctly recovered in all the studied methods.

In contrast to ICA/ERGM, the rest of the approaches (SPM, IADL and SDL) are able to properly separate the different tasks. However, it is observed that in the single subject case, SDL exhibits more spurious activity around the lateral ventricles compared to the IADL, which provides a more reliable spatial map. Similarly, although SPM present significant activity within the region of interest, the obtained significant activation are smaller than the expected activation areas of interest and the results exhibit a large amount of spurious false positives. Moreover, in the group analysis, SDL failed to recover the right hand. In contrast to that, IADL successfully unmix all the expected FBNs and exhibits an enhanced performance. Note that for IADL, the same sparsity percentage values were used for both, the single subject and group analyses, and they are defined in \Tab{\ref{Tab:Lvls}.c}.

\Fig{\ref{Fig:RealTC}} depicts, as an example, the estimated time courses corresponding to the visual area and the left hand. In particular, the red curve corresponds to the task-related time course, which is incorporated as knowledge in the case of IADL and it is fixed in the case of SPM and SDL. The green curve is the fully blind estimation obtained by ICA/ERBM. With respect to ICA, it is clear that it produces a relatively reliable estimate of the visual time course. This is in agreement with its performance in revealing the corresponding spatial maps (\Fig{\ref{Fig:Real}.a}). Still, however, there are some spurious peaks, which are associated with fixation (dark dotted curve). Furthermore, on the left hand case, ICA fails to recover one out of two block events and excessive volatility is also observed. On the contrary, in both conditions, IADL gently adjusts the task undershoots.

Finally, to further illustrate the potential of IADL, we performed a closer comparison of it with the standard SPM. Thus, \Fig{\ref{Fig:All_ROIs}} displays the significant activation clusters from the group analysis of the 15 subjects. For both approaches, the group maps are displayed between a lower threshold of $z=2.32$ ( $p<0.01$ uncorrected) and a upper threshold of $z=5$ (Bonferroni-corrected at $p<0.066$). We selected the same range of z-scores and the same coordinates as \Fig{7} in~\cite{Motor-ROIs}, to facilitate the comparison of the results.

From a general perspective, as we previously mentioned, the significant clusters from SPM, within each region of interest, are smaller compared with those from \Fig{7} in~\cite{Motor-ROIs} and IADL. Specifically, SPM seems to have problems to unmixes the areas related to the hands and the right foot within the motor areas. Moreover, SPM fails to recover most of the associated areas within the cerebellum --which are particularly more challeging \cite{Cerebellum-Buckner}-- for each movement type. In particular, SPM only exhibits significant activation clusters within the cerebellum for the tongue and the left foot. In contrast, the results of IADL are more consistent; we observe that the main activation clusters of IADL are located precisely as suggested by previous literature~\cite{Motor-ROIs, Cortex-Yeo}, even for the cerebellar areas~\cite{Cerebellum-Buckner} (see also \App{\ref{Sub:Extra_CompSPM}}).



\section{Conclusions}
In this paper, we present a new Dictionary Learning method that incorporates external information in a natural way via two novel convex constraints: a) a sparsity constraint based on the weighted $\ell_{1}$-norm, which allows to set row-wise sparsity constraints that naturally encapsulates the sparsity of the spatial maps, and b) a similarity constraint over the dictionary, which integrates external a priori information that is available from the experimental task.

The proposed sparsity constraint constitutes a natural alternative to the standard $\ell_{1}$-norm regularization, allowing to incorporate external sparsity-related information that is available in brain atlases, and by-passes the problem of selecting the regularization parameters by following cross-validation arguments that have no practical meaning, when real data are involved.

Furthermore, the incorporation of the task-related time courses from the experimental task enhances the performance decomposition of their corresponding sources. Moreover, the newly proposed constraints exhibit more tolerance and robustness against mis-modeling, compared to alternative approaches. 

Extensive simulation results, over realistic synthetic and real fMRI datasets, have verified the advantages and the enhanced performance obtained by the new proposed method.

\section*{Acknowledgment}
This research has been funded by the European Union's Seventh Framework Programme (H2020-MSCA-ITN-2014) under the grant agreement No. 642685 MacSeNet.

\end{multicols}

\begin{changemargin}{25mm}{25mm}

\newpage
\appendix
\addcontentsline{toc}{part}{Appendix}
\addcontentsline{toc}{section}{Contents}
\part*{Appendix}

\tableofcontents

\newpage

\section{Parameter setup, initialization guidelines and tips}
\label{Sec:ParamSetup}

In this section, we provide a setup guideline for running the proposed algorithm. The free parameters that need to be tuned are: a) the number of sources, $K$, b) the maximum sparsity per row, $\phi_i$, and c) the radius parameter $c_{\delta}$.  Concerning the parameters $c_{d}$ and $\varepsilon$, both are set to values so that  to satisfy straightforward stability arguments as we described in the main text. Put succinctly, we set $c_{d}$ equal to 1 to avoid ill-conditioned phenomena \cite{Allen} and $\varepsilon$ to a small value, e.g., $10^{-6}$, to avoid division by zero and to enhacne numerical stability \cite{Sergios,wl1-norm,Yan-wl1-norm}. The choice of their specific value is not critical and the algorithm is insensitive to it.

Unlike other alternative approaches, in our case, all the parameters of the proposed algorithm can be tuned based on physical arguments with respect to the fMRI study at hand. Specifically, 

\subsection{How to choose the number of sources $K$}
\label{Sub:Par_Sources}
The imposed number of sources, $K$, should be idealy equal to the true number of sources, which sums up to the number of active FBNs plus the number of machine-induced sources plus any physiological (non brain-induced) source, e.g., cardiac, movement residuals, etc. It is certainly expected that the true number of sources would vary in some degree from experiment to experiment and from subject to subject. However, the proposed algorithm is insensitive to overestimating $K$, as we illustrate in \App{\ref{Sub:Extra_EvalK}}. Therefore, it is practically easy to set $K$ to an approximate value, which grossly reflects the order of sources being present. Studies concerning the number of sources present in fMRI data, e.g., \cite{Intrinsic_Dim}, \cite{Dim-1}, \cite{Sources_2004}, have shown that the total number of sources sources are roughly 20 to 40. Therefore, an according to the results in \App{\ref{Sub:Extra_EvalK}}, the range from 20 to 40 contitutes a adequate and realistic interval, where the proposed algorithm produces similar outcomes, and one can safetly set the number of sources as any value in that interval. Note, however, that bigger number of sources will require more computation time, since larger number of sources increases the complexity of the decomposition.

Besides, there are specialized approaches that one could use to estimates the  intrinsic dimensionality of the fMRI data \cite{Intrinsic_Dim} directly; for example, Autoregresive models \cite{Intrinsic_Dim}, Principal Component Analysis (PCA) \cite{P_PCA} or Minimum Description Length (MDL) \cite{MDL}.

\subsection{How to choose the maximum sparsity per source, $\phi_i$, based on brain atlases}
\label{Sub:Par_Sparsity}
According to the constraint in \Eq{11} in the main text (see also \Eq{\ref{Eq:SetDic}}), the dictionary is split into $M$ columns, which correspond to the assisted time courses, and $K-M$ columns, which correspond to the rest of the sources. Accordingly, the first $M$ rows of $\mathbf{S}$, i.e., spatial maps, correspond to the assisted sources and the rest of the rows are related to any other source present in the data matrix. Therefore, the sparsity specification is handled differently for the two sets of rows. For those corresponding to the assisted sources, the maximum sparsity levels $\phi_i$, $i=1,\ldots,M$, need to be explicitly set for each one of them. To this end, we resort to information available via existing brain atlases, from which the sparsity level of the FBNs of interest can easily be extracted as it is described is \App{\ref{Sec:Atlas}}.

However, it should be emphasized that the algorithm does not requires the exact sparsity value; instead, an upper bound of it suffices. In other words, one needs to get an estimate of the maximum expected sparsity level. In addition, as we demonstrated in the simulations section (see \Sec{\ref{Sec:Experiments}} of the main text), the method is very robust to overestimates. This gives the users great flexibility: imagine the worst scenario; that is, setting the sparsity percentage of the assisted sources while no prior related knowledge concerning the corresponding FBNs is available. In such cases, one can safely give to all the assisted sources a sparsity percentage around, say, 85\%, which is close to the mean sparsity percentage of the primary anatomical division of the brain (see \Tab{\ref{Tab:Anatomical}}) and it is roughly the denser brain-induced source that we observed from the studied brain atlases. Of course, if one has a tighter value than that to provide --closer to the correct one-- then this can only be beneficial to the source decomposition task.

On the other hand, the sparsity level of the rest of the sources need not be tuned one by one. Parameters $\phi_i$, $i=M+1\ldots K$, can be assigned a series of values, which roughly represent the sparsity levels expected to be found in the dataset. Still the method is exceptionally robust to this parametrization and the ordering of the values does not have any effect on the result; the algorithm has the power to automatically distribute the estimated sources to the rows of $\mathbf{S}$ that they best fit; this behavior is guaranteed by the proper initialization, as we described latter in \ref{Sec:D-Init}. As a rule of thumb, we suggest to provide a gradient of ordered sparsity values having sparsity percentage from 99\% down to 80\%, in order to model expected brain-like sources and movement artifacts. Also, it is a good practice to include a few denser sparsity levels, e.g. 70\% - 0\% to model machine-induced artifacts, unmodeled movement corrections and potential noise-like residuals.

\subsection{How to select the radius parameter $c_{\delta}$}
\label{Sub:Par_cdelta}
By definition, the parameter $c_{\delta}$ constraints the energy of the residual between the task-related time course, $\bdelta$, and the subject-specific true time course, $\mathbf{d}$. In practice, these time courses are  estimated using the linear convolution with the corresponding HRF, that is:
\begin{equation}
	\left\Vert \bdelta-\mathbf{d}\right\Vert =\left\Vert \mathbf{u}*\mathbf{h}_{c}-\mathbf{u}*\mathbf{h}_{s}\right\Vert =\left\Vert \mathbf{u}*(\mathbf{h}_{c}-\mathbf{h}_{s})\right\Vert \leqslant c_{\delta}^{\nicefrac{1}{2}}
	\coma
	\label{Eq:Con-u}
\end{equation}
where $\mathbf{u}$ is the specific neuronal activation pattern that theoretically matches the experimental task and $\mathbf{h}_{c}$, $\mathbf{h}_{s}$ are the canonical HRF (cHRF) and the unknown subject-specific HRF. Note that, in general, the neuronal activation pattern and the experimental one might not be exactly the same: any miss-modeled activation pattern will increase the difference between the real and the considered time course.

Therefore, under the convolutional model, we observe that the parameter $c_{\delta}$ reflects our confidence on how accurate is the cHRF for the subject under consideration, as well as, how reliable is the imposed experimental condition, which theoretically matches the corresponding activation pattern of the subject. Setting $c_{\delta}=0$, no flexibility is allowed at all and the information regarding the task-induced time courses is equivalent to that provided in the GLM/SPM framework. Such a strategy is also acceptable and the whole analysis can still be benefited from the rest of the sources that will be estimated, in the same way as the SDL in \cite{SDL} does. However, there are good reasons to allow the task-related time courses to drift from the imposed cHRF-based ones: a) subject HRF will indeed diverge from the cHRF and b) the convolution of the HRF with the activation patterns is a linear function that is unable to account for natural nonlinear effects, produced when short interstimulus intervals are considered \cite{NonLinear-HRF}. Therefore, the use of a $c_{\delta}$ moderately larger than zero can only be of benefit. Note, however, that large values of $c_{\delta}$ will cancel the benefit of exploiting a priori imposed information. Indeed, the extreme case that $c_{\delta}=\infty$ corresponds to the case where the columns of $\mathbf{D}$ are not constrained at all.

In order to obtain a meaningful value, assuming that the specific neuronal activation pattern matches the experimental task, we used the results in \Eq{\ref{Eq:Con-u}} to obtain an adequate estimate of $c_{\delta}$, by explicitly measuring the distance between the task-related time courses generated by the HRF and a significantly different HRF, yet realistic. This will ensure enough flexibility to accommodate the expected natural variability of the HRF. In general, when two or more task-related time courses are implemented, a different estimate of the similarity parameter per task-related time course is obtained. All these estimates provide information about how strict/relaxed the similarity constraint needs to be for each specific task-related time course. Therefore, a good candidate for $c_{\delta}$ is the mean value of all the obtained estimates, since this value considers the expected mean variability among all the imposed task-related time courses. However, note that other potential alternatives may also be used; for example, one may set the similarity parameter, $c_{\delta}$, equal to the minimum (maximum) estimate among the task-related time courses for a more conservative (relaxed) approach.

In this paper, based on the previous rationale, we estimate the similarity parameter, $c_{\delta}$, for the different studied fMRI experiments as follows: first, given all the conditions, we determine their corresponding task-related time courses, $\bdelta$, using the cHRF and an extra task-related time course, $\bdelta^{*}$, generated using the HRF 5 from the \Fig{\ref{Fig:HRFs}}, which constitutes a  significantly different, yet realistic, HRF compared with the cHRF. Then, according to \Eq{\ref{Eq:Con-u}}, the difference $\left\Vert \bdelta-\bdelta^{*}\right\Vert ^{2}$ provides an estimate of the similarity parameter for each particular task. Finally, we set $c_{\delta}$ as the mean value obtained among the different tasks. For completeness, we provide a Matlab function that automatically performs this procedure given a particular set of experimental tasks. This function is available in AIDL$^{\ref{fnt:AIDL}}$ repository and is used to set the default value of this parameter in within the main algorithm.

\subsection{Dictionary Initialization}
\label{Sub:Par_D-Init}
In general, the matrix factorization approaches, as described in \Eq{\ref{Eq:Main}}, admit an infinite number of potential solutions and also constitutes a \emph{non-convex} optimization task. Consequently, the solution that will be obtained via the proposed algorithm, as well as by any BSS counterpart, depends on the initialization of $\mathbf{S}^{[0]}$ and $\mathbf{D}^{[0]}$. As a result, a careful initialisation is more likely to lead to a better outcome rather than initialising from a random guess of $\mathbf{S}^{[0]}$ and $\mathbf{D}^{[0]}$. For this reason, a systematic initialisation procedure comprising four steps is proposed here:
\begin{enumerate}
    \item An initial estimate of $\mathbf{S}$ and $\mathbf{D}$, denoted as $\bar{\mathbf{S}}$ and $\bar{\mathbf{D}}$ is obtained via  any ICA algorithm. A known issue with ICA methods is that the same FBN, from the same independent component, might be split into several time-courses/spatial-map vectors \cite{Correa},\cite{Sources_2004}. To address this issue, the split sources are merged back following the merging process described in \cite{Yannis-DL}.
    \item The columns of $\bar{\mathbf{D}}$ that are most correlated with the $M$ task-related time courses are detected. Then, the columns of $\bar{\mathbf{D}}$ and the corresponding rows of $\bar{\mathbf{S}}$ are rearranged in order to bring the $M$ most correlated columns on the far left of $\bar{\mathbf{D}}$ and the corresponding rows on the top of $\bar{\mathbf{S}}$. Such a rows-columns rearrangement is allowed since the product $\mathbf{D}\mathbf{S}$ is invariant to such permutations. Then, the $M$ first columns of $\mathbf{D}$ are replaced with the imposed task-related time courses. This step brings the initial $\bar{\mathbf{D}}$ in the form described in \Eq{11}.
    \item The initialization is further refined using the proposed algorithm for a few iterations but by imposing sparsity in the full coefficient matrix, i.e., using the alternative constrained set $\mathfrak{L}_{W}$ described in \Eq{\ref{Lwcap}}, rather than imposing sparsity row-by-row. The reason for including this extra step is twofold: first, the ICA-based estimate is gently improved by using the imposed information concerning both the overall sparsity and the imposed task-related time courses. Second, the resulting coefficient matrix $\bar{\mathbf{S}}$ will become sparse and in agreement to the imposed sparsity specifications. Indeed, this was not the case before the execution of this step, since the coefficient matrix produced by ICA is dense. In particular, using the pseudocode described in Algorithm 1 in the main text, the for loop in lines 6-10 is replaced by:\[\begin{array}{l}
\text{\textbf{if }}\left\Vert \mathbf{A}\right\Vert _{1,\mathbf{W}}\geqslant\Phi=\sum_{i=1}^{K}\phi_{i}\\
\;|\qquad\mathbf{A(:)}=\mathcal{P}_{B_{\ell_{1}}[\mathbf{W(:)},\Phi]}(\mathbf{A(:)})\\
\text{\textbf{end}}
\end{array}
    \]
    where the notation $\mathbf{A}(:)$ stands for the vectorized form of the matrix $\mathbf{A}$. 
    \item The last step is to properly fit the initialization of $\bar{\mathbf{D}}$ and $\bar{\mathbf{S}}$, computed in the previous step, with the row-by-row sparsity constraints required by the algorithm. In order to achieve this, the last $M+1,\ldots, K$ columns of $\bar{\mathbf{D}}$ and $\bar{\mathbf{S}}$ are permuted in order for the last $K-M$ rows of $\bar{\mathbf{S}}$ to exhibit a gradient of sparsity levels from the sparser to the denser. Then $\mathbf{D}^{[0]}$ and $\mathbf{S}^{[0]}$ in the algorithm is set equal to the finally resulted $\bar{\mathbf{D}}$ and $\bar{\mathbf{S}}$ respectively  
    \end{enumerate}
Although the user is advised to proceed with the initialisation method described above, the algorithm is theoretically guaranteed to converge to a local minimum irrespectively of the initial values of $\mathbf{S}^{[0]}$ and $\mathbf{D}^{[0]}$. Note, also, that using ICA as initialization point has also a Bayesian flavor: the imposed sparsity constraint that acts componentwise (see $\mathfrak{L}_{w}$) works similarly as an implicit independence assumption over a prior. Of course, independence over the prior does not necessarily mean independence over the posterior. In this way, initializing our algorithm from ICA constitutes an effective synergy between ICA and DL; using ICA only as initialization allows the DL algorithm to go beyond the independence assumption, yet preserving the contribution of ICA as a good reference point. Similarly, other approaches have tried to combine the benefits of sparsity and independence. For example, in \cite{Semi-ICA-Calhoun} a hybrid method called sparse-ICA tries to integrate both approaches into a single algorithm. However, the independence assumption as well as the lack of consistent ways to tune the sparsity constraint limit the power of these approaches in the fMRI case.

\subsection{Detection of the FBNs in Real datasets}
\label{Sub:Par_FBNs}

The standard practice when working with BSS techniques for fMRI analysis is the detection and representation of the obtained sources/FBNs on the brain. In the fully-blind case, the time courses of the dictionary, hopefully, produce reasonable estimates of their corresponding hemodynamic responses, including those induced by the FBNs of interest. Similarly, the semi-blind methods assume that the imposed task-related time course assembles an adequate representation of the real brain activity.

An alternative way to determine the spatial distribution of each source is to use the obtained time courses to get an estimate of the coefficient matrix directly, say $\tilde{\mathbf{S}}$, using the pseudoinverse of the dictionary, that is, $\tilde{\mathbf{S}}=\mathbf{D}^{-1}\mathbf{X}$, where $\mathbf{X}$ is the data matrix. In practice, this is a well-established approach particularly when working with ICA methods, which provides directly an estimate of the pseudoinverse of the dictionary, usually referred as ``unmixing matrix'', and is later used to determine the coefficient matrix from the data. Under this approach, each row of the obtained coefficient matrix represents the ``weights'' that specify the contribution of each time course to its corresponding spatial map. Then, the values of each spatial map can be mapped back to the brain volume space, representing its spatial distribution on the brain. Nevertheless, the obtained coefficient matrix, $\tilde{\mathbf{S}}$, potentially turns out dense, i.e., the most of its values are non-zero, since noise and small residuals from the decomposition may also appear, interfering with the location of the main patterns. Consequently, to obtain interpretable results, the spatial maps are ``sparsified'' using some significant tests, such as $z$-scores, which provide statistical support to infer significant activation patterns that correspond to each source \cite{Raj-2001}, \cite{Comparisons}, \cite{Zscore-1}, \cite{Zscore-2}, \cite{SDL}. Similarly, this modus operandi is valid for DL methods, where one can use the pseudoinverse to estimate the coefficient matrix and use statistical tests to infer their corresponding significant activation patterns.

On the other hand, sparsity-based approaches offer an alternative way for the representation of the spatial distribution of the brain activity. In general, sparsity-based approaches, such as DL, simultaneously produce estimates of the dictionary and its corresponding sparse representation over the coefficient matrix, $\mathbf{S}$. Theoretically, proper sparsity constraints will naturally force to zero the values of each spatial map, which appear unrelated to the specific source. On the contrary, the non-zero values represent the regions with important activity. Consequently, for each spatial map, the representation of the values $\mathbf{s}^{i}>0$ provides also meaningful information concerning to the spatial distribution of the $i^{\mathrm{th}}$ detected FBN.

In this paper, we followed the first standard procedure to represent the main activation patterns using statistical test, since it provides a fair comparison between the studied methods. Thus, the real fMRI results presented in the \Sec{\ref{Sub:Exp_AllComp}} of the main text were obtained as follows: first, using the pseudoinverse of the dictionary, we estimate its corresponding coefficient matrix. Then, we scaled each spatial map to $z$-values; voxels that exhibited values greater than $z\geqslant 1.97$ were considered as significant active voxels \cite{Raj-2001}. Finally, the $z$-statistical scores of the detected blobs were superimposed on its corresponding anatomical slices for visualization.

For completeness, we also explored the alternative way to display the spatial distribution of the obtained FBNs, using the sparse coefficient matrix directly. In this case, we only studied two sparsity-based methods: SDL and IADL. Thus, the comparison between the obtained sparse spatial maps are presented in \App{\ref{Sub:Extra_SparseRep}} as additional results.

\section{Brain Atlas Sparsity Information}
\label{Sec:Atlas}

In general, brain atlases contain information regarding the anatomical structure of the spatial distribution of some FBNs. In our context, this information can directly lead to infer the \emph{sparsity percentage} of a particular FBN; that is, to estimate the proportion of voxels that belong to that FBN, which can be used as extra a priori information for tuning the associated sparsity parameters.

In practice, we distinguish two types of atlases: functional connectivity atlases and anatomical/structural atlases. The function connectivity-based atlases contain information about the spatial distribution of different FBNs, regardless of their anatomical proximity or location. Some examples of this type of atlases are \cite{MSDL} and \cite{Pow_2011} , which contain some standard FBNs, such as Auditory, Default Mode Network, Language, etc. On the other hand, the anatomical-based atlases use several anatomical/structural arguments to define different brain partitions. Although a priori, anatomical atlases may not provide information concerning the sparsity percentage of a specific FBN, the particular value corresponding to the FBN can be easily estimated using the sparsity percentage of the principal regions of interest occupied by that FBN over the anatomical atlas. For example, assume one particular FBN distributed over $N$ distinct anatomical region of one anatomical brain atlas, with sparsity percentages $\theta_{1},\theta_{2},\ldots,\theta_{N}$. Then, the sparsity percentage for that FBN, $\theta_{FBN}$, is given by:

\begin{equation}
	\theta_{FBN}=100(1-N)+\sum_{i}^{N}\theta_{i}
	\punto
	\label{Eq:Th_FBN}
\end{equation}
Note than this equation holds if and only if the anatomical areas are not overlapping, which is the case for any anatomical brain atlas.

Consequently, we observed that both types of atlases are suitable for obtaining estimates of the sparsity percentage of a specific FBN. To illustrate the reliability and consistency of this procedure, we present the sparsity percentage of some common FBNs, using two different brain atlases: a) the Automatic Anatomical Labeling (AAL) atlas\footnote{The Automated Anatomical Labeling atlas is a digital human brain atlas with labeled volume. Labels are indicating macroscopic brain structures. See \url{http://neuro.imm.dtu.dk/wiki/Automated_Anatomical_Labeling}}, which is a well established anatomical-based brain atlas that contains a total of 116 regions of interest, and b) a Multi-Subject Dictionary Learning (MSDL) atlas \cite{MSDL}, which is a functional connectivity-based atlas that includes some common FBNs obtained from a resting state-fMRI study. \Tab{\ref{Tab:SparsityFBN}} shows the sparsity percentage with respect to the brain for some FBNs for both atlases. In this case, for the MSDL atlas, the FBNs were already defined, so the sparsity percentage was measured directly. With respect the AAL, we measured the total sparsity of the principal regions of interest that are occupied by the specific FBN of interest using \Eq{\ref{Eq:Th_FBN}}. Comparing the values of the sparsity percentage for some common FBNs (see \Tab{\ref{Tab:SparsityFBN}}), observe that the obtained values are similar in both atlases, despite that each atlas was independently defined following different approaches. Hence, the sparsity percentage is independent of the atlas and any brain atlas potentially contains exploitable information. For completeness, the \Tab{\ref{Tab:Anatomical}} also illustrates the sparsity of some of the principal anatomical structures of the human brain from the AAL atlas.

\begin{table}
\centering
\caption{Sparsity percentage of some FBNs using two brain atlases.}
\label{Tab:SparsityFBN}
\begin{tabular}{lcc}
\textbf{Functional Brain Network} & \textbf{MSDL (\%)} & \textbf{AAL (\%)}\tabularnewline
\hline 
\hline 
Auditory & 96.00 & 95.57\tabularnewline
Primary Visual Cortex & 93.11 & 93.20\tabularnewline
Language & 94.57 & 94.09\tabularnewline
Basal Area & 94.83 & 96.18\tabularnewline
Default Mode Network & 95.26 & 93.21\tabularnewline
\hline 
\end{tabular}
\end{table}

\begin{table}
	\centering
	\caption{Sparsity percentage of the lobes of the brain and the cerebellum}
	\label{Tab:Anatomical}
\begin{tabular}{lcc}
\textbf{Anatomical Region} &  & \textbf{AAL (\%)}\tabularnewline
\hline 
\hline 
Frontal Lobe &  & 74.07\tabularnewline
Temporal Lobe &  & 88.38\tabularnewline
Occipital Lobe &  & 88.26\tabularnewline
Partietal Lobe &  & 86.23\tabularnewline
Crebellum &  & 89.13\tabularnewline
\hline 
\end{tabular}
\end{table}

\section{Realistic Synthetic Dataset}
\label{Sec:SyntheticDataset}
In this section, a new, highly realistic synthetic dataset is developed using as a base, SimTB\footnote{See: \url{http://mialab.mrn.org/software}}, an open MATLAB toolbox for the design of fMRI-like data, which has been widely used in several studies, such as \cite{SimTB-1}, \cite{SimTB-2} and \cite{SimTB-3}.
This dataset allows to effectively evaluate the performance of the proposed DL method in comparison with the state-of-the-art of the blind and semi-blind approaches. 

The new dataset resembles demanding experimental designs, where the sources present highly relative different energy and their corresponding brain areas that correspond to different condition overlap. This renders them hardly detectable by conventional BSS methods. Moreover, the new dataset is examined under realistic noise levels and statistics and it also comprises a relatively large number of components/sources that exhibit a large range of sparsity characteristics. In order to succeed in the above and also to study the effect of HRF miss-modeling, certain interventions in the SimTB data generation procedure needed to take place. 

SimTB implements a simple spatiotemporal separable model characterized by three fundamental features: a) the spatial maps are smooth (Gaussian-shaped) b) the sources are barely overlapped and c) a fixed HRF is considered. Although these features simplify the study of synthetic experiments, they are unrealistic; the overlap between the standard sources of SimTB only affects small areas with relatively low intensity, whereas, in real fMRI, some sources may exhibit high overlap \cite{Protopapas}. Moreover,  HRF naturally varies among subjects \cite{HRF-Var}, \cite{HRF-Var-2}.

For these reasons, we modified the default mode of SimTB to accomplish a more realistic synthetic fMRI-like dataset: first, we translated in space some of the sources to increase their relative overlap. Second, instead of smooth Gaussian shaped spatial maps, which show maximum activity at only one specific peak/voxel, we prefer to have more than one neighbor voxels activated with the same intensity. We achieved that, by starting from the standard SimTB's spatial maps and reducing the highest activated voxels to have the same relative energy, within a small neighborhood.

Regarding the HRF, many different parametric models have been built that capture its shape, with the most popular being two gamma distributions model \cite{SPM-Book}, \cite{SPM-web}, \cite{HRF-Var-2}. A certain set of parameter values produces the cHRF  that represents a parametric approximation of the real HRFs \cite{SPM-Book}. The cHRF is the default model in many different toolboxes such as SimTB \cite{SimTB-1}, and and it is also used in the GLM/SPM analysis for the computation, via convolution, of the task related time-courses. Since one of the scopes of this study is to analyze the effects of HRF mismodelling, we account for the natural variability of the HRF, emulating six different \emph{synthetic subjects}, with each one of them having its own specific HRF. Then, we generate one dataset per subject since all the brain-induced sources are HRF-dependent.

The distinct subject-specific HRFs were generated using the two-gamma distribution model and in order to ensure that they are realistic enough, we followed the procedure described next. We fed the free parameters of the two-gamma distribution model with values drawn from a uniform distribution centered on the parameter values that correspond to the cHRF. The support of the aforementioned distribution was experimentally tuned to a level that led to HRFs, which visually resemble the natural variability of HRFs. For better inspection, a total of 100 HRFs was generated (gray lines in \Fig{\ref{Fig:HRFs}}) and qualitatively compared with natural HRFs estimated from real fMRI studies, e.g., \Fig{3} and \Fig{4} of \cite{HRF-Var-2}. It is clear that the 100 simulated HRFs are remarkably similar to true ones. For the simulation studies that we presented in the \Sec{\ref{Sec:Experiments}} of the main text, five randomly generated HRFs, as well as the cHRF, were used (depicted with colored curves in \Fig{\ref{Fig:HRFs}}). For simplicity, hereafter, and  we define the ``canonical'' subject as the subject that will use the cHRF,  we introduce the letters A, B, C, D and E to refers to the subjects that correspond to the HRFs 1, 2, 3, 4 and 5, respectively. 

The subject-specific datasets comprise 15 brain-like sources. Each source is defined by its spatial map, which is represented by a single brain slice of size $100 \times 100$ voxels and a time course that comprise 300-time instances with a difference of two seconds between acquisition times (TR=2). \Fig{\ref{Fig:Srcs}} shows the brain-like sources for the canonical subject, which includes all the modifications that we discussed above. The rest of the subjects correspond to the same spatial maps but their time courses differ, depending on their respective HRF. Moreover, in \Tab{\ref{Tab:Srcs}}, the anatomical correspondence as well as the sparsity level of each source are also listed.

\begin{figure}
	\centering
	\textbf{Synthetic Brain-like Sources}

	\includegraphics[width=0.40\columnwidth]{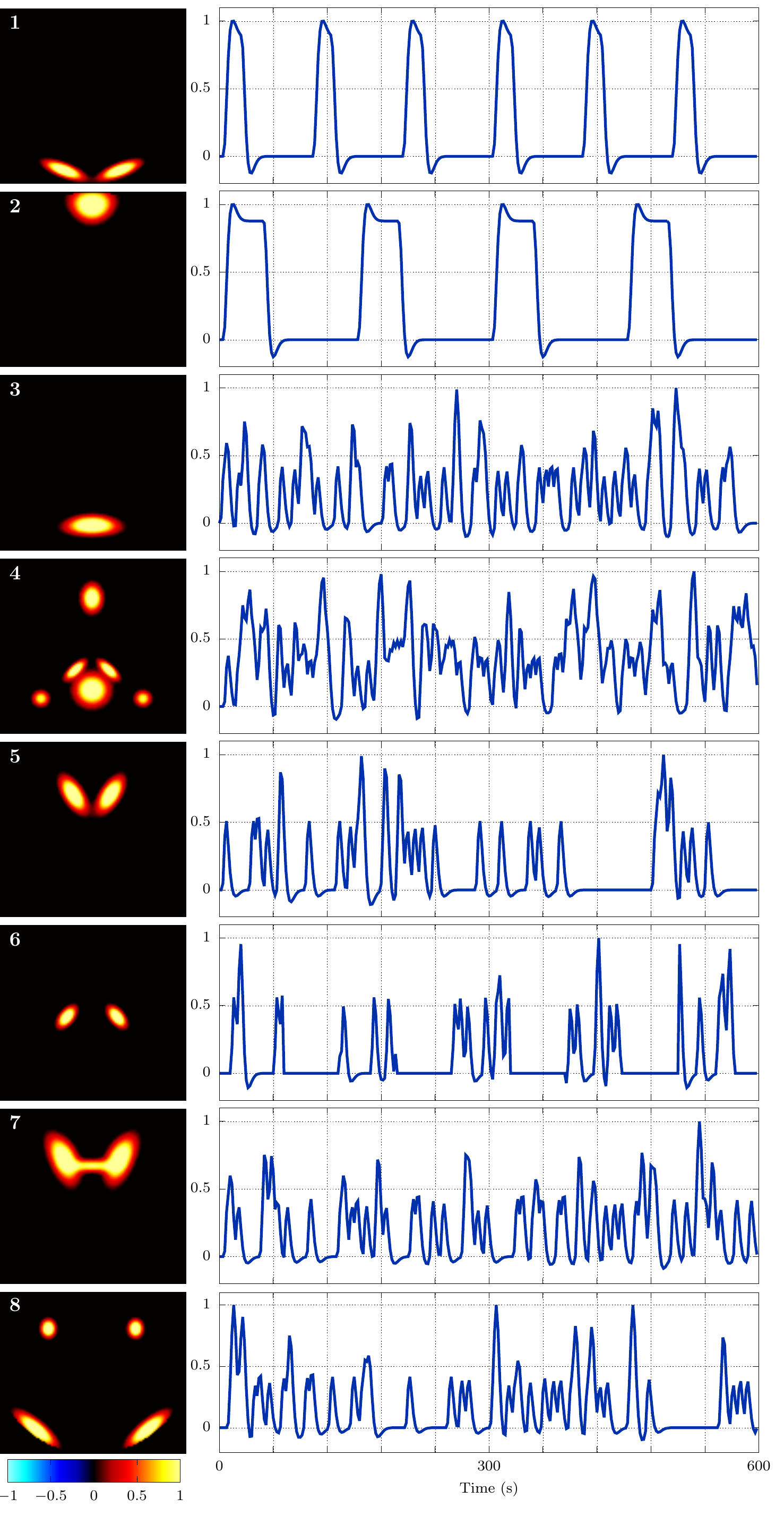}
	\includegraphics[width=0.40\columnwidth]{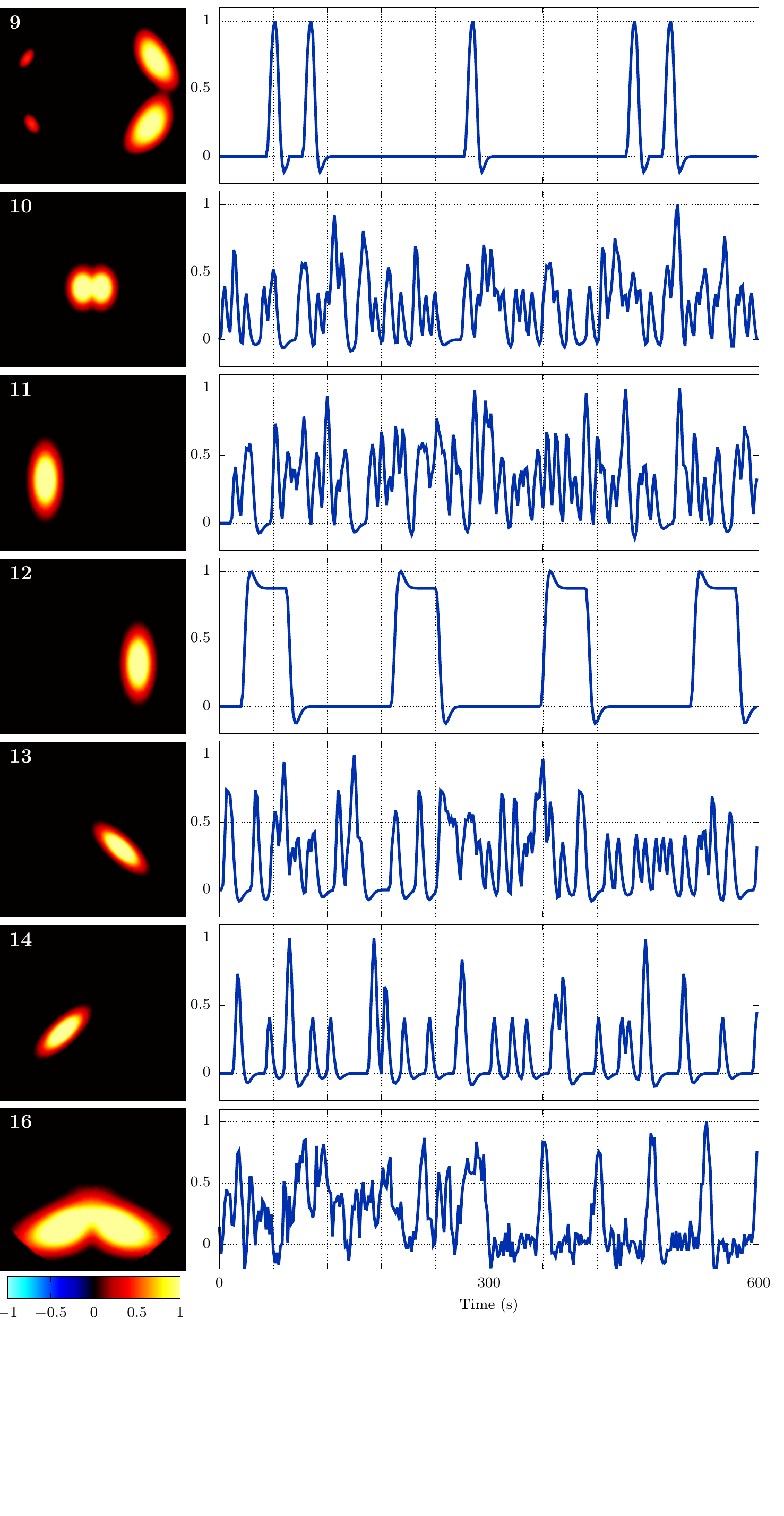}
	
	\vspace{-4mm}

	\caption{Visual representation of the synthetic spatial maps and their corresponding time courses generated with the canonical HRF. The intensity of the spatial maps and time courses were normalized to have maximum intensity at 1.}
	\label{Fig:Srcs}

	\textbf{Artifacts}
	
	\includegraphics[width=0.40\columnwidth]{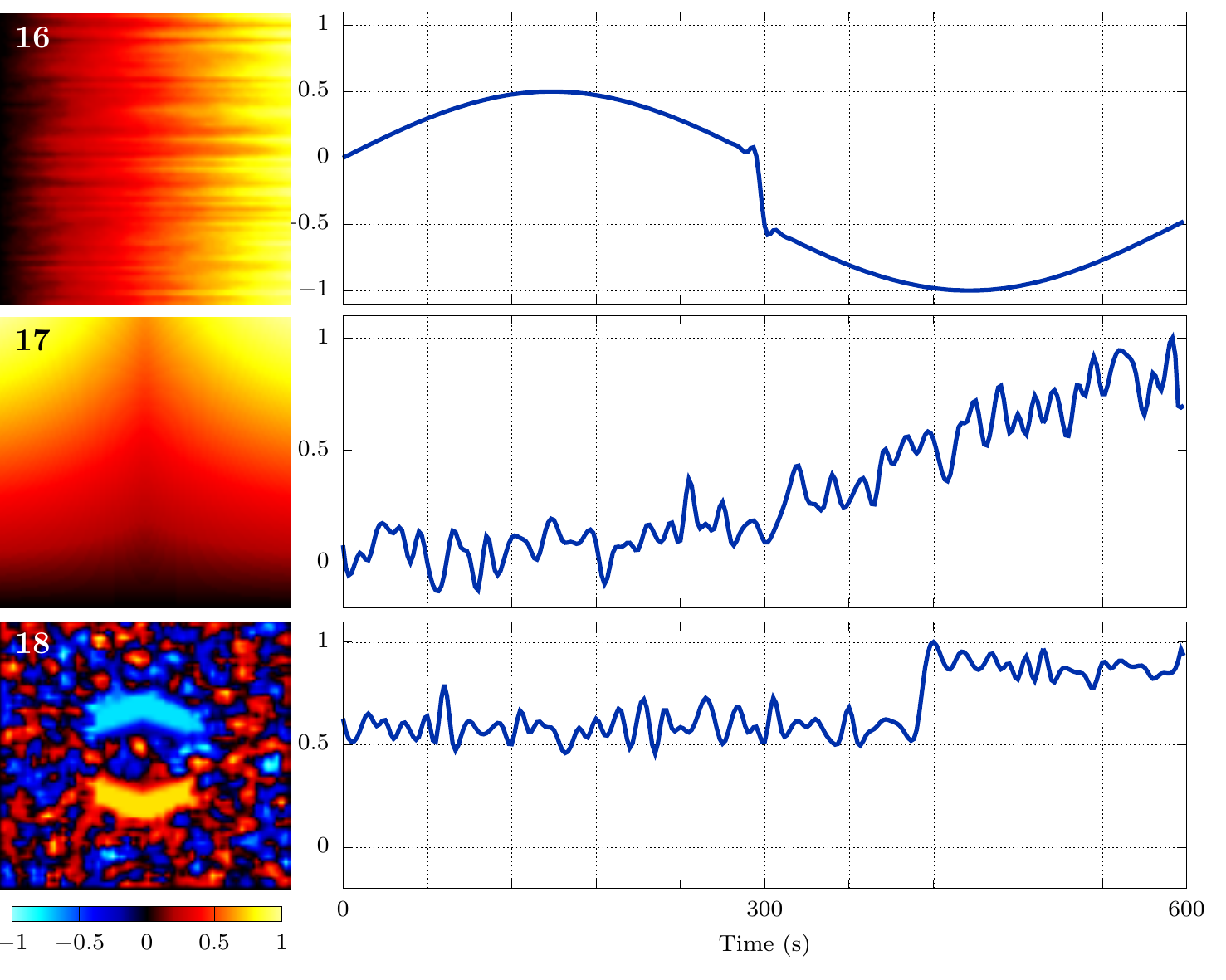}
	\includegraphics[width=0.40\columnwidth]{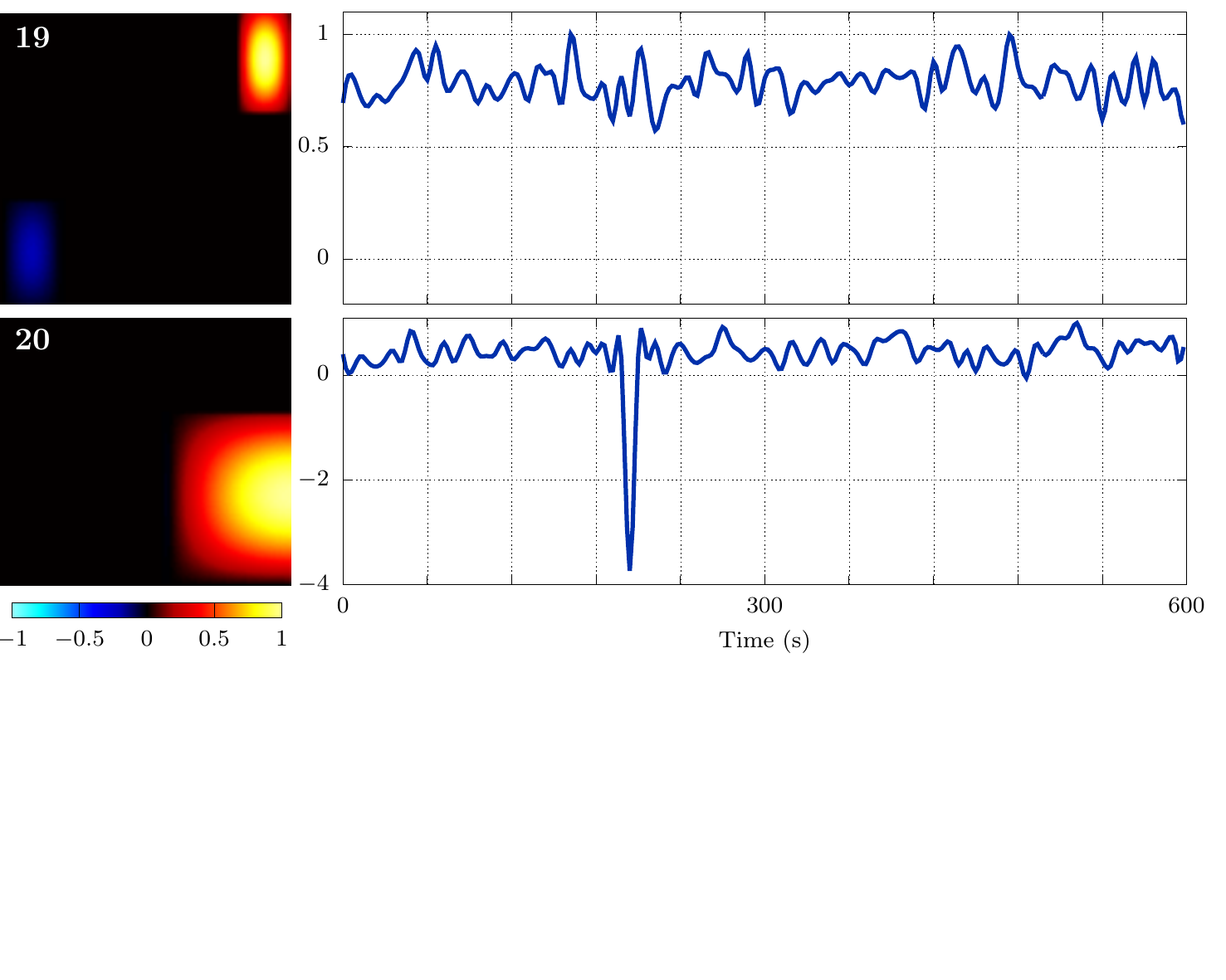}
	
	\vspace{-4mm}

	\caption{Visual representation of the synthetic artifacts. The first column depicts the spatial maps and their corresponding spatial maps for the artifacts 3, 4 and 5, whereas the second column shows the artifacts 7 and 8 from the dataset \cite{Artifacts}.}
	\label{Fig:Arts}	
\end{figure}
\begin{table*}
	\centering
	\caption{Summary of the main features of the spatial maps of the proposed synthetic dataset, including their corresponding sparsity level.}
	\label{Tab:Srcs}	
	\begin{tabular}{clcc}
\multicolumn{4}{c}{\textbf{Brain-like sources}}\tabularnewline
\hline 
Surce & Anatomical Correspondence${}^{*}$ & Active voxels ($\phi_{i}$) & Sparsity (\%) ($\theta_{i}$)\tabularnewline
\hline 
\hline 
1 & Bilateral Visual & 472 & 95.28 \%\tabularnewline
2 & Medial Frontal & 467 & 95.33 \%\tabularnewline
3 & Precuneous & 447 & 95.53 \%\tabularnewline
4 & Default Mode Network & 1175 & 88.25 \%\tabularnewline
5 & Subcortical Nuclei & 670 & 93.30 \%\tabularnewline
6 & Subcortical Nuclei - Putamen & 293 & 97.04 \%\tabularnewline
7 & White Matter Tracts (anterior) & 1193 & 88.07 \%\tabularnewline
8 & Dorsal Atention Network & 818 & 91.82 \%\tabularnewline
9 & Frontoparietal (Right dominance) & 1449 & 85.51 \%\tabularnewline
10 & Subcortical Nuclei - Thalamus & 733 & 92.67 \%\tabularnewline
11 & Right Auditory & 840 & 91.60 \%\tabularnewline
12 & Left Auditory & 847 & 91.53 \%\tabularnewline
13 & Right Hipocampus & 549 & 94.51 \%\tabularnewline
14 & Left Hipocampus & 543 & 94.57 \%\tabularnewline
15 & White Matter Tracts (posterior) & 2805 & 71.95 \%\tabularnewline
\hline 
 & \footnotesize{${}^{*}$ see SimTB toolbox \cite{SimTB-1}} & & \tabularnewline
 &  &  & \tabularnewline
\multicolumn{4}{c}{\textbf{Artifacts}}\tabularnewline
\hline 
Source & Main fatures${}^{**}$ & Active voxels ($\phi_{i}$) & Sparsity (\%) ($\theta_{i}$)\tabularnewline
\hline 
\hline 
16 & Sub-Gaussian & 9900 & 1.00 \%\tabularnewline
17 & Gaussian & 9900 & 1.00 \%\tabularnewline
18 & Super-Gaussian & 9801 & 1.99 \%\tabularnewline
19 & Sub-Gaussian & 1386 & 86.14 \%\tabularnewline
20 & Super-Gaussian & 2816 & 71.84 \%\tabularnewline
\hline 
 & \footnotesize{${}^{**}$ according to \cite{Artifacts}} & &\tabularnewline
 &  &  & \tabularnewline
\multicolumn{4}{c}{\textbf{Mean values}}\tabularnewline
\hline 
 & Group & Active voxels & Sparsity (\%)\tabularnewline
\hline 
\hline 
 & Whole dataset      &  2355 & 76.45 \%\tabularnewline
 & Brain-like sources &   887 & 91.13 \%\tabularnewline
 & Artifacts          &  6761 & 32.39 \%\tabularnewline
 \hline
	\end{tabular}
\end{table*}

Besides the brain-induced sources, real fMRI data also contains sources that are either scanner-induced or related to other biological processes unrelated to the brain activity, such as heart-beating, breathing, movements, etc. All these phenomena are collectively referred to as \emph{artifacts}. In the new dataset, we have also included realistic fMRI-like artifact sources, in particular the sources 3, 4, 5, 7 and 8 of the well-known dataset \cite{Artifacts}. Special information concerning each artifact is also provided in \Tab{\ref{Tab:Srcs}}. Note that the majority of the brain-like sources are sparse, whereas most of the artifacts are dense. Finally, the subject-specific datasets are corrupted by Rician noise of $\text{SNR}=0$. Both noise distribution and energy are realistic for the fMRI data case, \cite{Rician-Noise}, \cite{Noise-lvls}.

\newpage

\section{Pearson's correlation coefficient - based Performance measure}
\label{Sec:Measurements}
 In this direction, we are interested in measuring the discrepancy between the true and the estimated sources. Since each source expresses itself with a time course/spatial map pair, $(\mathbf{d}_{i}\in\mathbb{R}^{T\times1},\mathbf{s}^{i}\in\mathbb{R}^{1\times N})$, a way to jointly evaluate the decomposition performance with respect to both is to work with the matrix, $\mathbf{F}_{i}=\mathbf{d}_{i}\mathbf{s}^{i}\in\mathbb{R}^{T\times N}$, which represent the full $i^{\mathrm{th}}$ source as it is expressed in all voxels along time. Ideally, $\mathbf{F}_{i}$ should be equal to $\tilde{\mathbf{F}}_{j} = \tilde{\mathbf{d}}_{j} \tilde{\mathbf{s}}^{j}\in\mathbb{R}^{T\times N}$, where $\tilde{\mathbf{d}}_{j}$ and $\tilde{\mathbf{s}}^{j}$ denotes the time course and the spatial map of the corresponding true source. Then, to measure the quality of the performance for the $j^{\mathrm{th}}$ true source, we adopted:
\begin{equation}
    r_{j}=\left(\rho(\tilde{\mathbf{F}}_{j},\mathbf{F}_{i})\right)^{2}
    \coma
    \label{Eq:ri}
\end{equation}
where $\rho(\mathbf{A},\mathbf{B})$ $\mathbf{A},\mathbf{B}\in\mathbb{R}^{M\times N}$, is the Pearson's correlation coefficient for matrices given by:
\begin{equation}
	\rho\left(\mathbf{A},\mathbf{B}\right)=\frac{1}{M+N-1}\sum_{i,j=1}^{M,N}\left(\frac{a_{ij}-\mu_{\mathbf{A}}}{\sigma_{\mathbf{A}}}\right)\left(\frac{b_{ij}-\mu_{\mathbf{B}}}{\sigma_{\mathbf{B}}}\right)
	\coma
\end{equation}
where $\mu_{\mathbf{A}}$ and $\sigma_{\mathbf{A}}$ are the mean and the standard deviation of $\mathbf{A}$, respectively and $\mu_{\mathbf{B}}$, $\sigma_{\mathbf{B}}$ are the mean and the standard deviation of $\mathbf{B}$.

Observe that the measured value in \Eq{\ref{Eq:ri}} assumes  that the $i^{\mathrm{th}}$ unmixed source corresponds to the $j^{\mathrm{th}}$ true one. Nevertheless, in practice, the correspondence between the true sources and the unmixed sources is unknown (since the solution of the matrix decomposition is usually equivalent under permutation). Therefore, to measure the quality of the performance it is necessary to find the correspondence of each one among the  $\tilde{K}$ true sources with one of the  $K$ estimated sources (where in general $\tilde{K}\neq K$). The procedure followed is described next:
\begin{enumerate}
    \item Define the matrix $\mathbf{C}\in\mathbb{R}^{\tilde{K}\times K}$, with $c_{ij}=(\rho(\tilde{\mathbf{F}}_{i},\mathbf{F}_{j}))^{2}$. Observe that the larger an entry $c_{ij}$ is the more likely the $i^{\mathrm{th}}$ real source and the $j^{\mathrm{th}}$ source to be a match becomes.
    
    \item Detect the assisted sources: This is straightforward, since we know that these are the first M of the estimated sources. Denote $\mathbf{p}=[p_{1},\ldots,p_{M}]$ the vector with the indixes of the corresponding true sources associated to each one of the first $M$ assisted sources (In our specific dataset, $\mathbf{p}=[1,11,14]$). Then, the quality of the composition for the assisted sources, according to \Eq{\ref{Eq:ri}}, is given by $r_{p_{i}}=c_{p_{i},i}$ with $i=1,\ldots,M$.
    
    \item Remove the contribution of the assisted sources in the matrix $\mathbf{C}$, setting to zero their associated rows and columns, that is, $\mathbf{c}^{p_{i}}=\mathbf{0}$ and $\mathbf{c}_{i}=\mathbf{0}$ for $i=1,\ldots,M$. 
    
    \item Detection of the non-assisted sources: This can be efficiently done with the aid of matrix $\mathbf{C}$. First, we find the maximum entry of the matrix $\mathbf{C}$, say $c_{ij}$. This entry reveals the fact that the best match of the $i^{\mathrm{th}}$ true source is the $j^{\mathrm{th}}$ estimated source, consequently, $r_{i}=c_{ij}$. Then, in order to exclude the detected source, we set to zero their contribution to the rest of the matrix, that is, $\mathbf{c}^{i}=\mathbf{0}$ and $\mathbf{c}_{j}=\mathbf{0}$. This process is repeated for all the rest of the true sources, i.e., until $\mathbf{C}=\mathbf{O}$.
\end{enumerate}
After these steps, the result is a vector $\mathbf{r}$, where each component, $r_{i}$, reflects the quality of the decomposition associated to the $i^{\mathrm{th}}$ source. Then, this information can be used to quantify the performance in different ways. For example, we can take the mean of the values associated with the assisted sources only, or we can take the mean over all the values of $\mathbf{r}$ to have a general overview of the performance of the decomposition as a whole. 

\begin{figure}
	\centering
	\includegraphics[width=0.80\textwidth]{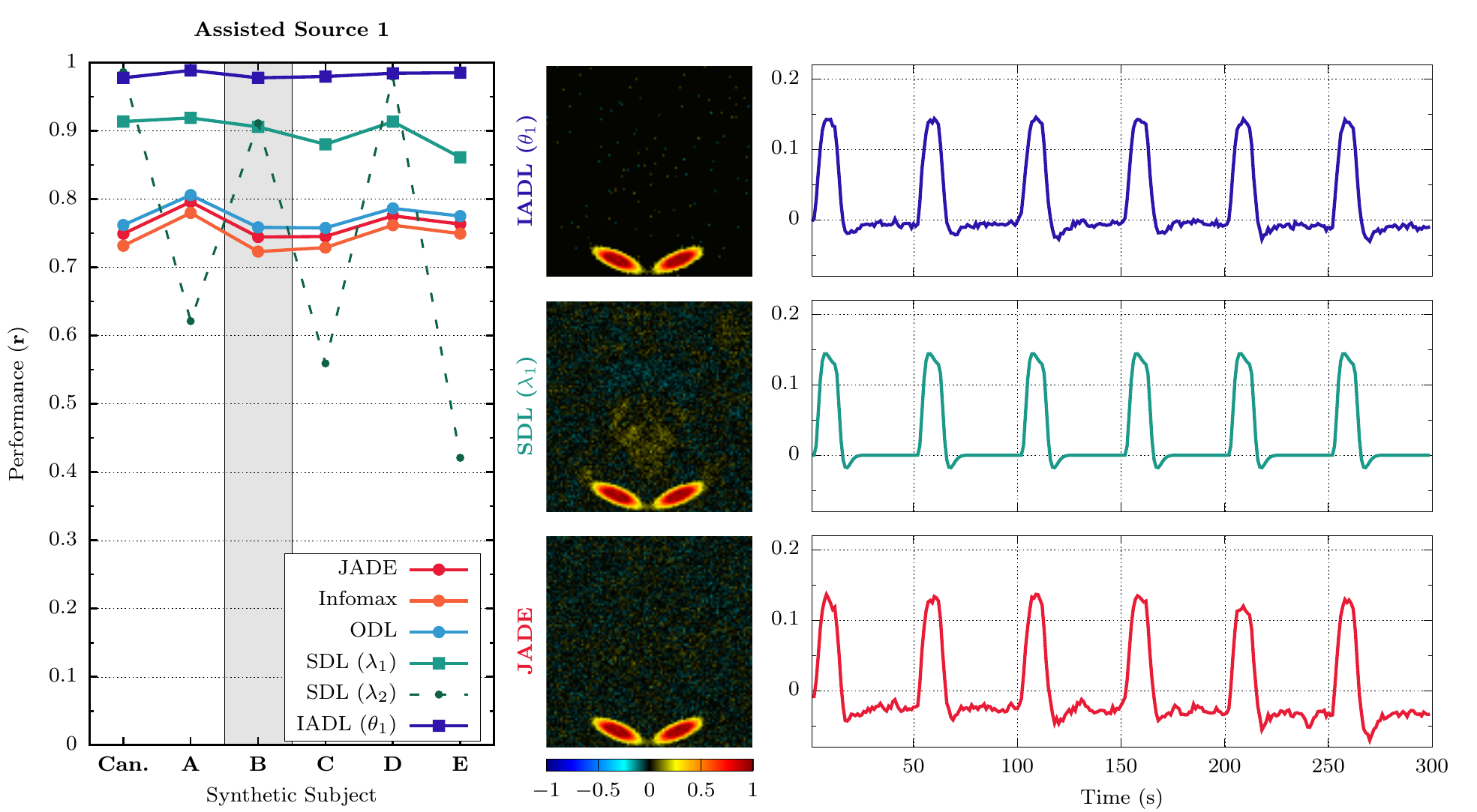}
	\includegraphics[width=0.80\textwidth]{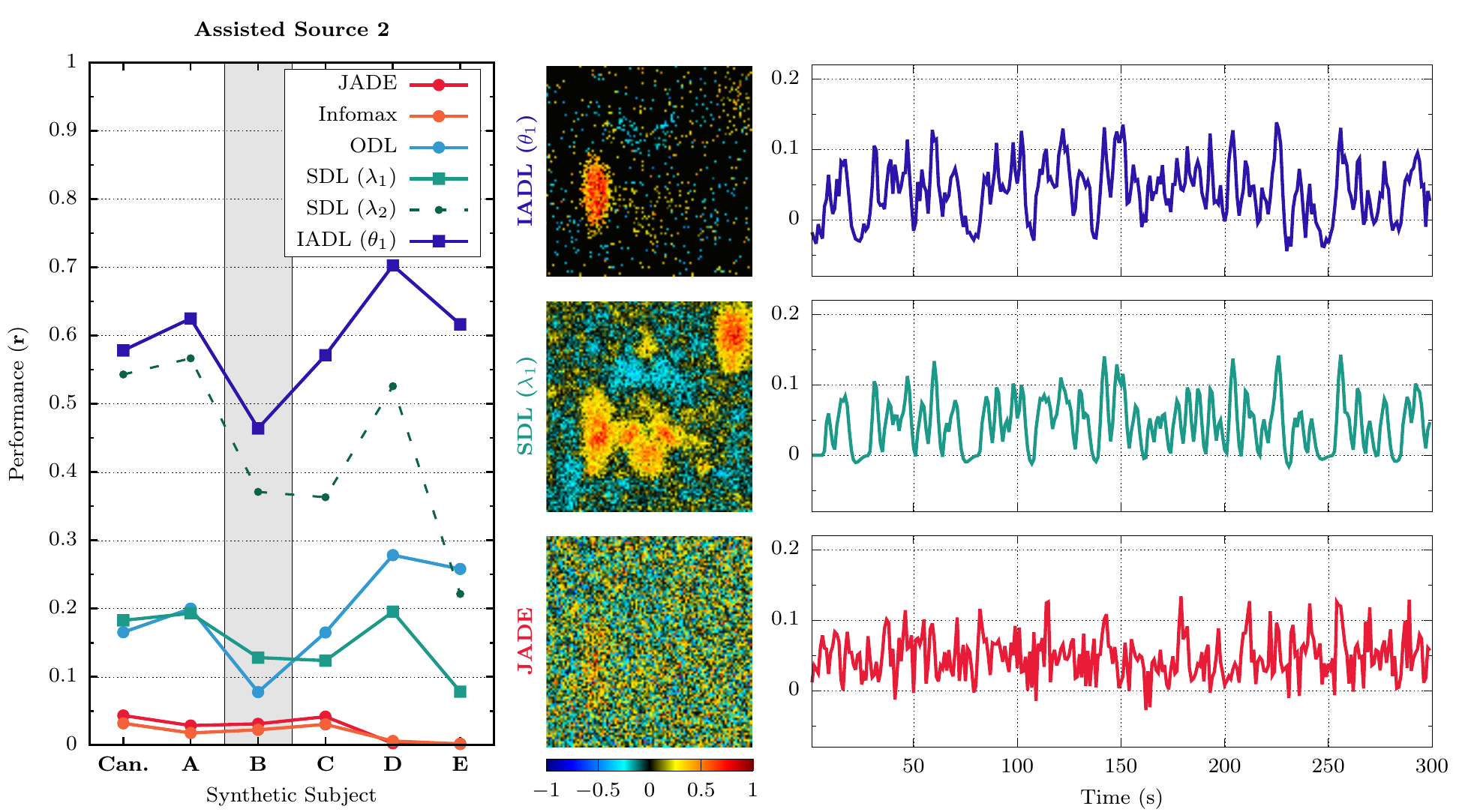}
	\includegraphics[width=0.80\textwidth]{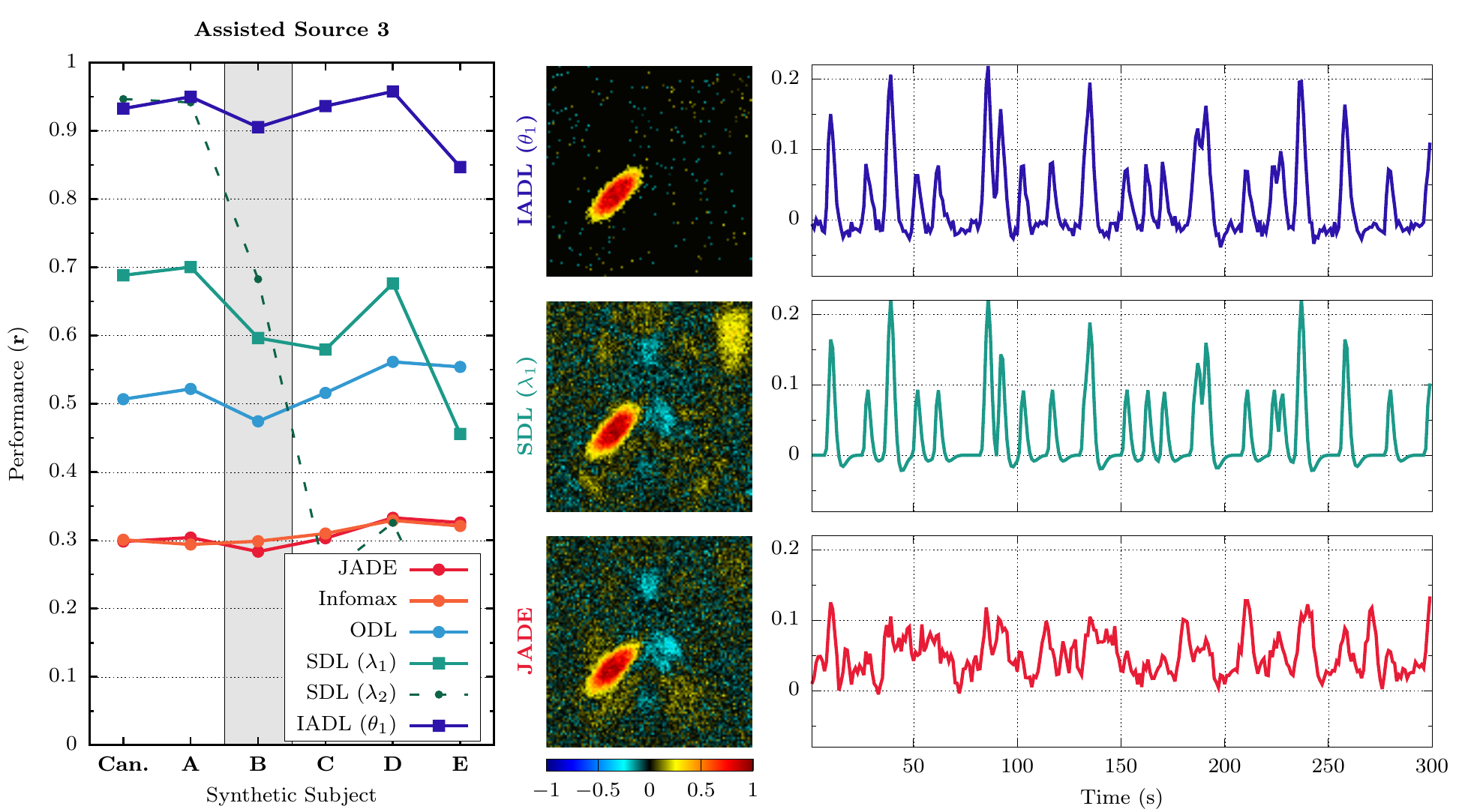}
	\caption{Individual results for the assisted soursces. The right graph shows the comparisons of all the method for each specific assisted source, which corresponds with the sources 1, 11 and 14. The left column illustrate three examples of the spatial maps and the time courses for IADL, SDL and JADE from one specif noise realization of the synthetic subject B.}
	\label{Fig:Ind_S}
\end{figure}

Apart from the full-source performance measure, the decomposition performance with respect the time courses only, i.e., with respect the estimated dictionary $\mathbf{D}$, is also of special interest. Indeed, as we discussed in the previous \Sec{\ref{Sub:Par_FBNs}}, the spatial distribution of the FBNs are commonly estimated applying certain statistical analysis over the spatial maps, which are commonly estimated via the pseudoinverse of the dictionary. Therefore, to determine the quality of the decomposition for the time course, we adopted as an alternative performance measurement:
\begin{equation}
	\hat{r}_{j}=\left(\rho(\tilde{\mathbf{d}}_{j},\mathbf{d}_{i})\right)^{2}
	\coma
	\label{Eq:ri-t}
\end{equation} 
where now $\mathbf{d}_{i}$ is the $i^{\mathrm{th}}$ obtained time course and $\tilde{\mathbf{d}}_{j}$ is its true corresponding time course. In order to compute the values for all the time courses, we followed the same procedure as described above, but changing the definition of the matrix $\mathbf{C}$ as $c_{ij}=(\rho(\tilde{\mathbf{d}}_{i},\mathbf{d}_{j}))^{2}$ instead. Similarly, the result is a vector $\hat{\mathbf{r}}$ , where each value, $\hat{r}_{i}$, reflects the quality of the decomposition of the time course of the $i^{\mathrm{th}}$ source.

\section{Additional results}
\label{Sec:Extras}

\subsection{Individual results from the synthetic fMRI analysis}
\label{Sub:Extra_Individual}
In \Sec{\ref{Sub:Exp_Synthetic}} of the main text, we present the performance results based on the synthetic dataset, where \Fig{2} summarizes the comparison results of the different studied methods. Therefore, to further investigate the synthetic results shown in \Fig{2}, we depicted the individual results of the performance comparison of the assisted sources, including the obtained time courses and spatial maps obtained by some of the different methods. Thus, the \Fig{\ref{Fig:Ind_S}} shows the individual results for each assisted source. The left graph includes the main comparison using the introduced performance measurement among sources, whereas the right part displays one specific noise realization of one particular subject, in this case, the subject B.

Observe that all the methods are capable of separating the assisted source 1 correctly, as we expected, since we selected this assisted source as an ``easy'' source example. On the other hand, the assisted sources 2 and 3 (which represents the sources 11 and 14 of the synthetic dataset \Fig{\ref{Fig:Srcs}}) explicitly require assistance to unmix them properly. Moreover, for the SDL method, we observe the same issues as in the \Fig{3} in the main text: the SDL method only works if the HRF of the subject is close enough to the cHRF. On the contrary, IADL successfully unmixes all the assisted sources and exhibits a more stable performance among the synthetic subjects.

\subsection{GLM analysis for the synthetic dataset}
\label{Sub:Extra_GLM}
For the study of GLM over the synthetic dataset, we performed a comparison between the proposed IADL and the standard GLM approach, using the software toolbox SPM12. The design matrix of SPM includes the three task-related time courses that we used for IADL and the rest of information-assisted algorithms, as it is described in \Sec{\ref{Sub:Exp_Robustness}} of the main text. \Fig{\ref{Fig:Synthetic_Prm}} shows the normalized parameter maps for the assisted sources corresponding to (a) the canonical subject and (b) the subject~E. Furthermore, for completeness, \Fig{\ref{Fig:Synthetic_zScr}} depicts the corresponding statistical parametric maps for both studied approaches with $z>1.97$ ($p<0.05$, uncorrected), respectively.\\

In both figures, we observe that SPM and IADL recover the assisted sources 1 and 3 sucessfully, similarly to the other studied algorithms, since these two synthetic sources are relatively easy to identify. However, for the assisted source 2, the result of SPM is significantly inferior to that of IADL. Specifically, the result of SPM contains considerable residual from other overlapped sources, which partially remains as false positives in \Fig{\ref{Fig:Synthetic_zScr}}, whereas the source of interest appears considerably deteriorated. On the other hand, IADL does not have residual from other sources, and the source of interest is recovered correctly.

\begin{figure}
	\centering
	\textbf{(A) Canonical Subject} \hspace{3cm} \textbf{(B) Subject E}\\
	\includegraphics[scale=0.7]{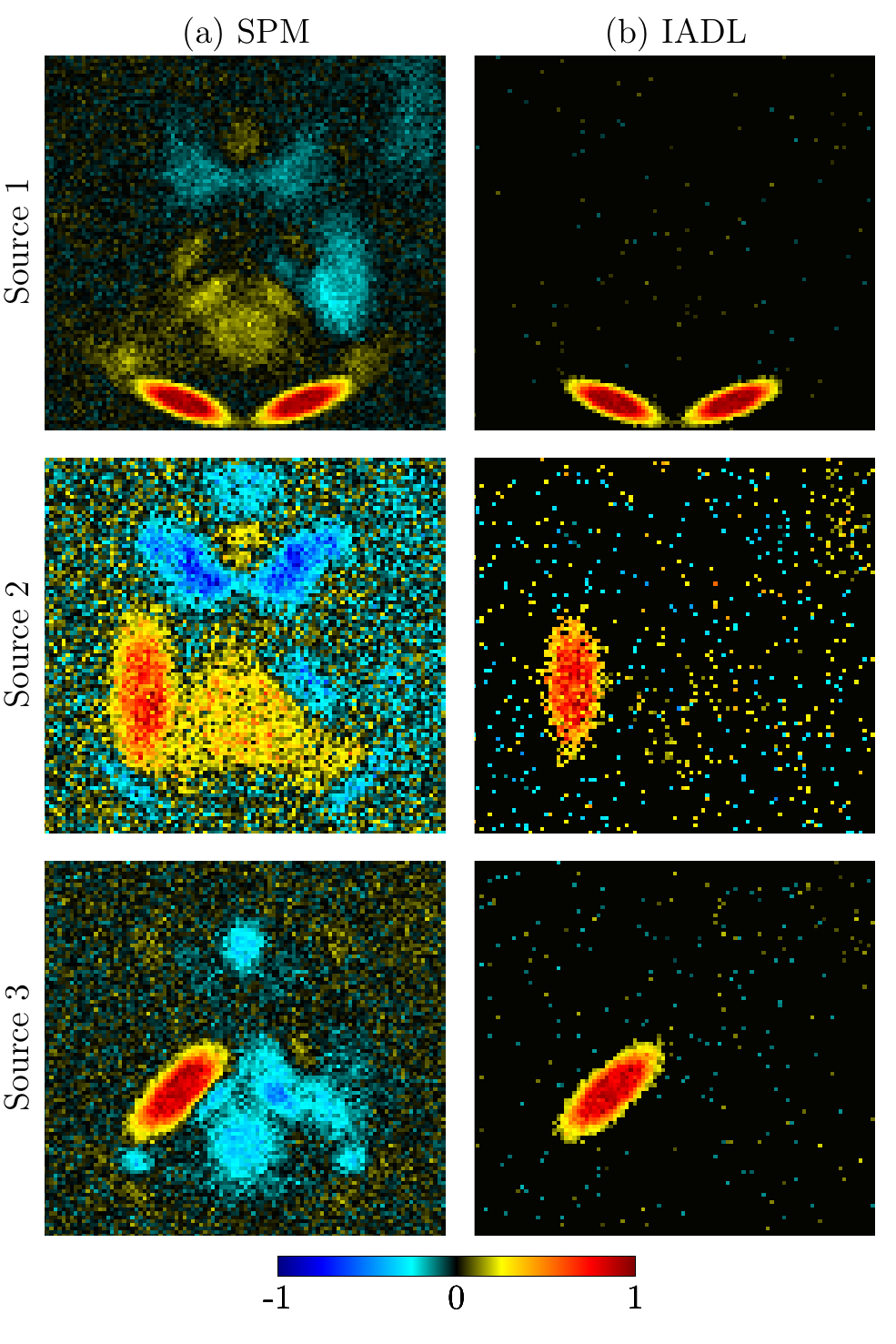}
	\includegraphics[scale=0.7]{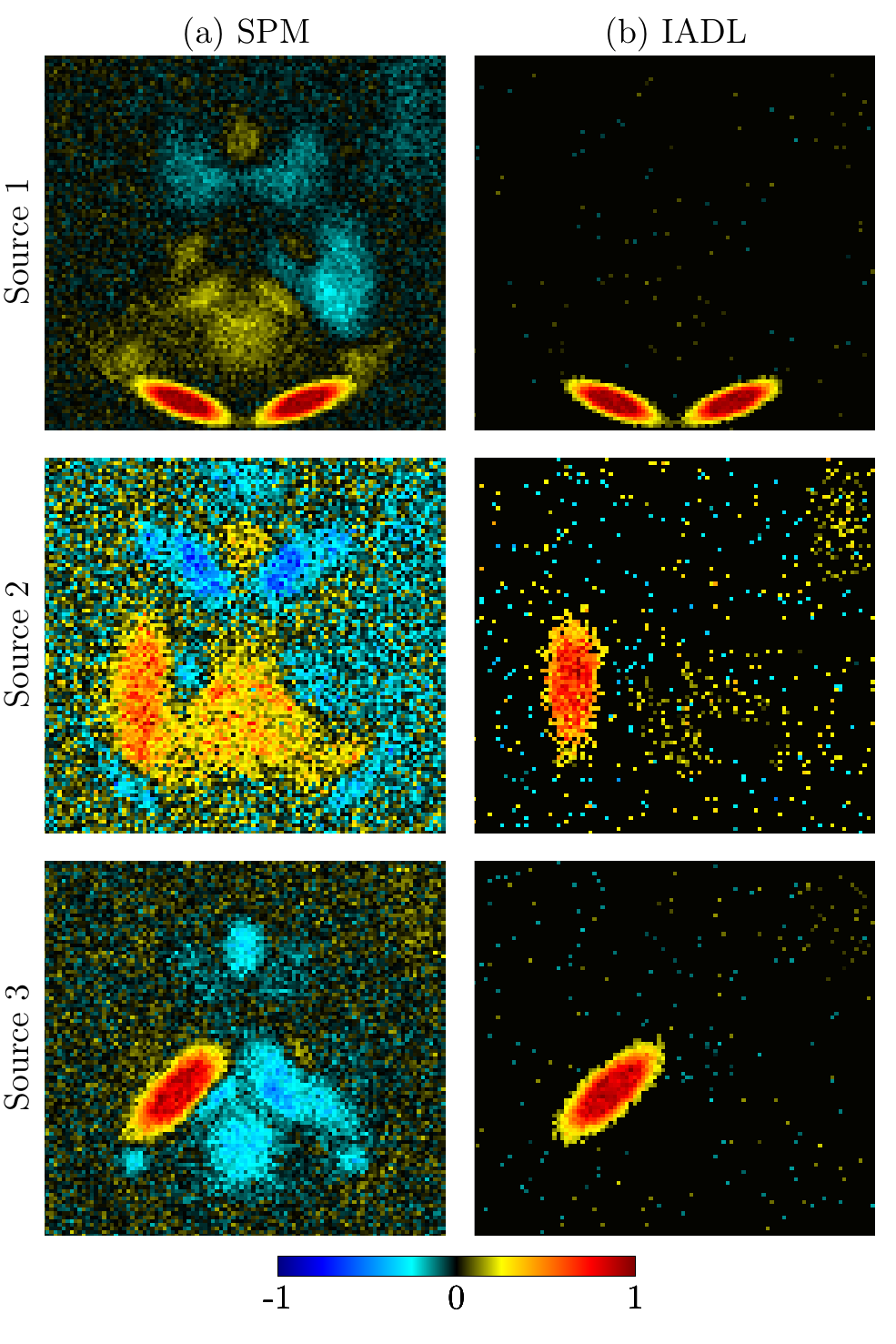}
	\caption{Comparison between the normalized parameter maps of SPM and IADL, for the canonical subject (A) and the subject E (B) from the synthetic dataset.}
	\label{Fig:Synthetic_Prm}
\end{figure}

\begin{figure}
	\centering
	\textbf{(A) Canonical Subject} \hspace{3cm} \textbf{(B) Subject E}\\
	\includegraphics[scale=0.7]{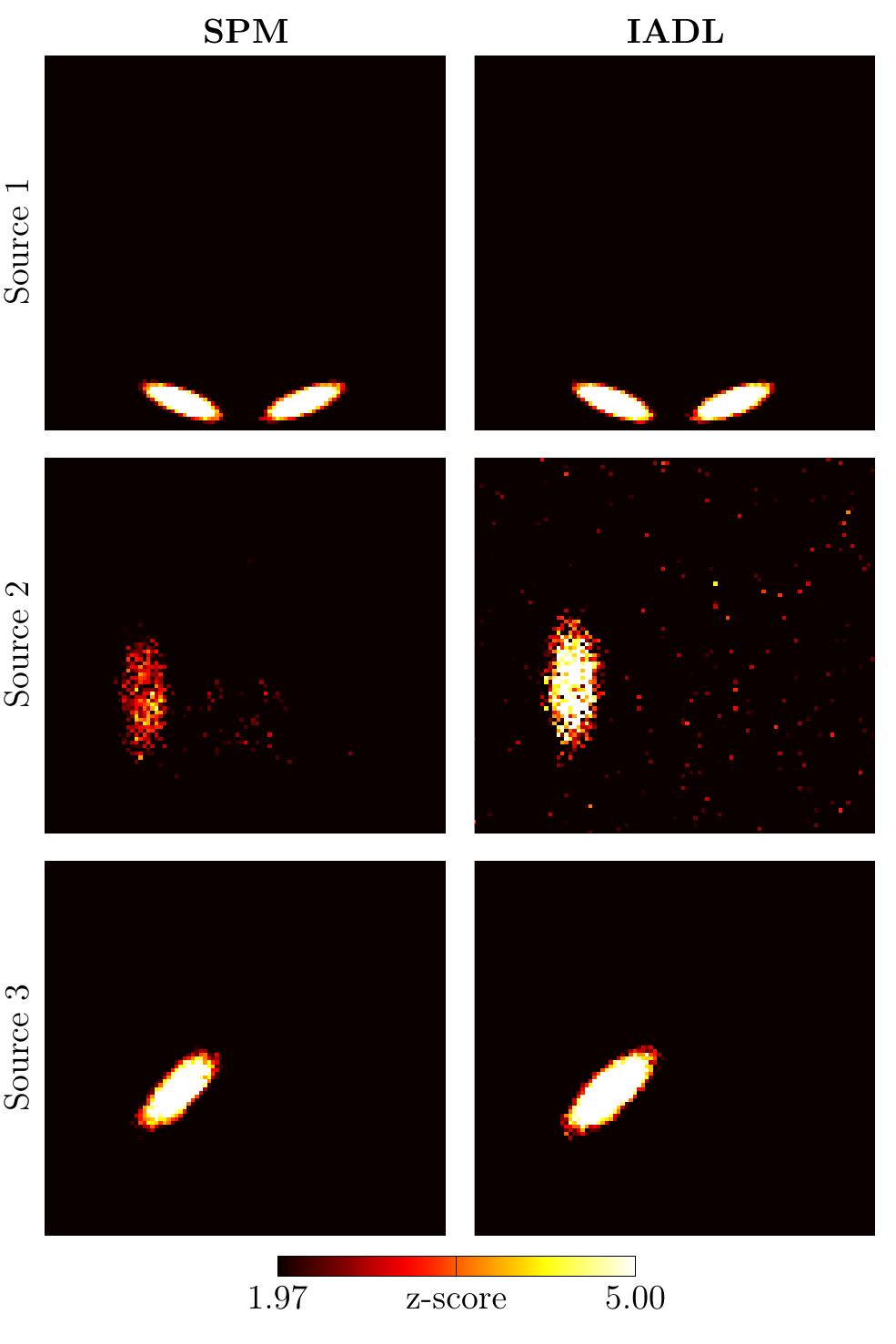}
	\includegraphics[scale=0.7]{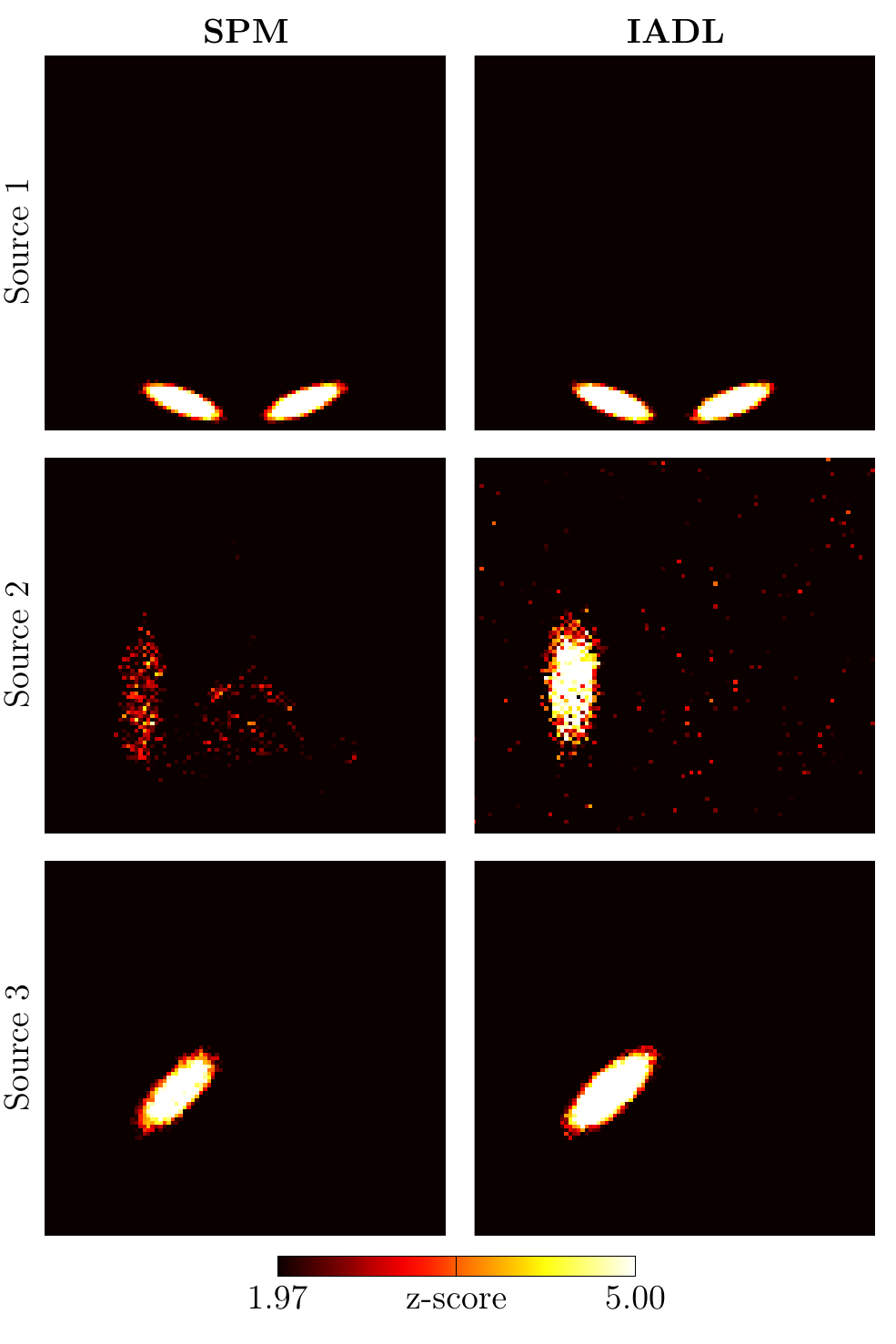}
	\caption{Significant active voxels for $z>1.97$ (uncorrected) for SPM and IADL, for the canonical subject (A) and the subject E (B) from the synthetic dataset.}
	\label{Fig:Synthetic_zScr}
\end{figure}

\subsection{Performance evaluation in function of the number of sources}
\label{Sub:Extra_EvalK}
In order to evaluate the tolerance of the proposed algorithm to the choice of the number of sources, $K$, we studied the performance of IADL for different number of sources over the synthetic and the real dataset.

\subsubsection*{Synthetic Data Analysis}
Concerning to the synthetic dataset, we studied the performance of IADL for different number of sources, $K=5,10,15,\ldots,40$, keeping the same three assisted sources with the same parameters, as we described in \Sec{\ref{Sub:Exp_Synthetic}} in the main text. For each number of sources, we implemented a different set of sparsity percentage for the free part of the dictionary following the principal guidelines described in \App{\ref{Sub:Par_Sparsity}}. \Tab{\ref{Tab:Sparsity-Srcs}.a} contains the particular values used for each number of sources. Note that the sparsity percentage for the assisted sources is the same in all the studied cases. Furthermore, the sparsity percentage for $K=25$ corresponds to the parameter $\bm{\theta}_{2}$ from \Tab{II.b} in the main text.

\begin{figure}[t]
	\centering
	{\footnotesize\textbf{I. Performance with respect to the full sources}}
	
	\includegraphics[width=0.77\textwidth]{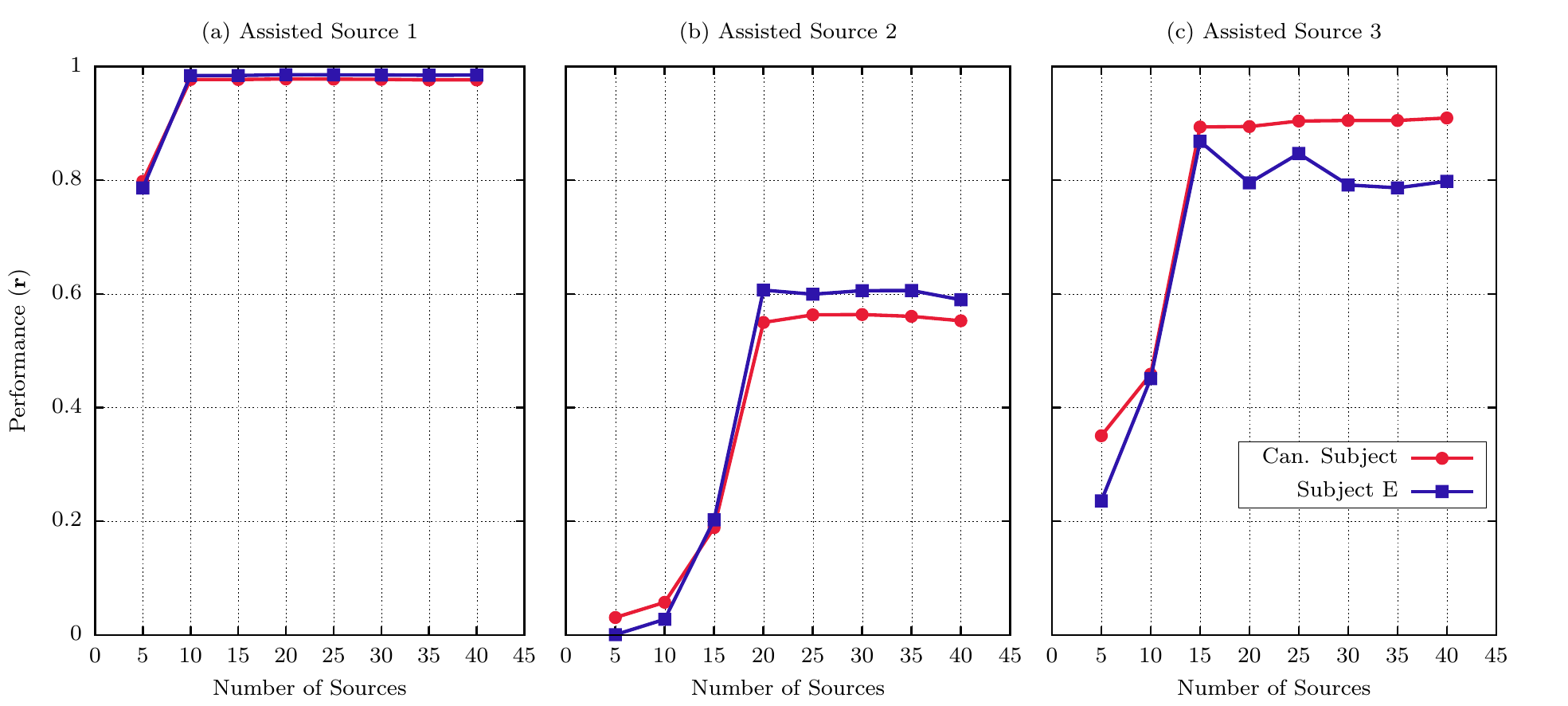}
	
	{\footnotesize\textbf{II. Performance with respect to the time courses}}
	\includegraphics[width=0.77\textwidth]{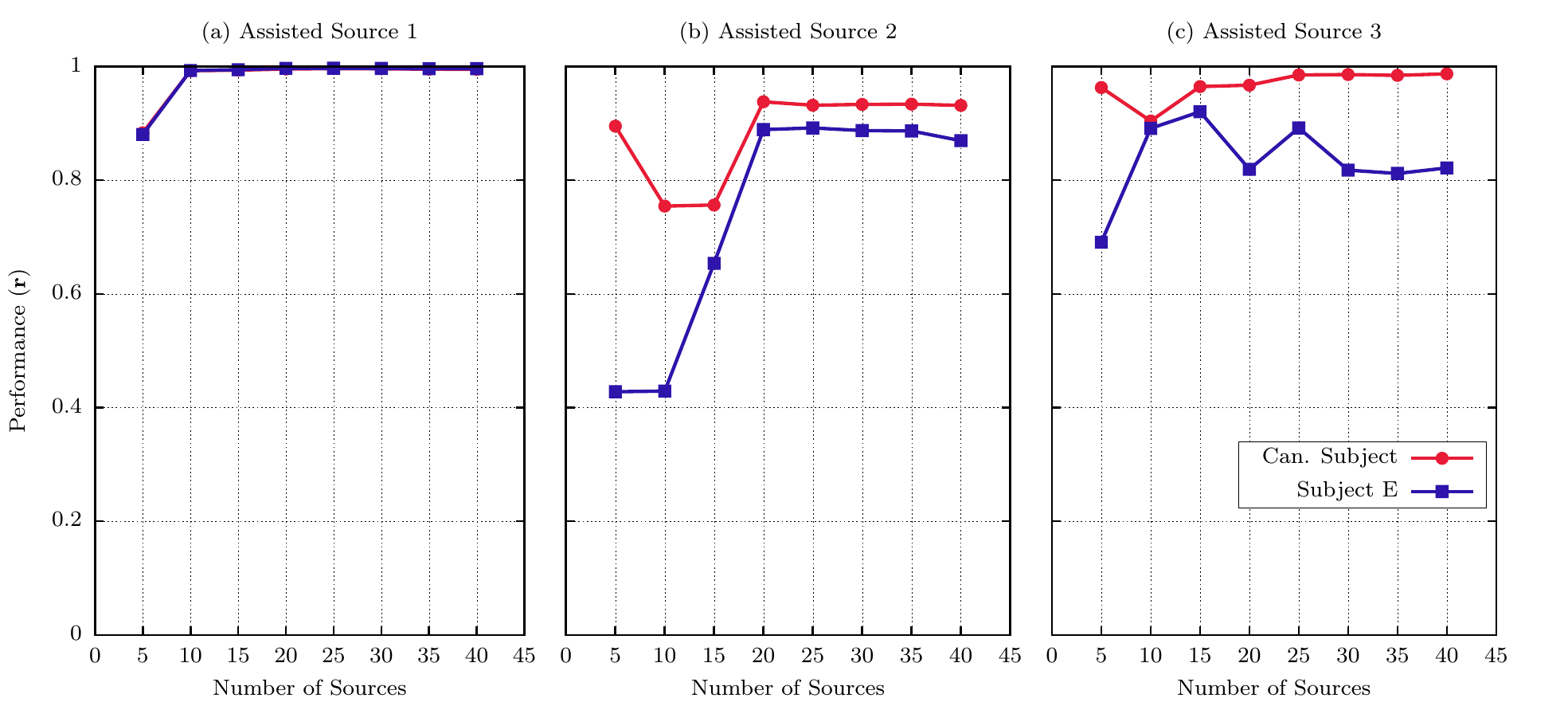}
	\caption{Performance comparison with respect the full source (I) and the time courses (II). Each subfigure presents the performance of one particular assisted source  in function of the number of sources.}
	\label{Fig:Srcs_1}
	
\end{figure}

\begin{figure}[t]
	\centering	
	{\footnotesize\textbf{I. Performance with respect to the full sources}}
	
	\includegraphics[width=0.77\textwidth]{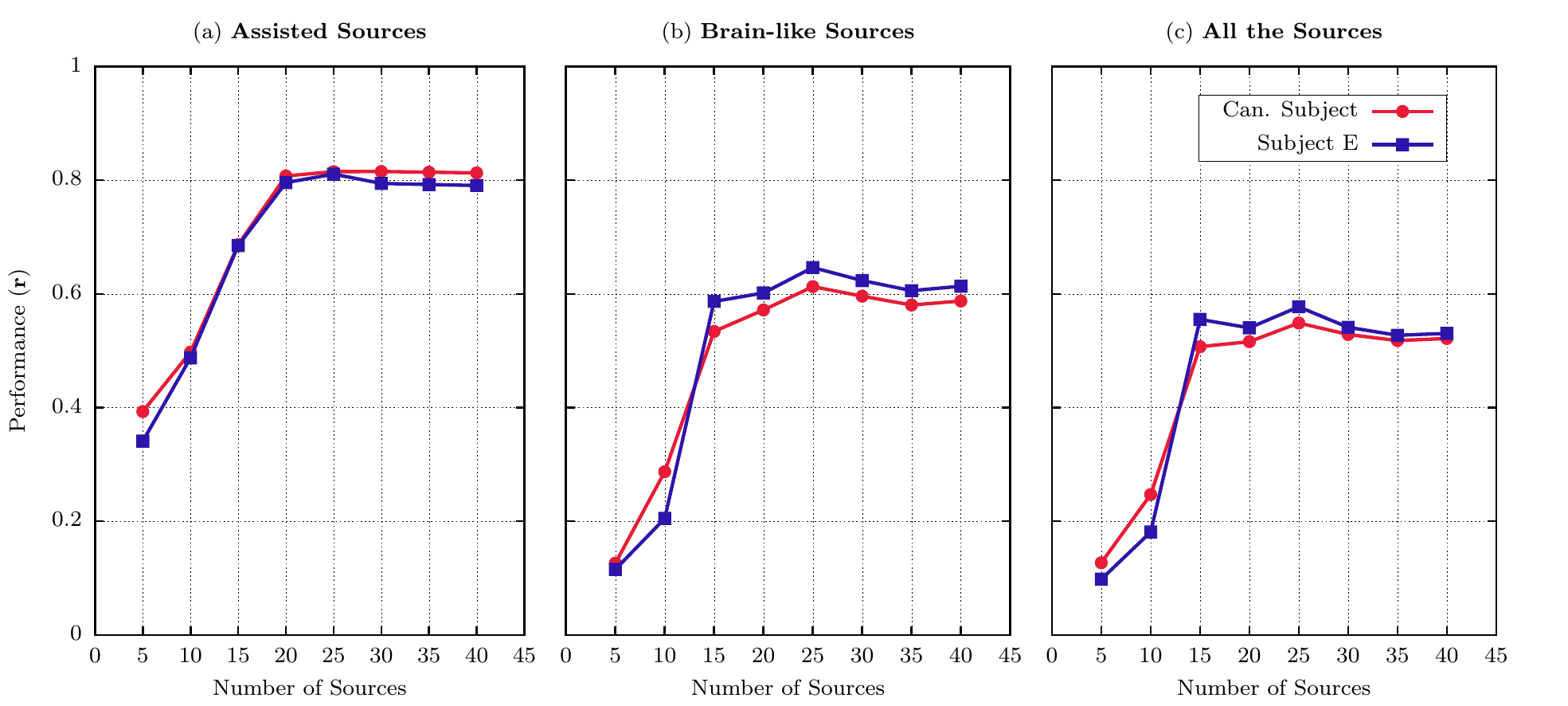}
	
	{\footnotesize\textbf{II. Performance with respect to the time courses}}
	\includegraphics[width=0.77\textwidth]{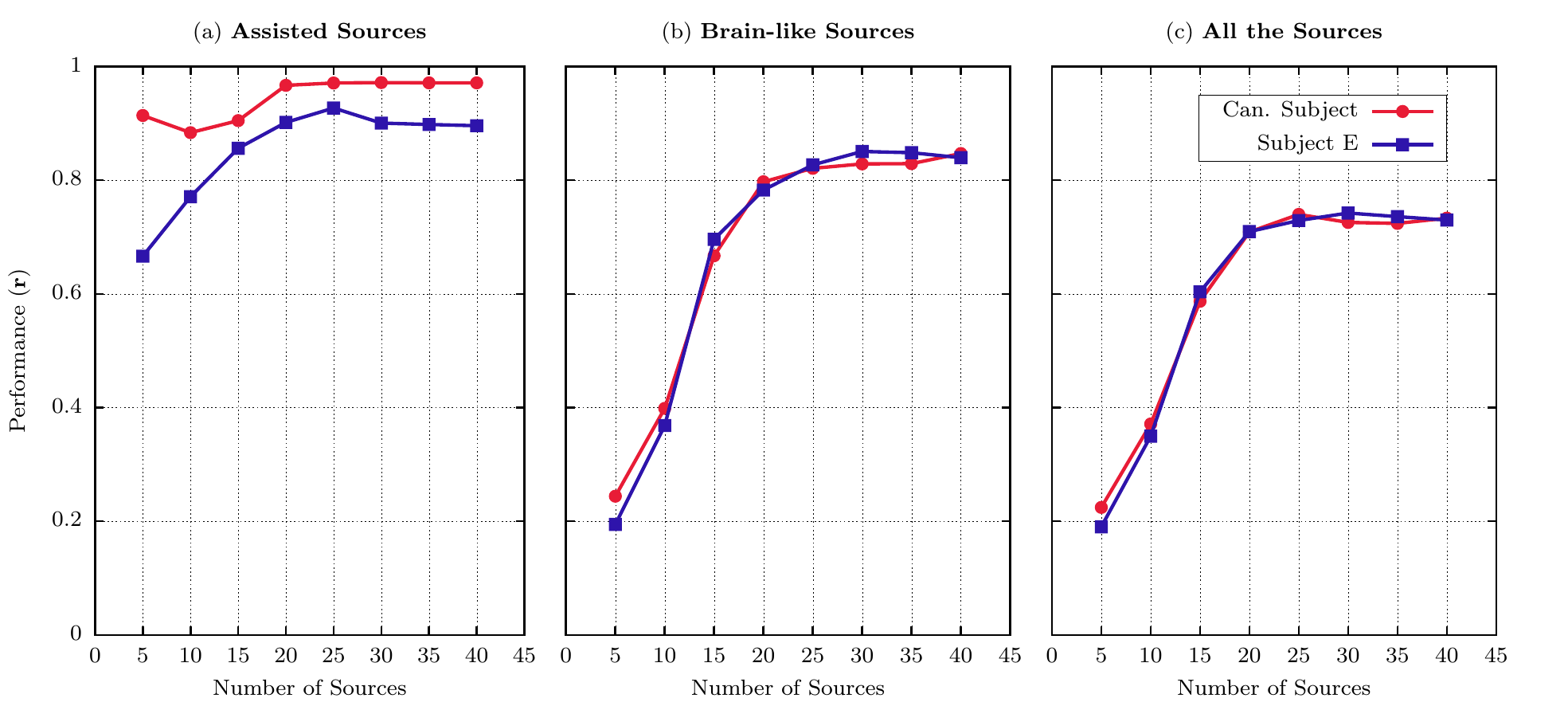}
	\caption{Performance comparison with respect to the full source (I) and the time courses (II). The sub-figures correspond to a) the assisted source, b) the brain-like sources only, and c) all sources (including artifacts) depict the performance decomposition in function of the number of sources.}
	\label{Fig:Srcs_2}
	
\end{figure}

\Fig{\ref{Fig:Srcs_1}} shows the performance comparison obtained for each specific assisted source in function of the number of sources. Similarly, \Fig{\ref{Fig:Srcs_2}} depicts the mean performance of the three assisted sources, the brain-like sources and all the sources (including artifacts) respectively. In both figures, the red line depicts the obtained results of the canonical subject and the blue line represents the results of the subject E from the synthetic dataset.

Observe that the obtained results agree with the expected behavior of the standard DL techniques: the algorithm exhibits problems to unmix the different components if the number of sources is inferior to the correct one, whereas the performance of IADL remains more or less stable for both subjects if the number of sources is overestimated.

Furthermore, we observe that IADL seems particularly robust against the overestimation of $K$; comparing the results of 35 and 40 with 20 (the correct one), we observe that all the sources are present in all the compositions and most of the sources of the overestimated cases appears filled with random noise.

\begin{table}[H]
	\centering
	\caption{Sparsity percentage used for each number of components.}
	\label{Tab:Sparsity-Srcs}
\begin{tabular}{c|llllllll}
\multicolumn{9}{c}{\textbf{(a)} \textsc{Sparsity (\%) for the Synthetic Dataset}}\tabularnewline
\tabularnewline
$K$ & 5 & 10 & 15 & 20 & 25 & 30 & 35 & 40\tabularnewline
\hline 
\hline 
\multirow{3}{*}{\rotatebox[origin=c]{90}{Assist.}} & 95 & 95 & 95 & 95 & 95 & 95 & 95 & 95\tabularnewline
 & 90 & 90 & 90 & 90 & 90 & 90 & 90 & 90\tabularnewline
 & 90 & 90 & 90 & 90 & 90 & 90 & 90 & 90\tabularnewline
\hline
 \multirow{9}{*}{\rotatebox[origin=c]{90}{Non-assisted Sources}}& 75 & 90 & 95 & 95 & 95 & 95 & $95\:{\scriptstyle (\times3)}$ & $95\:{\scriptstyle (\times5)}$\tabularnewline
 &  0 & 85 & $90\:{\scriptstyle (\times2)}$ & $90\:{\scriptstyle (\times3)}$ & $90\:{\scriptstyle (\times5)}$ & $90\:{\scriptstyle (\times5)}$ & $90\:{\scriptstyle (\times5)}$ & $90\:{\scriptstyle (\times5)}$\tabularnewline
 &    & 80 & $85\:{\scriptstyle (\times2)}$ & $85\:{\scriptstyle (\times3)}$ & $85\:{\scriptstyle (\times5)}$ & $85\:{\scriptstyle (\times5)}$ & $85\:{\scriptstyle (\times5)}$ & $85\:{\scriptstyle (\times5)}$\tabularnewline
 &    & 75 & $80\:{\scriptstyle (\times2)}$ & $80\:{\scriptstyle (\times2)}$ & $80\:{\scriptstyle (\times3)}$ & $80\:{\scriptstyle (\times4)}$ & $80\:{\scriptstyle (\times5)}$ & $80\:{\scriptstyle (\times5)}$\tabularnewline
 &    & 70 & 75 & $75\:{\scriptstyle (\times2)}$ & $75\:{\scriptstyle (\times2)}$ & $75\:{\scriptstyle (\times2)}$ & $75\:{\scriptstyle (\times3)}$ & $75\:{\scriptstyle (\times4)}$\tabularnewline
 &    & 10 & 70 & 70 & 70 & $70\:{\scriptstyle (\times2)}$ & $70\:{\scriptstyle (\times2)}$ & $70\:{\scriptstyle (\times3)}$\tabularnewline
 &    &  0 & 10 & 10 & 10 & $10\:{\scriptstyle (\times2)}$ & $10\:{\scriptstyle (\times3)}$ & $10\:{\scriptstyle (\times3)}$\tabularnewline
 &    &    & 5 & 5 & 5 & $5\:{\scriptstyle (\times2)}$ & $5\:{\scriptstyle (\times2)}$ & $5\:{\scriptstyle (\times2)}$\tabularnewline
 &    &    & 0 & $0\:{\scriptstyle (\times3)}$ & $0\:{\scriptstyle (\times3)}$ & $0\:{\scriptstyle (\times4)}$ & $0\:{\scriptstyle (\times4)}$ & $0\:{\scriptstyle (\times5)}$\tabularnewline
 \hline
\end{tabular}

\vspace{1cm}

\begin{tabular}{c|llllllll}
\multicolumn{9}{c}{\textbf{(b)} \textsc{Sparsity (\%) for the Real Dataset}}\tabularnewline
\tabularnewline
K & 10 & 15 & 20 & 25 & 30 & 35 & 40 & 60\tabularnewline
\hline 
\hline 
\multirow{6}{*}{\rotatebox[origin=c]{90}{Assisted}} & 90 & 90 & 90 & 90 & 90 & 90 & 90 & 90\tabularnewline
 & 95 & 95 & 95 & 95 & 95 & 95 & 95 & 95\tabularnewline
 & 95 & 95 & 95 & 95 & 95 & 95 & 95 & 95\tabularnewline
 & 95 & 95 & 95 & 95 & 95 & 95 & 95 & 95\tabularnewline
 & 95 & 95 & 95 & 95 & 95 & 95 & 95 & 95\tabularnewline
 & 95 & 95 & 95 & 95 & 95 & 95 & 95 & 95\tabularnewline
\hline 
\multirow{18}{*}{\rotatebox[origin=c]{90}{Non-assisted Sources}} & 95 & 95 &  $95\:{\scriptstyle (\times2)}$ &  $95\:{\scriptstyle (\times3)}$ &  $95\:{\scriptstyle (\times3)}$ &  $95\:{\scriptstyle (\times4)}$ &  $95\:{\scriptstyle (\times4)}$ &  $95\:{\scriptstyle (\times5)}$\tabularnewline
 & 70 & $90\:{\scriptstyle (\times2)}$ &  $90\:{\scriptstyle (\times3)}$ &  $90\:{\scriptstyle (\times3)}$ &  $90\:{\scriptstyle (\times3)}$ &  $90\:{\scriptstyle (\times3)}$ &  $90\:{\scriptstyle (\times4)}$ &  $90\:{\scriptstyle (\times5)}$\tabularnewline
 & 10 & 80 & 85 & $85\:{\scriptstyle (\times2)}$ & $85\:{\scriptstyle (\times3)}$ & $85\:{\scriptstyle (\times3)}$ & $85\:{\scriptstyle (\times4)}$ & $85\:{\scriptstyle (\times5)}$\tabularnewline
 & 0 & 70 &  80 &  $80\:{\scriptstyle (\times2)}$ &  $80\:{\scriptstyle (\times3)}$ &  $80\:{\scriptstyle (\times3)}$ &  $80\:{\scriptstyle (\times4)}$ &  $80\:{\scriptstyle (\times5)}$\tabularnewline
 &  & 50 & 70 & 75 & 75 & $75\:{\scriptstyle (\times3)}$ & $75\:{\scriptstyle (\times3)}$ & $75\:{\scriptstyle (\times5)}$\tabularnewline
 &  & 10 & 50 & 70 & $70\:{\scriptstyle (\times2)}$ & $70\:{\scriptstyle (\times2)}$ & $70\:{\scriptstyle (\times3)}$ & $70\:{\scriptstyle (\times5)}$\tabularnewline
 &  & $0\:{\scriptstyle (\times2)}$ & 30 & 50 & 50 & 50 & 50 & 60 \tabularnewline
 &  &  & 20 & 30 & 30 & 40 & 40 & 50 \tabularnewline
 &  &  & 10 & 20 & 20 & 30 & 30 & 45\tabularnewline
 &  &  & $0\:{\scriptstyle (\times2)}$ & 10 & 10 & 20 & 20 & 40\tabularnewline
 &  &  &  & $0\:{\scriptstyle (\times3)}$ & 5 & 10 & 10 & 35\tabularnewline
 &  &  &  &  & $0\:{\scriptstyle (\times4)}$ & $5\:{\scriptstyle (\times2)}$ & $5\:{\scriptstyle (\times2)}$ & 30\tabularnewline
 &  &  &  &  &  & $0\:{\scriptstyle (\times4)}$ & $0\:{\scriptstyle (\times5)}$ & 25\tabularnewline
 &  &  &  &  &  &  &  & 20\tabularnewline
 &  &  &  &  &  &  &  & 15\tabularnewline
 &  &  &  &  &  &  &  & 10\tabularnewline
 &  &  &  &  &  &  &  & $5\:{\scriptstyle (\times5)}$\tabularnewline
 &  &  &  &  &  &  &  & $0\:{\scriptstyle (\times9)}$\tabularnewline
\hline
\end{tabular}

\end{table}

\newpage

\begin{figure}
	\centering
	\includegraphics[width=0.8\textwidth]{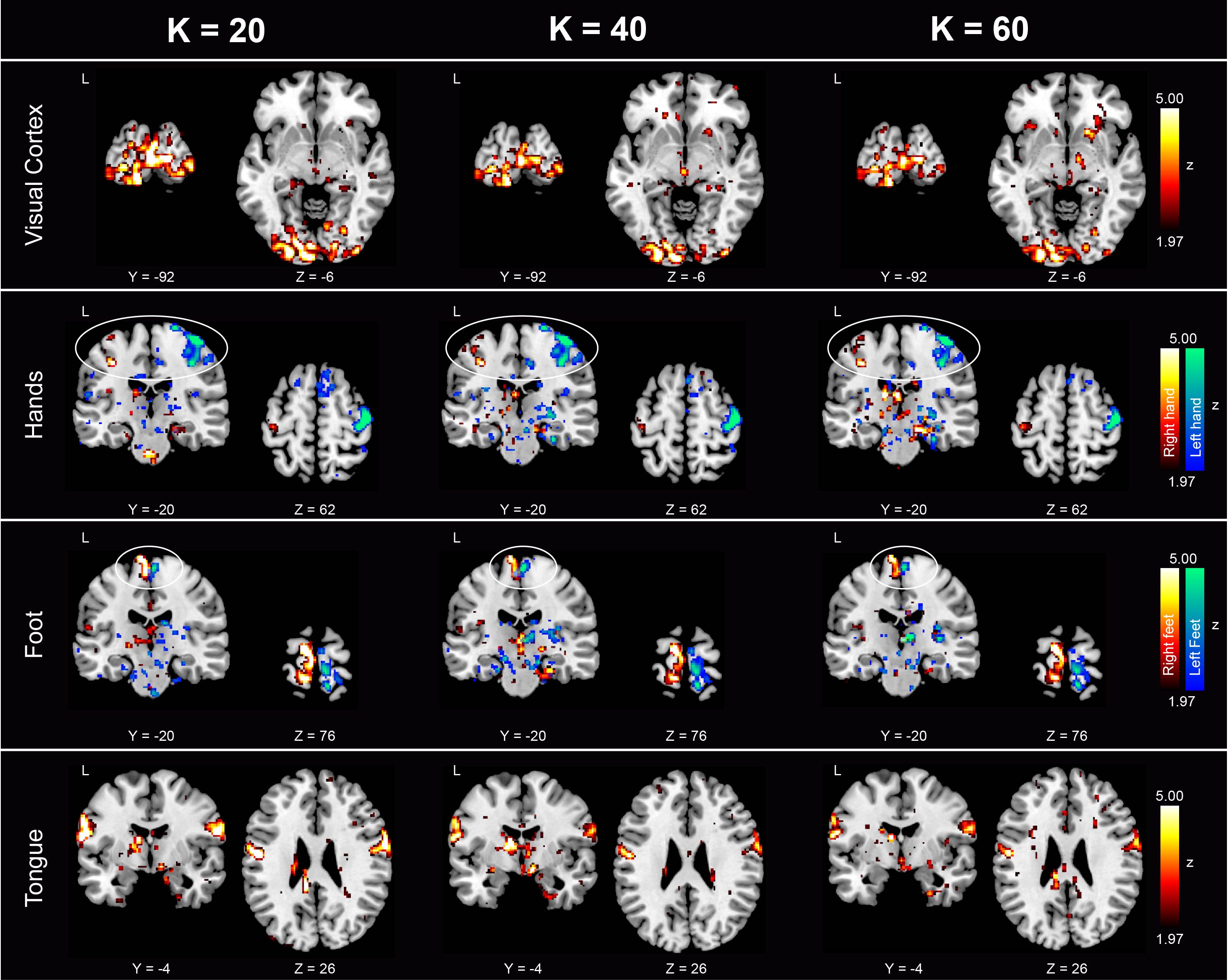}
	\caption{Significant active voxles ($z > 1.97$) for one randomly selected subject, where each column represent the obtained results for the different studied number of sources. Each row shows the obtained statistical maps for each specific task: visual, left/right hand, left/right feet and tongue.}
	\label{Fig:Rnd_Src_A}
\end{figure}

\begin{figure}
	\centering
	\includegraphics[width=0.8\textwidth]{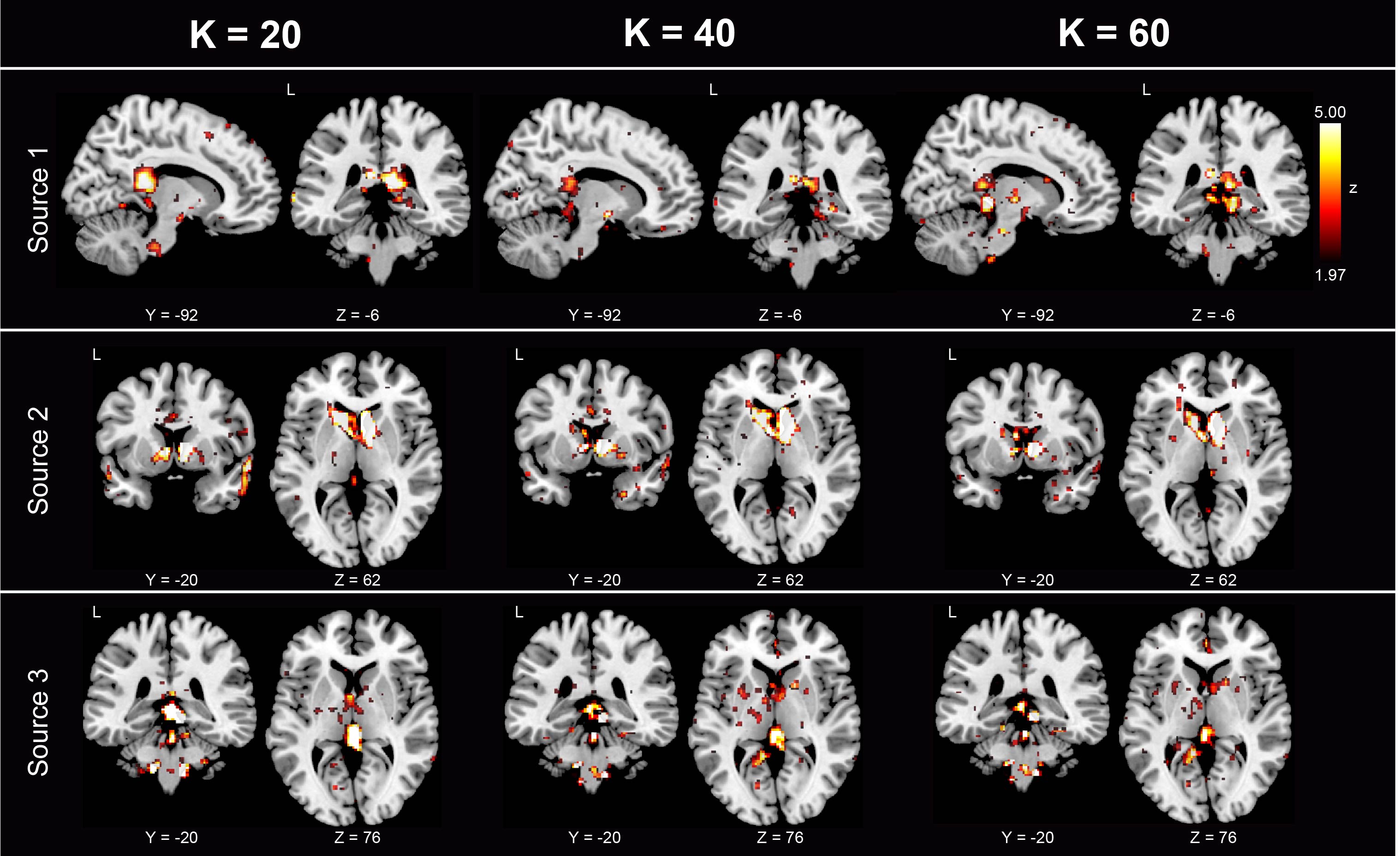}
	\caption{Significant active voxles ($z > 1.97$) for one randomly selected subject, where each column represents the obtained results for different studied number of sources. Each row contain three common components found in the free part of the dictionary.}
	\label{Fig:Rnd_Src}
\end{figure}

\subsubsection*{Real Data Analysis}
In this case, we studied the same task-related fMRI experiment used in the main text, implementing different number of sources, $K=10,15,20,25,\ldots,40$ and 60. For this analysis, we randomly selected a single subject from the studied dataset. The rest of the parameters, including the imposed task-related time courses for the six assisted sources were the same as they are described in \Sec{\ref{Sub:Exp_Real-fMRI}} of the main text. As in the synthetic experiment, we implemented a different set of sparsity parameters for the free part of the dictionary, following the guideline described in \App{\ref{Sub:Par_Sparsity}}. \Tab{\ref{Tab:Srcs}.b} depicts the particular selection of the sparsity percentage used in this experiment. The parameter for $K=20$ corresponds to the parameter $\bm{\theta}$ from \Tab{II.c}, which was implemented for the analysis of the real data in the main text.

The absence of ground truth in real data hinders a direct performance comparison in contrast to the synthetic data. Therefore, for the real data, we analyzed the consistency of the obtained results searching for common components among all the studied number of components and visually analyzing their corresponding spatial maps. First of all, for the results from $K=10$ and $K=15$, the performance of the algorithms deteriorates compared with the results observed in the main text ($K=20$). However, for the results between $K=20$ and $K=40$ we observe that all the obtained results present between 17-20 common components. Between these common components, we observed the six different spatial maps that correspond to the six different assisted sources, and the rest belong to extra common sources observed within the free part of the dictionary. For the case $K=60$ we only observed 14 common component and these results were slightly worse compared with the rest of the analysis.

For completeness, we present some of the obtained common components in \Fig{\ref{Fig:Rnd_Src_A}} and \Fig{\ref{Fig:Rnd_Src}}. \Fig{\ref{Fig:Rnd_Src_A}} depicts the results of the assisted sources for $K=20$, $K=40$ and $K=60$. Observe that IADL provides similar results independently of the number of sources. Furthermore, \Fig{\ref{Fig:Rnd_Src}} shows three common components observed in the free part of the dictionary for different number of sources. Consequently, after this study we observe that the performance of the IADL algorithm is consistently robust to overestimating the number of sources, which agrees with the expected behavior of standard DL methods.

\subsection{Spatial maps from the sparse representation}
\label{Sub:Extra_SparseRep}

The sparsity-based BSS method simultaneously provides an estimate of the dictionary and its sparse representation over the coefficient matrix. As we discussed in \App{\ref{Sub:Par_FBNs}}, an alternative way to display the spatial distribution of the detected FBNs of interest is to use the sparse spatial maps from the coefficient matrix.

Therefore, to study the reliability of the sparse spatial maps, we decided to depict the obtained results of the group analysis of the motor task-fMRI preformed in \Sec{\ref{Sec:Experiments}} of the main text, using just the obtained coefficient matrix. Thus, \Fig{\ref{Fig:RealAlt}} depicts the spatial maps of the different studied conditions. The first two columns show the sparse spatial maps of SDL and IADL from the group analysis, where we only display the values bigger than zero ($\mathbf{s}^{i}>0$). This is because theoretically, the sparsity constraint will push to zero values with null contribution to that source. Then, we normalized the maximum intensity to 1 for its visualization. The third column shows the same result as we presented in the \Fig{5.b}, where the spatial maps were obtained using a z-scores. We included this extra column to facilitate a visual comparison.

Observe that, although SDL and IADL are sparsity-based approaches, the results obtained are entirely different: first, SDL present an overwhelming ratio of false positives, particularly for the hands and foot tasks, which makes impossible to infer any interpretable result. On the contrary, IADL presents more clear results, even comparable to those obtained using $z$-statistical tests. 

\begin{figure}
	\centering
	\includegraphics[width=0.90\textwidth]{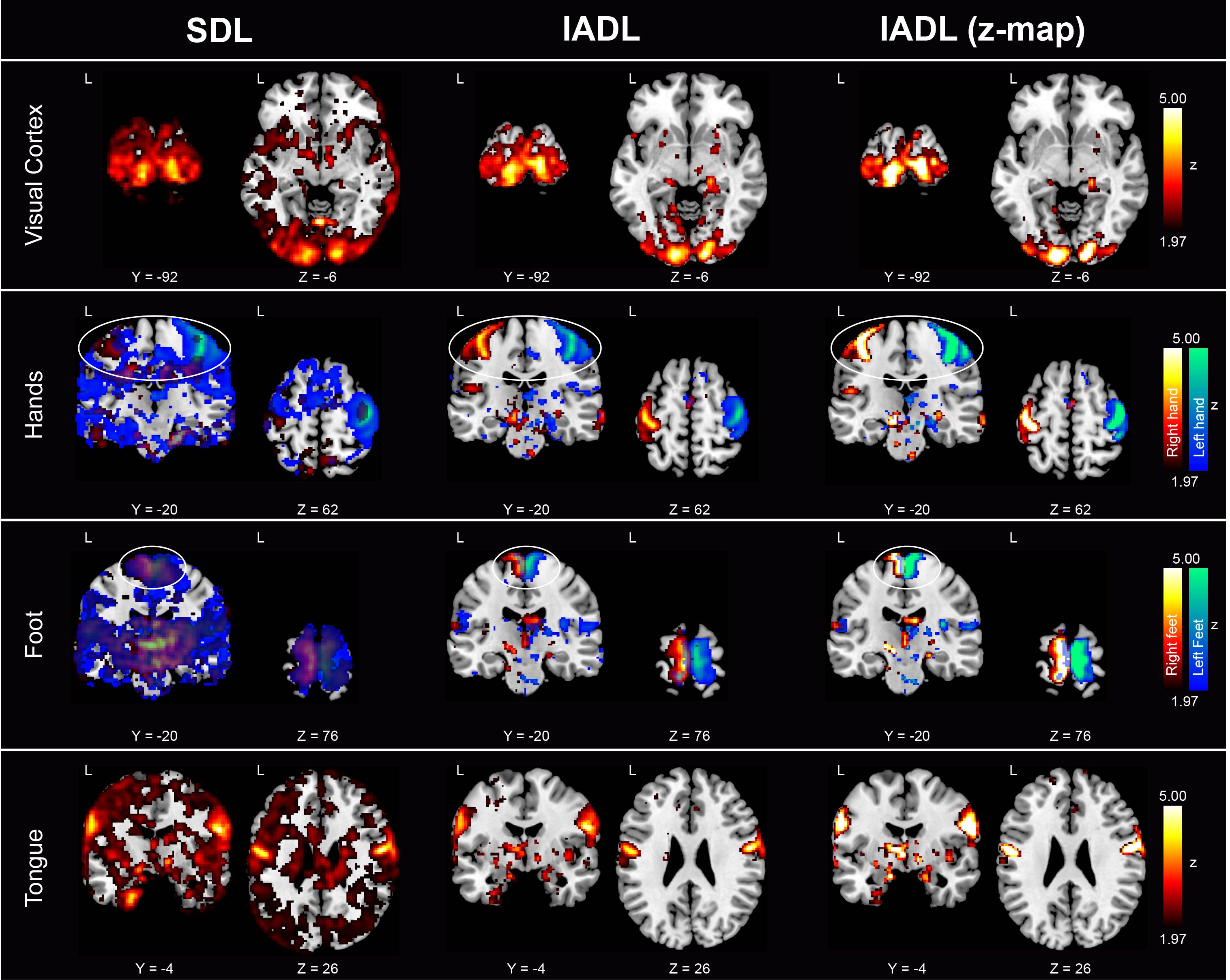}
	\caption{Comparison between the obtained sparsity map for the group analysis. The first and the second columns represent shows the positive values ($\mathbf{s}^{i}>0$) of the spatial maps of SDL and IADL respectively. The third columns shows the same results depicted in \Fig{5} of the main text as reference.}
	\label{Fig:RealAlt}
\end{figure}

\subsection{Results from the single subject analysis}
\label{Sub:Extra_Single}

For completeness, \Fig{\ref{Fig:RandomSubject}} depicts the obtained significant activation maps for one randomly selected subject for all the studied matrix factorization method. This Figure complements the results depicted in \Fig{5} in the main text, which contains the results of the group analysis.

\begin{figure}
	\centering
	\includegraphics[width=0.9\textwidth]{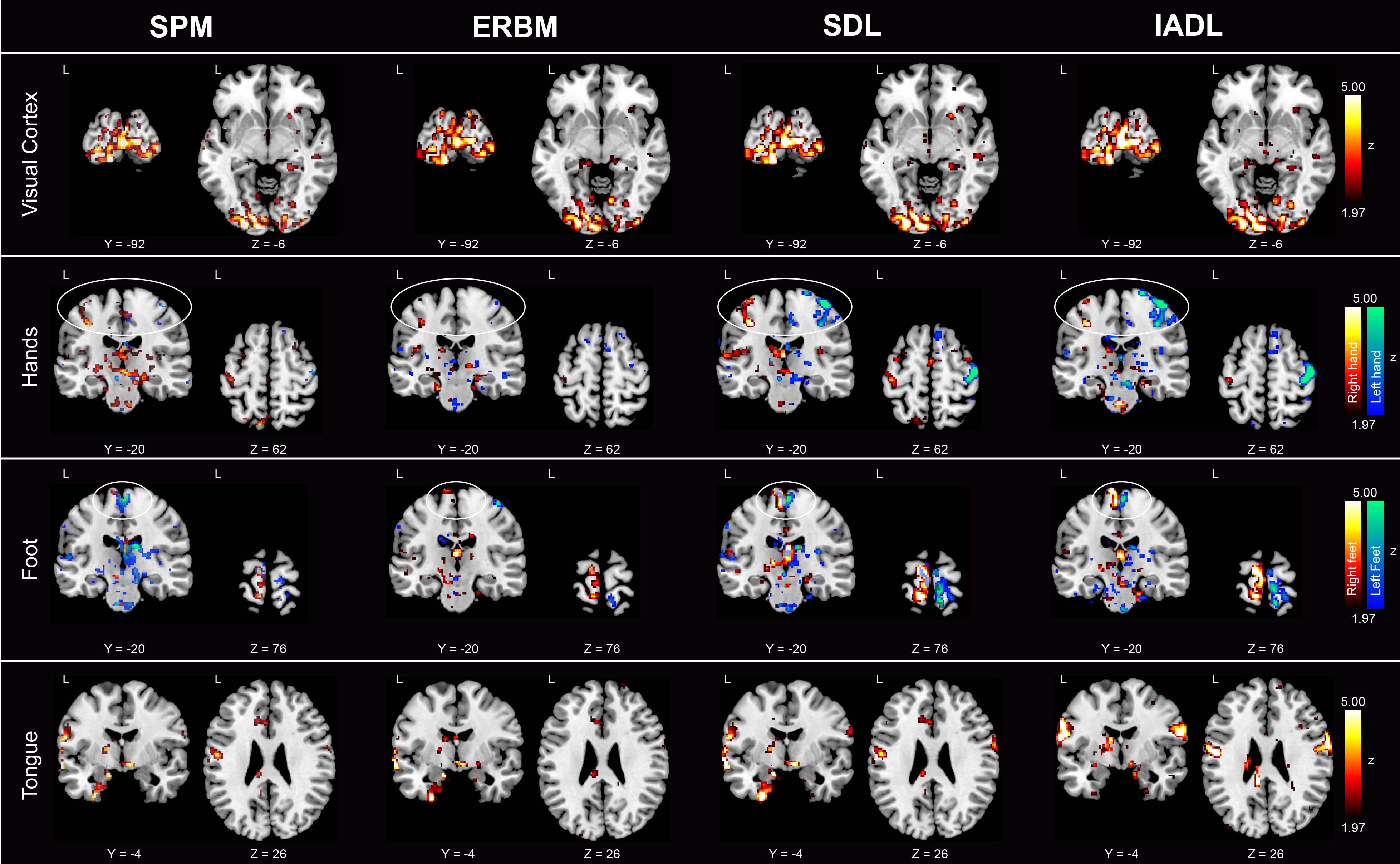}
	\caption{Significant active voxles ($z > 1.97$) for one randomly selected subject. Each row shows the most representative
positions for each specific task: visual, left/right hand, left/right feet and tongue}
	\label{Fig:RandomSubject}
\end{figure}

\subsection{Extra results from the comparison between IADL and SPM}
\label{Sub:Extra_CompSPM}
\Fig{\ref{Fig:All_ROIs}} within the main text shows a comparison between IADL and SPM. That figure is displayed between a lower and higher threshold. We specifically selected this particular interval to help to visualize what type of activation would be present at a specific threshold, in the same way as it is described in \cite{Motor-ROIs}.

Therefore, to further evaluate the performance of both approaches as well as to evaluate the presence of potential false positives, \Fig{\ref{Fig:All_Bonf}} shows the significant active voxel after applying Bonferroni correction, which is considered a a conservative correction. Observe that all the majority of false positive detected in \Fig{\ref{Fig:All_ROIs}} disappeared after the correction, as it was expected. Interestingly, SPM still display some significant voxels within the expected region of interest after this correction, whereas IADL still produces a good visualization of the correct region of interest. Thus, \Fig{\ref{Fig:All_Rend}} displays the significant active voxels after the Bonferroni correction rendered over the cortex. Observe that SPM still haves some small voxels within the regions of interest, pecesiely, over the left hemisphere, whereas IADL consistently depicts the main activation areas expected for each movement type \cite{Cortex-Yeo, Cerebellum-Buckner}.

\begin{figure}[H]
	\centering
	\includegraphics[width=0.8\textwidth]{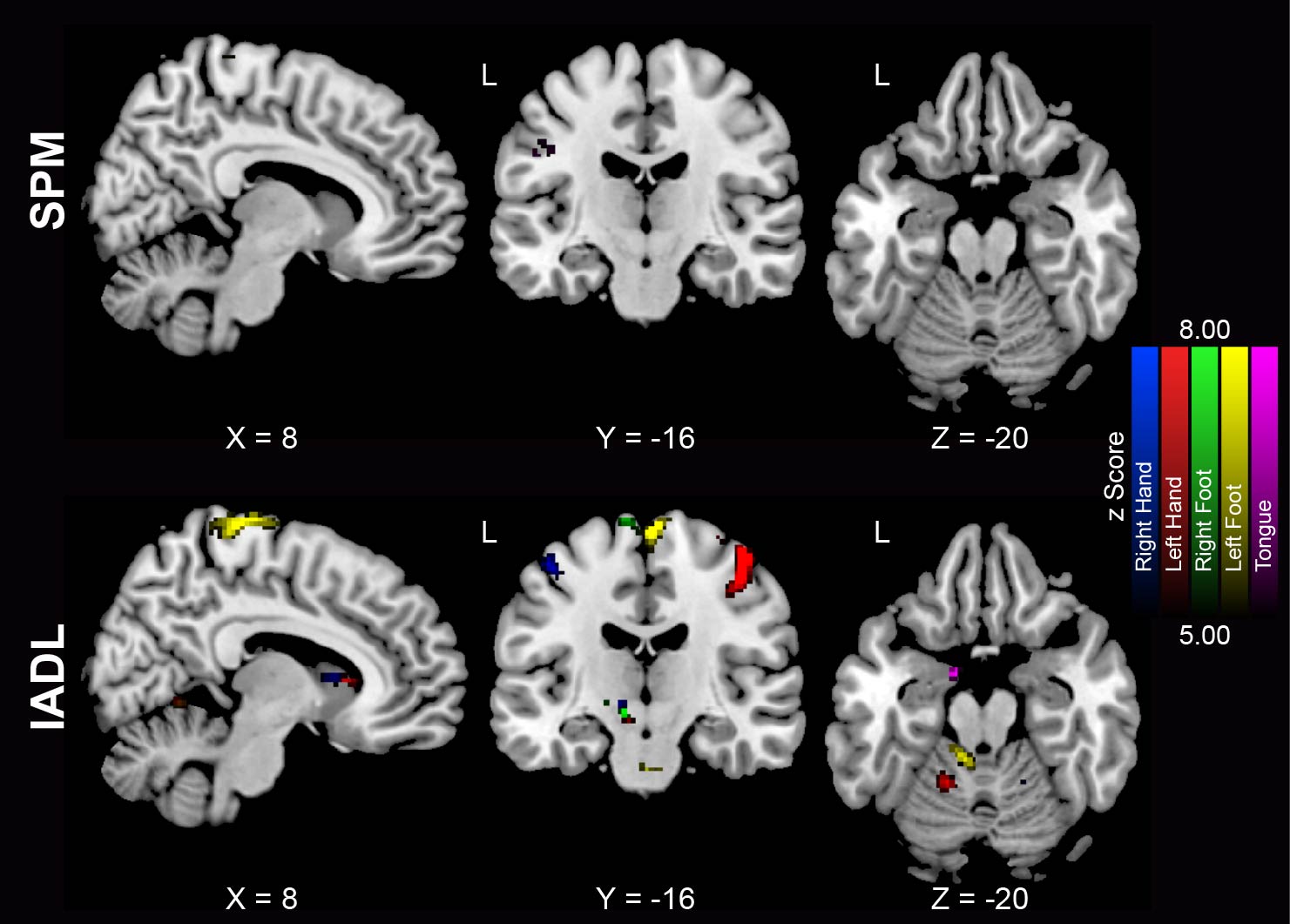}
	\caption{Significant active voxles $z > 5.00$ (Bonferroni-corrected at $p>0.066$) from the group parametric maps for SPM and IADL.}
	\label{Fig:All_Bonf}
\end{figure}

\begin{figure}[H]
	\centering
	\includegraphics[width=0.8\textwidth]{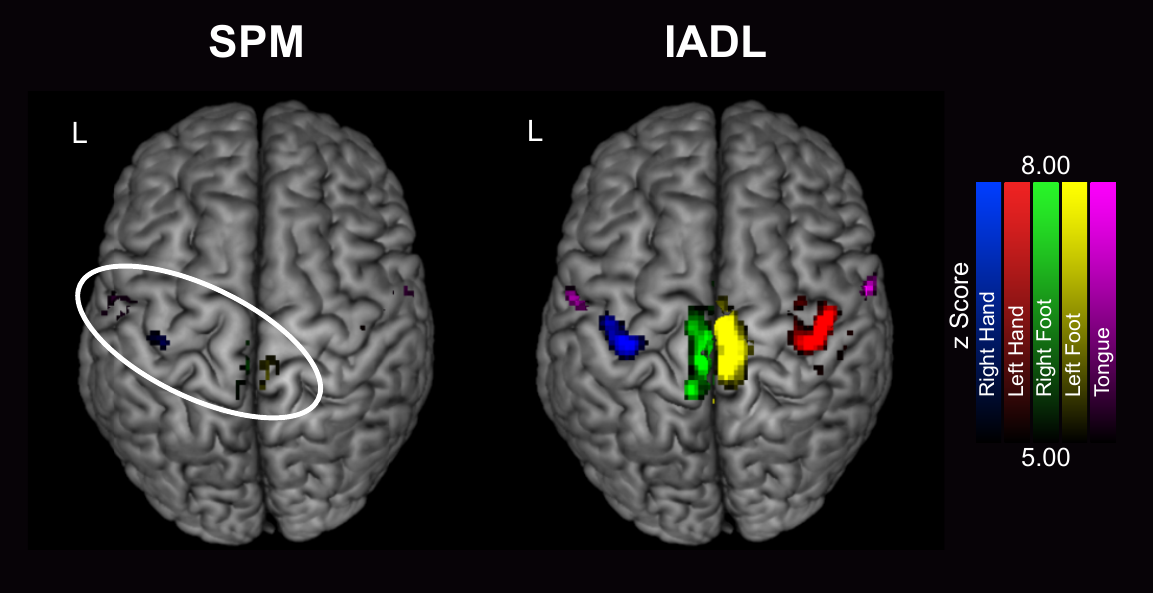}
	\caption{Rendered significant active voxles $z > 5.00$ (Bonferroni-corrected at $p>0.066$) from the group parametric maps for SPM and IADL.}
	\label{Fig:All_Rend}
\end{figure}

\newpage

\section{Mathematical deviation of the Proposed Algorithm}
\label{Sec:Mathematical}
In this section, we analytically provide the full mathematical derivation of the propose Information Assisted Dictionary Learning (IADL).

\subsection{Notation}
A lower case letter, $x$, denotes scalars, a bold capital letter, $\mathbf{X}$, denotes a matrix, and a bold lower case letter, $\mathbf{x}$, denotes a vector with its $i^{\mathrm{th}}$ component denoted as $x_{i}$. The $i^{\mathrm{th}}$ row and the $i^{\mathrm{th}}$ column of a matrix, $\mathbf{X}\in\mathbb{R}^{M\times N}$, are represented as $\mathbf{x}^{i}\in\mathbb{R}^{1\times N}$ and $\mathbf{x}_{i}\in\mathbb{R}^{M\times 1}$, respectively. Moreover, $x_{ij}$ denotes the element ``located'' at row $i$ and column $j$ of the matrix $\mathbf{X}$. The vector $\mathbf{1}_{N}$ denotes a column vector of size $N$ having all its components equal to one and $\mathbf{I}_{N}$ is the identity matrix of size $N\times N$. The notation $\mathbf{A}=\text{diag}(\mathbf{x})$ stands for the diagonal matrix having $a_{ii}=x_{i}$. Eventually, given an arbitrary vector, $\mathbf{x}\in\mathbb{R}^N$, the notation $\mathbf{x}>a$ stands for $x_{i}>a\,\forall i=1,2,\ldots,N$.

\subsection{Optimization Task}
\label{Sub:Math_Opt}
In this paper, we are focused on the study the matrix factorization problem:
\begin{equation}
	\mathbf{X}\approx\mathbf{D}\mathbf{S}
	\coma
\end{equation}
where, $\mathbf{D}\in\mathbb{R}^{T\times K}$, is the dictionary and, $\mathbf{S}\in\mathbb{R}^{K\times N}$, is the coefficient matrix as we presented them in the main text. In general, this matrix factorization problem can be seen as a constrained optimization task as follows:
\begin{equation} \label{Eq:MainOpt}
	 (\hat{\mathbf{D}},\hat{\mathbf{S}})=\underset{\mathbf{D},\mathbf{S}}{\text{argmin }}\left\Vert \mathbf{X}-\mathbf{D}\mathbf{S}\right\Vert _{F}^{2}\;\text{ s.t. }\;\begin{array}{c}
\mathbf{D}\in\mathfrak{D}\\
\mathbf{S}\in\mathfrak{L}
\end{array}
\coma
\end{equation}
where, $\mathfrak{D}$, $\mathfrak{L}$, are two sets of admissible constraints. 

According to the nature of the studied problem, we set $\mathfrak{D}=\mathfrak{D}_{\delta}$ and $\mathfrak{L}=\mathfrak{L}_{w}$, where those two convex set (see convexity proofs in \App{1}) were defined as:
\begin{equation}
	\mathfrak{D}_{\delta}=\left\{ \mathbf{D}\in\mathbb{R}^{T\times K}\;\left|\begin{array}{lcl}
\left\Vert \mathbf{d}_{i}-\bdelta_{i}\right\Vert _{2}^{2}\leqslant c_{\delta} &  & i=1,\ldots,M\\
\left\Vert \mathbf{d}_{i}\right\Vert _{2}^{2}\leqslant c_{d} &  & i=M+1,\ldots,K
\end{array}\right.\right\} 
\coma
\label{Eq:SetDic}
\end{equation}
where $\left\Vert \,\cdot\,\right\Vert _{2}$ denotes the Euclidean norm, $\mathbf{d}_{i}$ is the $i^{\mathrm{th}}$ column of the dictionary $\mathbf{D}$ and $\bdelta_{i}$ is the $i^{\mathrm{th}}$ a priori selected task-related time course and $c_{\delta}$, $c_{d}$ are two positive user-defined constants.
\begin{equation}
	\mathfrak{L}_{w}=\left\{ \mathbf{S}\in\mathbb{R}^{K\times N}\;|\;\left\Vert \mathbf{s}^{i}\right\Vert _{1,\mathbf{w}^{i}}\leqslant\phi_{i}\quad\forall i=1,2,\ldots,K\right\} 
	\coma
\end{equation}
where $\left\Vert \cdot\right\Vert _{1,\mathbf{w}}$ denotes the weighted $\ell_{1}$-norm, $\mathbf{w}^{i}$ is the vector of weights that corresponds to the vector $\mathbf{s}^{i}$ and $\phi_{i}$ is the expected number of non-zeros of the $i^{\mathrm{th}}$ row of the coefficient matrix.

As an alternative to the constrained set above, we also introduce a new convex set (see proof in \App{\ref{Sub:Math_Convex}}), where we used the weighted $\ell_{1}$-norm to impose sparsity over the full coefficient matrix as:
\begin{equation}
\label{Lwcap}
	\mathfrak{L}_{W}=\left\{ \mathbf{S}\in\mathbb{R}^{K\times N}\;|\;\left\Vert \mathbf{S}\right\Vert _{1,\mathbf{W}}\leqslant\Phi\right\} 
	\coma
\end{equation}
where $\Phi$ is the maximum number of non-zero elements of the full coefficient matrix, $\mathbf{S}$, and $\mathbf{W}\in\mathbb{R}^{K\times N}$ is the matrix of the associated weights with $w_{ij}=\frac{1}{|s_{ij}|+\varepsilon}$ (the same definition that we introduced in the \Eq{6}, in the main text). The notation $\left\Vert \,\cdot\,\right\Vert _{1,\mathbf{W}}$ stands for a generalization of the weighted $\ell_{1}$-norm for matrices, which is compactly written as:
\begin{equation}
	\left\Vert \mathbf{S}\right\Vert _{1,\mathbf{W}}=\text{tr}\left[\left|\mathbf{S}\right|\mathbf{W}^{T}\right],
\end{equation}
where $|\mathbf{S}|$ returns the absolute value of each element of $\mathbf{S}$. At this point, observe that if we have a good guess about the values $\phi_{i}$, it is possible to set the sparsity level of $\mathfrak{L}_{W}$ as $\Phi=\sum_{i=1}^{K}\phi_{i}$ forcing both constrained sets to lead to the same overall sparsity of $\mathbf{S}$.

In general, the simultaneous optimization for $\mathbf{D}$ and $\mathbf{S}$ for \Eq{\ref{Eq:MainOpt}} is challenging, due to both the non-convexity and the potential large size of the data matrix. For the above reasons, we adopted the \emph{Block Majorized Minimization} (BMM) rationale, which provides a powerful framework for the solution of such as optimization tasks~\cite{Conv-MM,Gen-MM}. The main strategy of the BMM consists of changing the actual cost function at each step by an approximation function called \emph{surrogate function}, which majorizes the former and simultaneously facilitates the optimization process.

For example, starting from an arbitrary set of estimates $\mathbf{D}^{[0]}$ and $\mathbf{S}^{[0]}$ at the $t^{\mathrm{th}}$ iteration, the BMM for the studied optimization task comprises into the following two steps:
\begin{align}
    \text{I.} & \quad \mathbf{S}^{[t+1]}=\min_{\mathbf{S}}\psi_{S}(\mathbf{S},\mathbf{S}^{[t]})\text{ s.t. }\mathbf{S}\in\mathfrak{L}_{w} \label{Eq:StepI}\coma\\
    \text{II.} & \quad \mathbf{D}^{[t+1]}=\min_{\mathbf{D}}\psi_{D}(\mathbf{D},\mathbf{D}^{[t]})\text{ s.t. }\mathbf{D}\in\mathfrak{D}_{\delta} \label{Eq:StepII}
    \coma
\end{align}
where, $\psi_{S}$ is the so called surrogate function of the first step, given $\mathbf{D}$ fixed to its current estimate $\mathbf{D}^{[t]}$, and $\psi_{D}$ is the corresponding surrogate function of the second step, gives $\mathbf{S}$ fixed to its current estimate $\mathbf{S}^{[t+1]}$.

In practice, finding an appropriate surrogate function --despite it may not be unique-- is a complicated task and also problem dependent. Furthermore, the surrogate function has to satisfy specific constraints \cite{Conv-MM}, to guarantee that the proposed two-step iterative minimization will converge to an optimal value of the original optimization task, as we discuss in detail bellow in \App{\ref{Sub:Math_Convergence}}.

After studying several candidates for the surrogate function \cite{Gen-MM}, we finally implemented the second order Taylor's expansion for matrices of the original cost functions to obtains their corresponding surrogate functions, motivated by two reasons: a) the Taylor's expansions guarantees that all the conditions of the surrogate functions are satisfied by construction \cite{Gen-MM} and b) the Taylor's expansion simplifies the optimization process, avoiding computationally expensive operations, e.g., matrix inversions, which are notably restrictive in this particular case due to the natural size of the fMRI data.\\

\subsubsection*{Step I -- Coefficient Update}

According to the \Eq{\ref{Eq:StepI}}, the update stage at the $t^{\mathrm{th}}$ iteration is  formulated as:
\[
	\mathbf{S}^{[t+1]}=\min_{\mathbf{S}}\psi_{S}(\mathbf{S},\mathbf{S}^{[t]})\quad\text{s.t.}\quad\mathbf{S}\in\mathfrak{L}_{w}
	\punto
\]
Regarding to the surrogate function, $\psi_{S}$, fixed to the current estimates the coefficient matrix $\mathbf{D}=\mathbf{D}^{[t]}$, we use the matrix form of the second order Taylor's expansion (see Appendix D in \cite{ConvexOpt}), to derive the following surrogate function:
\begin{equation}
	\psi_{S}(\mathbf{S},\mathbf{S}^{[t]})=\left\Vert \mathbf{X}-\mathbf{D}\mathbf{S}\right\Vert _{F}^{2}-\left\Vert \mathbf{D}\mathbf{S}-\mathbf{D}\mathbf{S}^{[t]}\right\Vert _{F}^{2}+c_{S}\left\Vert \mathbf{S}-\mathbf{S}^{[t]}\right\Vert _{F}^{2}
	\coma
\end{equation}
where $c_{S}\geqslant\left\Vert \mathbf{D}^{T}\mathbf{D}\right\Vert $ is a constant and $\left\Vert \,\cdot\,\right\Vert $ stands for the spectral norm. Observe that this current surrogate function is convex and majorizes its corresponding loss function by construction.

At this point, the BMM does not impose any specific optimization process. Then, we implement a Lagrangian relaxation to solve the current minimization task. Implementing the Lagrangian multipliers, the optimization becomes into:
\begin{equation}\label{Eq:Opt_Task_I}
	\min_{\mathbf{S}}\psi_{S}(\mathbf{S},\mathbf{S}^{[t]})+\mathcal{P}_{\gamma}(\mathbf{S})
	\coma
\end{equation}
where $\mathcal{P}_{\gamma}$ is a penalty term that depends on the weighted $\ell_{1}$-norm constraints and it is defined as:
\begin{equation}
	\mathcal{P}_{\gamma}(\mathbf{S})=\sum_{i=1}^{K}\gamma_{i}\left(\left\Vert \mathbf{s}^{i}\right\Vert _{1,\mathbf{w}_{i}}-\phi_{i}\right)
	\coma
\end{equation}
where the introduced parameters $\gamma_{i}\geqslant0$, for $i=1,2,\ldots,K$ are the Lagrangian multipliers, $K$ is the number of consider spatial maps (rows of $\mathbf{S}$), and $\phi_{i}$ is the expected maximum sparsity, that is, the maximum expected number of active voxels of the $i^{\mathrm{th}}$ spatial map.

From optimization theory, Lagrangian multipliers at the optimal point satisfy their corresponding KKT conditions, i.e.,
\begin{equation}
	\gamma_{i}\left(\left\Vert \mathbf{s}^{i}\right\Vert _{1,\mathbf{w}_{i}}-\phi_{i}\right)=0
	\coma\label{Eq:S_Lag}
\end{equation}
with $i=1,2,\ldots,K$.

Now, observe that the optimization task in \Eq{\ref{Eq:Opt_Task_I}} involves a \emph{convex} loss function. Hence, given the surrogate function at the $t^{\mathrm{th}}$ iteration, a global minimum exists.

In order to determine the global minimum, although the function is convex and we know that such a point exists, the loss function is not differentiable. Nevertheless, we can obtain the minimum at a point with zero subgradient. That is, a minimizer of the function exists if and only if the zero (matrix) belong to the subdifferential set, which is given by:
\begin{equation}
	\mathbf{0}\in\partial\left(\psi_{S}(\mathbf{S},\mathbf{S}^{[t]})+\mathcal{P}_{\gamma}(\mathbf{S})\right)
	\punto
\end{equation}
Observe that the first term is fully differentiable, and the subdifferential set becomes a singleton given by the standard derivative:
\[
    \partial\left(\psi_{S}(\mathbf{S},\mathbf{S}^{[t]})\right)=\left\{ \nabla\psi_{S}(\mathbf{S},\mathbf{S}^{[t]})\right\} =\left\{ -2\mathbf{D}^{T}\mathbf{X}+2c_{S}\mathbf{S}-2c_{S}\mathbf{S}^{[t]}-2\mathbf{D}\mathbf{D}^{T}\mathbf{S}^{[t]}\right\} 
    \coma
\]
that is,
\begin{equation}
    \partial\left(\psi_{S}(\mathbf{S},\mathbf{S}^{[t]})\right)=\left\{ 2c_{S}(\mathbf{S}-\mathbf{A})\right\} 
    \coma
\end{equation}
where $\mathbf{A}$ is matrix defined as:
\begin{equation}
    \mathbf{A}\triangleq\frac{1}{c_{S}}\left[\mathbf{D}^{T}\mathbf{X}+\left(c_{S}\mathbf{I}_{K}+\mathbf{D}^{T}\mathbf{D}\right)\mathbf{S}^{[t]}\right]
    \punto
\end{equation}
In contrast, the second term is not differentiable, and its corresponding subdifferential set is:
\begin{equation}
    \partial\mathcal{P}_{\gamma}(s_{ij})=\begin{cases}
\gamma_{i}w_{ij}\text{sign}(s_{ij}) & \text{if }s_{ij}\neq0\\
\gamma_{i}w_{ij}g_{ij}\text{ with }g_{ij}\in[-1,1] & \text{if }s_{ij}=0
\end{cases}
    \punto\label{Eq:PPg}
\end{equation}
However, the imposed Lagrangian multipliers address the definition of this set. Therefore, the task becomes to find the matrix $\mathbf{S}$  in the subdifferential set that simultaneously satisfies
\[
    2c_{S}(\mathbf{S}-\mathbf{A})+\partial\mathcal{P}_{\gamma}(\mathbf{S})=\mathbf{0}
    \coma
\]
and its corresponding KKT conditions \Eq{\ref{Eq:S_Lag}}, imposed by the Lagrangian multipliers.

Thus, following standard arguments and after some algebraic manipulations, it can be demonstrated that the solution of this task is given by:
\begin{equation} \label{Eq:Sol_s}
    \mathbf{s}^{i}=\begin{cases}
\mathbf{a}^{i} & \text{if }\left\Vert \mathbf{a}^{i}\right\Vert _{1,\mathbf{w}^{i}}\leqslant\phi_{i}\\
\mathcal{P}_{B_{\ell_{1}}[\mathbf{w}^{i},\phi_{i}]}(\mathbf{a}^{i}) & \text{if }\left\Vert \mathbf{a}^{i}\right\Vert _{1,\mathbf{w}^{i}}>\phi_{i}
\end{cases}
    \coma
\end{equation}
with $i=1,2,\ldots,K$, where $K$ is the total number of sources, $\phi_{i}$, is the expected maximum number of active voxels of the $i^{\mathrm{th}}$ source, $\mathbf{s}^{i}$ and $\mathbf{a}^{i}$ are the $i^{\mathrm{th}}$ row of the matrix $\mathbf{S}$ and $\mathbf{A}$ respectively, the vector $\mathbf{w}^{i}$ is the associated vector of weights, and $\mathcal{P}_{B_{\ell_{1}}[\mathbf{w}^{i},\phi_{i}]}$ is the projection operator over the weighted $\ell_{1}$-norm ball, $B_{\ell_{1}}[\mathbf{w}^{i},\phi_{i}]=\left\{ \mathbf{x}\in\mathbb{R}^{N}\;|\;\left\Vert \mathbf{x}\right\Vert _{1,\mathbf{w}^{i}}\leqslant\phi_{i}\right\} $, of weights $\mathbf{w}^{i}$ and radius $\phi_{i}$. Eventually, with respect to the Lagrangian multipliers, $\gamma_{i}$, the values are given by:
\begin{equation}
    \gamma_{i}=\max\left\{ 0,\frac{2c_{S}}{\left\Vert \mathbf{w}^{i}\right\Vert ^{2}}\left(\left\Vert \mathbf{a}^{i}\right\Vert _{1,\mathbf{w}^{i}}-\phi_{i}\right)\right\} 
    \coma
\end{equation}
with $i=1,2,\ldots,K$.\\

\subsubsection*{Step II -- Dictionary Update}
In the second step of the alternating minimization (see \Eq{\ref{Eq:StepII}}), the optimization task at the $t^{\mathrm{th}}$ step is given by:
\[
	\mathbf{D}^{[t+1]}=\min_{\mathbf{D}}\psi_{D}(\mathbf{D},\mathbf{D}^{[t]})\quad\text{s.t.}\quad\mathbf{D}\in\mathfrak{D}_{\delta}
	\punto
\]

Regarding to the surrogate function, $\psi_{D}$, fixed to the current estimates the coefficient matrix $\mathbf{S}=\mathbf{S}^{[t]}$, we again use the second order Taylor's expansion (see Appendix D in \cite{ConvexOpt}) to derive the following surrogate function:
\begin{equation}
	\psi_{D}(\mathbf{D},\mathbf{D}^{[t]})=\left\Vert \mathbf{X}-\mathbf{D}\mathbf{S}\right\Vert _{F}^{2}-\left\Vert \mathbf{D}\mathbf{S}-\mathbf{D}^{[t]}\mathbf{S}\right\Vert _{F}^{2}+c_{D}\left\Vert \mathbf{D}-\mathbf{D}^{[t]}\right\Vert _{F}^{2}
	\coma
\end{equation}
where $c_{D}\geqslant\left\Vert \mathbf{S}\mathbf{S}^{T}\right\Vert $ and $\left\Vert \,\cdot\,\right\Vert $ stands for the spectral norm. Observe that the surrogate function is convex and majorizes its corresponding loss function by construction.

Again, the BMM gives freedom for choosing any procedure for solving the current optimization task. For simplicity, we again implemented a Lagrangian relaxation introducing a new penalty term into the original problem:
\begin{equation}
		\min_{\mathbf{D}}\psi_{D}(\mathbf{D},\mathbf{D}^{[t]})+\mathcal{P}_{\gamma}(\mathbf{D})
		\coma
\end{equation}
where $\mathcal{P}_{\gamma}$ is defined according to the constraints over the dictionary:
\begin{equation}
	\mathcal{P}_{\gamma}(\mathbf{D})=\sum_{i=1}^{M}\gamma_{i}\left[(\mathbf{d}_{i}-\bdelta_{i})^{T}(\mathbf{d}_{i}-\bdelta_{i})-c_{\delta}\right]+\sum_{i=M+1}^{K}\gamma_{i}\left(\mathbf{d}_{i}^{T}\mathbf{d}_{i}-c_{d}\right)
	\coma\label{Eq:P_D}
\end{equation}
where $\gamma_{i}$, $i=1,2,\ldots,K$, are the respective $K$ Lagrange multipliers, $\mathbf{d}_{i}$ are the $i^{\mathrm{th}}$ column of the matrix $\mathbf{D}$, $\bdelta_{i}$ is the $i^{\mathrm{th}}$ task-related time course, and $c_{\delta}$ and $c_{d}$ are two user-defined parameters to control the norm and the similarity constraint respectively. 

From the optimization theory, Lagrangian multipliers at the optimal point satisfy their associated KKT conditions:
\begin{align}
	&\gamma_{i}\left(\left\Vert \mathbf{d}_{i}-\bdelta_{i}\right\Vert ^{2}-c_{\delta}\right)=0 && i=1,2,\ldots,M \coma \\
	&\gamma_{i}\left(\left\Vert \mathbf{d}_{i}\right\Vert ^{2}-c_{d}\right)=0 && i=M+1,\ldots,K
	\punto
\end{align}

For simplicity, we can rewrite the penalty term in \Eq{\ref{Eq:P_D}} in a more compact form introducing the following matrix notation: Let's define the mask matrix, $\mathbf{M}\in\mathbb{R}^{M\times K}$, which is a rectangular matrix $(M<K)$ with zero elements everywhere except for ones along its principal diagonal, i.e.,
\begin{equation}
	\mathbf{M}\triangleq(m_{ij})_{ij}\longrightarrow\begin{cases}
m_{ij}=0 & \text{if }i\neq j\\
m_{ij}=1 & \text{if }i=j
\end{cases}
	\punto
\end{equation}

Using this mask matrix and the properties of the trace, we can compact the term, $\mathcal{P}_{\gamma}$, as follows:
\[
	\mathcal{P}_{\mathbf{\Gamma}}(\mathbf{D})=\text{tr}\left[\mathbf{\Gamma}\left(\left(\mathbf{D}-\mathbf{\Delta}\mathbf{M}\right)^{T}\left(\mathbf{D}-\mathbf{\Delta}\mathbf{M}\right)-\mathbf{C}\right)\right]
	\coma
\]
where $\mathbf{\Gamma}$ is a diagonal matrix $\mathbf{\Gamma}=\mathrm{diag}\{\gamma_{1},\gamma_{2},\ldots,\gamma_{K}\}$ and $\mathbf{C}$ is another diagonal matrix given by $\mathbf{C}=\mathrm{diag}\{c_{\delta}\mathbf{1}_{M},c_{d}\mathbf{1}_{K-M}\}$. Therefore, the optimization problems is given by:
\[
	\mathbf{D}^{[t+1]}=\min_{\mathbf{D}}\psi_{D}(\mathbf{D},\mathbf{D}^{[t]})+\mathcal{P}_{\mathbf{\Gamma}}(\mathbf{D})
	\punto
\]

Note that the current problems involves a \emph{convex} loss function. Hence, fixed the coefficient matrix to the current estimate, at the $t^{\mathrm{th}}$ iteration, a global minimum exists.

In this case, since the loss function is fully differentiable, we can find the minimum as the point with zero (matrix) gradient, i.e.:
\begin{equation}
	\nabla_{\mathbf{D}}\left(\psi_{D}(\mathbf{D},\mathbf{D}^{[k]})+\mathcal{P}_{\mathbf{\Gamma}}(\mathbf{D})\right)=\mathbf{0}
	\punto\label{Eq:Grad_II}
\end{equation}
The derivative of the first term is given by:
\[
	\nabla_{\mathbf{D}}\psi_{D}=-2\mathbf{X}\mathbf{S}^{T}+2\mathbf{D}^{[k]}\mathbf{S}\mathbf{S}^{T}+2c_{D}\mathbf{D}-2c_{D}\mathbf{D}^{[k]}
	\coma
\]
and the derivative of the second term is given by:
\[
	\nabla_{\mathbf{D}}\mathcal{P}_{\mathbf{\Gamma}}(\mathbf{D})=2\mathbf{D}\mathbf{\Gamma}-2\mathbf{\Delta}\mathbf{M}\mathbf{\Gamma}
	\coma
\]
Now, solving with respect to $\mathbf{D}$, we obtain:
\begin{equation}
	\mathbf{D}=\left[\mathbf{B}+\frac{1}{c_{D}}\mathbf{\Delta}\mathbf{M}\mathbf{\Gamma}\right]\left(\frac{1}{c_{D}}\mathbf{\Gamma}+\mathbf{I}_{K}\right)^{-1}
	\coma
\end{equation} 
where $\mathbf{B}$ is a matrix defined as:
\[
	\mathbf{B}\triangleq\frac{1}{c_{D}}\left[\mathbf{X}\mathbf{S}^{T}+\mathbf{D}^{[k]}\left(c_{D}\mathbf{I}_{K}-\mathbf{S}\mathbf{S}^{T}\right)\right]
	\coma
\]
and the Lagrange multipliers in $\mathbf{\Gamma}$ are obtained so that to satisfy the corresponding KKT conditions.

Thus, following standard arguments and after some algebraic manipulation, it can be easily demonstrated that the solution of this task is given by:
\begin{equation}
	\mathbf{d}_{i}=\begin{cases}
i \leqslant M & \begin{cases}
\mathbf{b}_{i} & \text{if }\left\Vert \mathbf{b}_{i}-\bdelta_{i}\right\Vert ^{2}\leqslant c_{\delta}\\
\frac{c_{\delta}^{\nicefrac{1}{2}}\left(\mathbf{b}_{i}-\bdelta_{i}\right)}{\left\Vert \mathbf{b}_{i}-\bdelta_{i}\right\Vert }+\bdelta_{i} & \text{otherwise}
\end{cases}\\ \\
i>M & \begin{cases}
\mathbf{b}_{i} & \text{if }\left\Vert \mathbf{b}_{i}\right\Vert ^{2}\leqslant c_{d}\\
\frac{c_{d}^{\nicefrac{1}{2}}}{\left\Vert \mathbf{b}_{i}\right\Vert }\mathbf{b}_{i} & \text{otherwise}
\end{cases}
\end{cases}
\coma
\end{equation}
with $i=1,2,\ldots,K$. Eventually, concerning to their respective Lagrangian multipliers, they are given by:
\begin{equation}
	\gamma_{i}=\begin{cases}
i\leqslant M & \max\left\{ 0,c_{D}\left(\frac{\left\Vert \mathbf{b}_{i}-\bdelta_{i}\right\Vert }{c_{\delta}^{\nicefrac{1}{2}}}-1\right)\right\} \\
i>M & \max\left\{ 0,c_{D}\left(\frac{\left\Vert \mathbf{b}_{i}\right\Vert }{c_{d}^{\nicefrac{1}{2}}}-1\right)\right\} 
\end{cases}
\punto
\end{equation}

For completeness, the Algorithm 1 in the main text summarizes a pseudocode with the main steps of the proposed algorithm. Furthermore, we provided a Matlab implementation of the proposed algorithm, which is available in the AIDL$^{\ref{fnt:AIDL}}$ github repository.

\subsection{Convexity Proofs of the constrained sets}
\label{Sub:Math_Convex}
In the following section, we present the convexity proofs of the proposed constrained sets:
\paragraph{\textbf{Constrained set} $\mathfrak{D}_{\delta}$} Given $M$ different imposed task-related time courses $\bdelta_{1},\bdelta_{2},\ldots,\bdelta_{M}$, and the specific constants $c_{\delta}$ and $c_{d}$, the proposed set is convex:

\begin{proof}
By definition, if $\mathfrak{D}_{\delta}$ is convex, then for any $\mathbf{X},\mathbf{Y}\in\mathfrak{D}_{\delta}$ the matrix $\lambda\mathbf{X}+\left(1-\lambda\right)\mathbf{Y}\in\mathfrak{D}_{\delta}$ $\forall\lambda\in[0,1]$.
	
Then, let $\mathbf{X},\mathbf{Y}\in\mathfrak{D}_{\delta}$ and the real parameter $\lambda\in[0,1]$:
\[
	\mathbf{Z}\coloneqq\lambda\mathbf{X}+\left(1-\lambda\right)\mathbf{Y}\quad\mathfrak{D}_{\delta}\text{ is convex }\Leftrightarrow\mathbf{Z}\in\mathfrak{D}_{\delta}	
	\punto
\]

Thus, the $i^{\mathrm{th}}$ column of $\mathbf{Z}$ is given by $\mathbf{z}_{i}=\lambda\mathbf{x}_{i}+\left(1-\lambda\right)\mathbf{y}_{i}$ with $i=1,2,\ldots,K$.
	
\textit{Question:} Is $\mathbf{Z}\in\mathfrak{D}_{\delta}$? 
\[
	\mathbf{Z}\in\mathfrak{D}_{\delta}\Leftrightarrow\begin{cases}
\text{a) }\left\Vert \mathbf{z}_{i}-\bdelta_{i}\right\Vert ^{2}\leqslant c_{\delta} & i=1,2,\ldots,M\\
\text{b) }\left\Vert \mathbf{z}_{i}\right\Vert ^{2}\leqslant c_{d} & i=M+1,\ldots,K
\end{cases}
\punto
\]
	
With respect to the condition a) for $i=1,2,\ldots,M$:
\[
	\left\Vert \mathbf{z}_{i}-\bdelta_{i}\right\Vert =\left\Vert \mathbf{z}_{i}-\bdelta_{i}+\lambda\bdelta_{i}-\lambda\bdelta_{i}\right\Vert =\left\Vert \mathbf{z}_{i}-\lambda\bdelta_{i}-\left(1-\lambda\right)\bdelta_{i}\right\Vert = 
\]
\[
	=\left\Vert \lambda\mathbf{x}_{i}-\lambda\bdelta_{i}+\left(1-\lambda\right)\mathbf{y}_{i}-\left(1-\lambda\right)\bdelta_{i}\right\Vert \leqslant\left\Vert \lambda\mathbf{x}_{i}-\lambda\bdelta_{i}\right\Vert +\left\Vert \left(1-\lambda\right)\left(\mathbf{y}_{i}-\bdelta_{i}\right)\right\Vert =
\]
\[
	=\lambda\left\Vert \mathbf{x}_{i}-\bdelta_{i}\right\Vert +\left(1-\lambda\right)\left\Vert \mathbf{y}_{i}-\bdelta_{i}\right\Vert \leqslant\lambda c_{\delta}^{\nicefrac{1}{2}}+\left(1-\lambda \right)c_{\delta}^{\nicefrac{1}{2}}=c_{\delta}^{\nicefrac{1}{2}}
\]
\[
	\Rightarrow\left\Vert \mathbf{z}_{i}-\bdelta_{i}\right\Vert ^{2}\leqslant c_{\delta}\punto
\]
For the condition b) with $i=M+1,\ldots,K$:
\[
	\left\Vert \mathbf{z}_{i}\right\Vert =\left\Vert \lambda\mathbf{x}_{i}+\left(1-\lambda\right)\mathbf{y}_{i}\right\Vert \leqslant\left\Vert \lambda\mathbf{x}_{i}\right\Vert +\left\Vert \left(1-\lambda\right)\mathbf{y}_{i}\right\Vert =
\]
\[
	=\lambda\left\Vert \mathbf{x}_{i}\right\Vert +\left(1-\lambda\right)\left\Vert \mathbf{y}_{i}\right\Vert \leqslant\lambda c_{d}^{\nicefrac{1}{2}}+\left(1-\lambda\right)c_{d}^{\nicefrac{1}{2}}=c_{d}^{\nicefrac{1}{2}}
\]
\[
	\Rightarrow\left\Vert \mathbf{z}_{i}\right\Vert ^{2}\leqslant c_{d}
\]
Finally, we obtain that $\mathbf{Z}\in\mathfrak{D}_{\delta}$ for any value of $\lambda\in[0,1]$, consequently by definition $\mathfrak{D}_{\delta}$ in convex.
\end{proof}

\paragraph{\textbf{Constrained set} $\mathfrak{L}_{w}$} Given a specific matrix of weights $\mathbf{W}$ and a vector with the expected number of non-zero values per row $\bphi=[\phi_{1},\phi_{2},\ldots,\phi_{K}]$, this proposed set is convex:
\begin{proof}
By definition, if $\mathfrak{L}_{w}$ is convex, then for any $\mathbf{X},\mathbf{Y}\in\mathfrak{L}_{w}$, the matrix $\lambda\mathbf{X}+\left(1-\lambda\right)\mathbf{Y}\in\mathfrak{L}_{w}$ $\forall\lambda\in[0,1]$.
	
Let $\mathbf{X},\mathbf{Y}\in\mathfrak{L}_{w}$ and the real parameter $\lambda\in[0,1]$:
\[
	\mathbf{Z}\coloneqq\lambda\mathbf{X}+\left(1-\lambda\right)\mathbf{Y}\quad\mathfrak{L}_{w}\text{ is convex }\Leftrightarrow\mathbf{Z}\in\mathfrak{L}_{w}
	\punto
\]
Thus, the $i^{\mathrm{th}}$ row of $\mathbf{Z}$ is given by $\mathbf{z}^{i}=\lambda\mathbf{x}^{i}+\left(1-\lambda\right)\mathbf{y}^{i}$ with $i=1,2,\ldots,K$.

\textit{Question:} Is $\mathbf{Z}\in\mathfrak{L}_{w}$? 
\[
	\mathbf{Z}\in\mathfrak{L}_{w}\Leftrightarrow\left\Vert \mathbf{z}^{i}\right\Vert _{1,\mathbf{w}^{i}}\leqslant\phi_{i}
	\punto
\]
Then, for this case:
\[
	\left\Vert \mathbf{z}^{i}\right\Vert _{1,\mathbf{w}^{i}}=\sum_{j=1}^{N}w_{ij}\left|\lambda x_{ij}+\left(1-\lambda\right)y_{ij}\right|\leqslant\sum_{j=1}^{N}w_{ij}\left(\left|\lambda x_{ij}\right|+\left|\left(1-\lambda\right)y_{ij}\right|\right)=
\]
\[
	=\sum_{j=1}^{N}\lambda w_{ij}\left|x_{ij}\right|+\sum_{j=1}^{N}\left(1-\lambda\right)w_{ij}\left|y_{ij}\right|=\lambda\sum_{j=1}^{N}w_{ij}\left|x_{ij}\right|+\left(1-\lambda\right)\sum_{j=1}^{N}w_{ij}\left|y_{ij}\right|=
\]
\[
	=\lambda\left\Vert \mathbf{x}^{i}\right\Vert _{1,\mathbf{w}^{i}}+\left(1-\lambda\right)\left\Vert \mathbf{y}^{i}\right\Vert _{1,\mathbf{w}^{i}}\leqslant\lambda\phi_{i}+\left(1-\lambda\right)\phi_{i}=\phi_{i}
\]
\[
	\Rightarrow\left\Vert \mathbf{z}^{i}\right\Vert _{1,\mathbf{w}^{i}}\leqslant\phi_{i}
\]
We obtained that $\mathbf{Z}\in\mathfrak{L}_{w}$ for any value of $\lambda\in[0,1]$, consequently by definition $\mathfrak{L}_{w}$ is convex.
\end{proof}

\paragraph{\textbf{Constrained set} $\mathfrak{L}_{W}$} Given the matrix of weights $\mathbf{W}$ and the expected number of non-zeros of the full coefficient matrix $\Phi$, the set $\mathfrak{L}_{W}$ is convex:
\begin{proof}
By definition, if $\mathfrak{L}_{W}$ is convex, then for any $\mathbf{X},\mathbf{Y}\in\mathfrak{L}_{W}$ the matrix $\lambda\mathbf{X}+\left(1-\lambda\right)\mathbf{Y}\in\mathfrak{L}_{W}$ $\forall\lambda\in[0,1]$.

Let $\mathbf{X},\mathbf{Y}\in\mathfrak{L}_{W}$ and a real parameter $\lambda\in[0,1]$:
\[
	\mathbf{Z}\coloneqq\lambda\mathbf{X}+\left(1-\lambda\right)\mathbf{Y}\quad\mathfrak{L}_{W}\text{ is convex }\Leftrightarrow\mathbf{Z}\in\mathfrak{L}_{w}
\]
\textit{Question:} Is $\mathbf{Z}\in\mathfrak{L}_{w}$? 
\[
	\mathbf{Z}\in\mathfrak{L}_{W}\Leftrightarrow\left\Vert \mathbf{Z}\right\Vert _{1,\mathbf{W}}\leqslant\Phi
	\punto
\]
Then,
\[
	\left\Vert \mathbf{Z}\right\Vert _{1,\mathbf{W}}=\text{tr}\left[\left|\mathbf{Z}\right|\mathbf{W}^{T}\right]=
\]
\[
	=\text{tr}\left[\left|\lambda\mathbf{X}+\left(1-\lambda\right)\mathbf{Y}\right|\mathbf{W}^{T}\right]\leqslant\text{tr}\left[\left(\lambda\left|\mathbf{X}\right|+\left(1-\lambda\right)\left|\mathbf{Y}\right|\right)\mathbf{W}^{T}\right]=
\]
\[
	=\text{tr}\left[\lambda\left|\mathbf{X}\right|\mathbf{W}^{T}+\left(1-\lambda\right)\left|\mathbf{Y}\right|\mathbf{W}^{T}\right]=\lambda\text{tr}\left[\left|\mathbf{X}\right|\mathbf{W}^{T}\right]+\left(1-\lambda\right)\text{tr}\left[\left|\mathbf{Y}\right|\mathbf{W}^{T}\right]=
\]
\[
	=\lambda\left\Vert \mathbf{X}\right\Vert _{1,\mathbf{W}}+\left(1-\lambda\right)\left\Vert \mathbf{Y}\right\Vert _{1,\mathbf{W}}\leqslant\lambda\Phi+\left(1-\lambda\right)\Phi=\Phi
	\punto
\]
\[
	\Rightarrow\left\Vert \mathbf{Z}\right\Vert _{1,\mathbf{W}}\leqslant\Phi
	\punto
\]
We then obtained that $\mathbf{Z}\in\mathfrak{L}_{W}$ for any value of $\lambda\in[0,1]$, consequently by definition $\mathfrak{L}_{W}$ is convex.

\end{proof}

\subsection{Convergence Analysis}
\label{Sub:Math_Convergence}
According to the specification of the Block Majorized Minimization (BMM) algorithm, the convergence of the proposed algorithm to a local minimum of the main optimization task (see \Eq{\ref{Eq:MainOpt}}) is guaranteed if the following conditions are satisfied \cite{Conv-MM}:
\begin{enumerate}
	\item The objective function of the original problem is regular.
	\item Each surrogate function is quasi-convex, defined over a convex set and majorizes its corresponding cost function.
	\item Each minimization block has a unique solution.
\end{enumerate}
\mbox{}

\paragraph*{Condition 1} By definition, the main loss function is regular, since $\left\Vert \mathbf{X}-\mathbf{D}\mathbf{S}\right\Vert _{F}^{2}$ is infinitely differentiable and single-valued.\\

\paragraph*{Condition 2} For each block, we obtained the surrogate function based on a second order Taylor's expansion. As we described above, one of the reasons for selecting this kind of surrogate function is because the second order Taylor's expansion is convex and majorize their corresponding loss functions by construction. Besides, the imposed set of constraints for each block, $\mathfrak{D}_{\delta}$, $\mathfrak{L}_{w}$ (and $\mathfrak{L}_{W}$) are convex as we proved it in the previous Section.\\

\paragraph*{Condition 3} For each block, each surrogate function is convex and defined over a convex set. Therefore, each block has a unique solution.\\

Consequently, the proposed algorithm satisfies the convergence conditions of the BMM algorithm, which guarantee that the proposed optimization algorithm converges \emph{monotonically} to a \emph{stationary} point of the main optimization problem \cite{Conv-MM}, \cite{Gen-MM}.

\end{changemargin}
\newpage

\begin{multicols}{2}

\begin{small}
\bibliographystyle{IEEEtran}
\bibliography{MyBiblio}
\end{small}

\end{multicols}

\end{document}